\documentclass[letterpaper]{article} 
\usepackage{aaai2026}  
\usepackage{times}  
\usepackage{helvet}  
\usepackage{courier}  
\usepackage[hyphens]{url}  
\usepackage{graphicx} 
\usepackage{natbib}  
\usepackage{caption} 
\usepackage{algorithm}
\usepackage{algorithmic}
\usepackage{subcaption}
\usepackage{multirow}
\usepackage{xspace}
\usepackage{comment}
\usepackage{tabularray}
\usepackage{colortbl}
\usepackage{tabularx}
\usepackage{arydshln}

\usepackage[table]{xcolor}

\newcommand{\exfull}{$\mathtt{F3}$-$\mathtt{6}$\xspace}
\newcommand{\exfullinit}{$\mathtt{F5}$-$\mathtt{3}$\xspace}
\newcommand{\exnostrat}{-$\mathtt{SD}$\xspace}
\newcommand{\exnoref}{-$\mathtt{CR}$\xspace}
\newcommand{\exnomulti}{-$\mathtt{MC}$\xspace}
\newcommand{\exbaseline}{$\mathtt{Bas}$\xspace}
\newcommand{\exsilver}{$\mathtt{Sil}$\xspace}
\newcommand{\exFF}{$\mathtt{ff}$\xspace}
\newcommand{\exLMC}{$\mathtt{lm}$\xspace}

\newcommand{\costume}{costumed\xspace}
\newcommand{\random}{anonymized\xspace}
\newcommand{\costumeabb}{costume\xspace}
\newcommand{\randomabb}{anonym\xspace}

\newcolumntype{s}{>{\hsize=.7\hsize}X}
\newcolumntype{h}{>{\hsize=1.25\hsize}X}

\usepackage{newfloat}
\usepackage{listings}
\DeclareCaptionStyle{ruled}{labelfont=normalfont,labelsep=colon,strut=off} 
\floatstyle{ruled}
\newfloat{listing}{tb}{lst}{}
\floatname{listing}{Listing}

\DeclareFixedFont{\ttb}{T1}{txtt}{bx}{n}{8} 
\DeclareFixedFont{\ttm}{T1}{txtt}{m}{n}{8}  

\definecolor{deepblue}{rgb}{0,0,0.5}
\definecolor{deepred}{rgb}{0.6,0,0}
\definecolor{deepgreen}{rgb}{0,0.5,0}
\definecolor{gray}{rgb}{0.5,0.5,0.5}

\lstdefinelanguage{ExtendedPython}{
    language=Python,
    morekeywords={
        as, assert, async, await,
        bool, bytearray, bytes,
        classmethod, complex,
        dict, enumerate, filter, float, frozenset,
        getattr, globals, hasattr, hash, help, hex,
        id, input, int, isinstance, issubclass, iter,
        len, list, locals, map, max, min, next, object,
        oct, open, ord, pow, property, range, repr,
        reversed, round, set, setattr, slice, sorted,
        staticmethod, str, sum, super, tuple, type, vars,
        zip,
        True, False, None
    },
    morekeywords={
        BaseException, Exception, ArithmeticError,
        AssertionError, AttributeError, BufferError,
        EOFError, ImportError, IndexError, KeyError,
        KeyboardInterrupt, MemoryError, NameError,
        OSError, OverflowError, RuntimeError, StopIteration,
        SyntaxError, TypeError, ValueError, ZeroDivisionError
    }
}

\newcommand\pythonstyle{\lstset{
    language=ExtendedPython,
    basicstyle=\ttm,
    keywordstyle=\ttb\color{deepblue},
    commentstyle=\color{gray},
    stringstyle=\color{deepgreen},
    emph={__init__,__main__,MyClass},
    emphstyle=\ttb\color{deepred},
    showstringspaces=false,
    frame=tb,
    tabsize=4,
    breaklines=true
}}

\lstnewenvironment{python}[1][]{
    \pythonstyle
    \lstset{#1}
}{}

\title{Improved Generalized Planning with LLMs through Strategy Refinement and Reflection}
\author{
     Katharina Stein\textsuperscript{\rm 1},
    Nils Hodel\textsuperscript{\rm 1},
    Daniel Fi\v{s}er\textsuperscript{\rm 2},
    J\"org Hoffmann\textsuperscript{\rm 1,3},
    Michael Katz\textsuperscript{\rm 4},
    Alexander Koller\textsuperscript{\rm 1}
}
\affiliations{
    \textsuperscript{\rm 1}Saarland Informatics Campus, Saarland University, Saarbr\"ucken, Germany\\
    
    \textsuperscript{\rm 2}Aalborg University, Denmark;
    \textsuperscript{\rm 3}German Research Center for Artificial Intelligence (DFKI), Saarbr\"ucken, Germany\\
    \textsuperscript{\rm 4}IBM T. J. Watson Research Center, Yorktown Heights, NY, USA\\
    Correspondence: kstein@lst.uni-saarland.de

}

\usepackage{bibentry}

\begin{document}

\maketitle

\begin{abstract}
LLMs have recently been used to generate Python programs representing generalized plans in PDDL planning, i.e., plans that generalize across the tasks of a given PDDL domain. 
Previous work proposed a framework consisting of three steps: 
the LLM first generates a summary and then a strategy for the domain, both in natural language, and then implements that strategy as a Python program, that gets debugged on example planning tasks. 
In that work, only one strategy is generated and passed directly to the program generation. If the strategy is incorrect, its implementation will therefore result in an incorrect generalized plan. 
Here, we introduce an approach that generates the strategy in the form of pseudocode and enables automatic debugging of the pseudocode, hence allowing us to identify and fix errors prior to the generation of the generalized plan itself. 
Additionally, we extend the Python debugging phase with a reflection step prompting the LLM to pinpoint the reason for the observed plan failure. Finally, we take inspiration from LLM code generation to produce several program variants and pick the best one. 
Running experiments on 17 benchmark domains with two reasoning and two non-reasoning LLMs, we show that these extensions substantially improve the quality of the generalized plans. Our best performing configuration achieves an average coverage of 82$\%$ across the domains.

\end{abstract}

\begin{links}
     \link{Code and Dataset}{https://github.com/coli-saar/genplan-strategy-refine}
\end{links}

\section{Introduction}
\label{introduction}

Large Language Models (LLMs) have revolutionized a large variety of tasks not only from the field of natural language processing but also from other areas of AI research. One very active area of research deals with LLMs in the context of reasoning problems, and there has been growing interest in using LLMs for symbolic planning in the PDDL language \cite{mcdermott20001998,DBLP:series/synthesis/2019Haslum}.

First approaches use LLMs to generate a plan based on the PDDL or natural language (NL) definition of a task. 
Non-reasoning LLMs tend to not perform well in this set-up \cite[e.g.][]{stein-25-icaps, kambhampati2024position, silver2022pddl}. 
Improvements have been achieved by incorporating thoughts and automatic corrections based on feedback into the process \cite[e.g.][]{stein-25-icaps, stechly2025on}, and reasoning LLMs achieve much better results. Yet scalability to larger tasks still tends to be inferior to the symbolic state of the art \cite[e.g.][]{corrêa20252025planning, valmeekam2025a}, especially on unseen domains.
In addition, even where they scale, these approaches can become costly in the number of LLM calls and processed tokens, being called on every planning instance (sometimes on every state in a plan), and with the number of tokens generated growing linearly in plan length.

\citet{Silver_2024} proposed an approach that has the potential to overcome these issues. Instead of using LLMs to generate plans for individual tasks, they prompt the LLM to produce a \textit{generalized plan}, that generalizes across the tasks of a given PDDL domain \cite[e.g.][]{DBLP:journals/ai/SrivastavaIZ11}. A generalized plan contains branches (if-then-else behavior) and loops to deal with different cases and scaling task size. 
\citet{Silver_2024} show how to use LLMs to generate Python programs representing such plans. This solves the issue regarding cost for LLM calls, as that cost is now per-domain instead of per-task. It also potentially addresses the scalability issue: if the generalized plan is correct, planning tasks of arbitrary size can be solved.

\begin{figure*}
    \centering
    \includegraphics[width=0.85\linewidth]{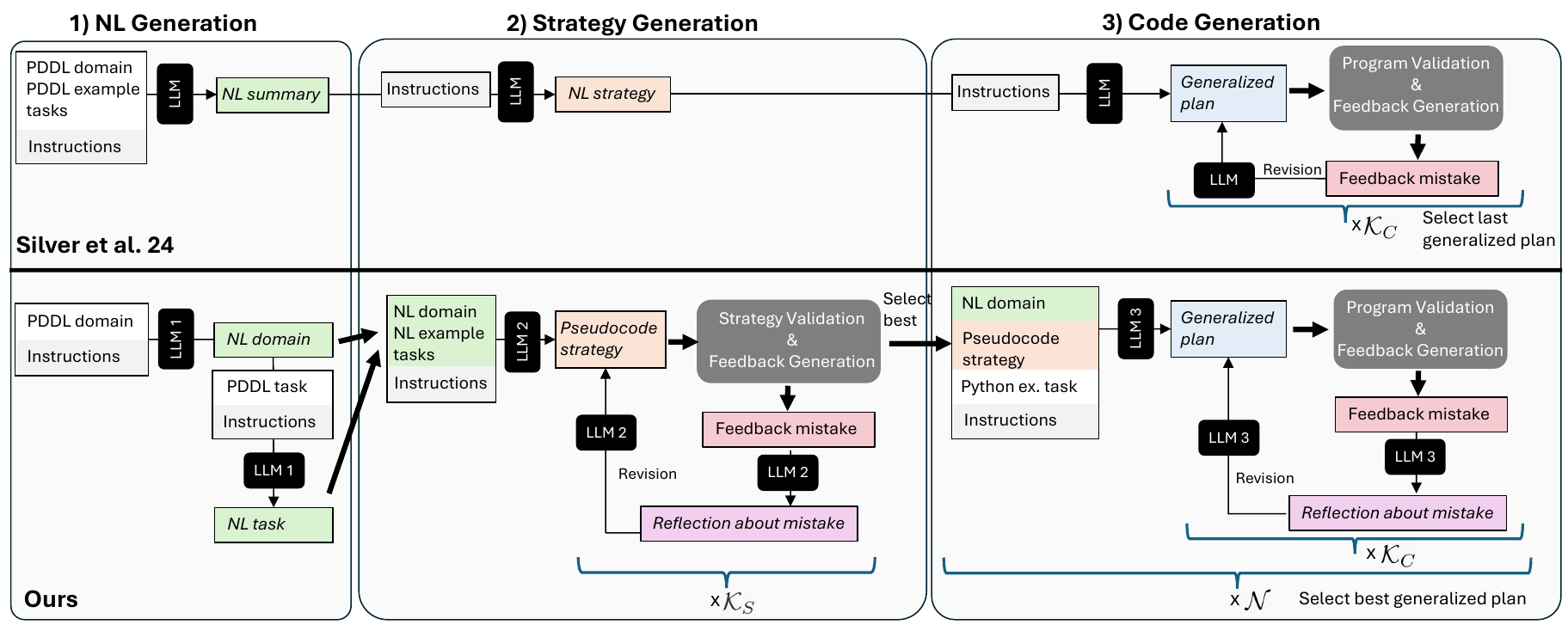}
    \caption{Overview of the framework of \citet{Silver_2024} (top) and our framework (bottom). The main three parts for both are the NL generation, the strategy generation and lastly the code generation, i.e. the generation of the generalized plan.}
    \label{fig:pipelines}
\end{figure*}

Silver et al. let an LLM generate a strategy in NL for the given PDDL domain.
They then prompt the LLM to generate Python code for that strategy, that is then debugged (see Figure \ref{fig:pipelines}, top part).
Silver et al.'s approach achieves good performance on tasks of varying size in 5 out of 7 tested domains when using GPT-4. 
However, when extending their evaluation to a larger set of domains, we find that their approach struggles with generating correct generalized plans. 

A key bottleneck of their approach is the strategy generation step:
they use a simple prompting approach to let the LLM create a generalizing NL strategy, which is directly passed to the code generation. If the strategy is incorrect, the LLM is hence prompted to generate a Python function that implements an inadequate logic.

Here we address this limitation by treating the strategy generation not only as a Chain-of-Thought (CoT) step \cite{wei22cot} but as a central part of the generalized planning framework that is responsible for an important sub-task.
Figure \ref{fig:pipelines} (bottom) provides an overview of our pipeline. 
Our main contribution is an approach that allows us to automatically validate and refine the strategy before passing it to the code generation. Furthermore, our approach generates the strategy in the form of \textit{pseudocode}, that is already closer to the final target structure.
For the refinement, we let an LLM generate PDDL plans for a set of debugging tasks based on the pseudocode, and we check correctness of these plans. We then pass the feedback about errors 
into a reflection step (inspired by e.g. \citeauthor{reflextion23}, \citeyear{reflextion23}; \citeauthor{self-refine23}, \citeyear{self-refine23}). In that step, the LLM is prompted to identify the responsible location in the pseudocode, and the reason for the mistake. The LLM is then prompted to update the pseudocode accordingly. We select the best pseudocode based on the debugging tasks as the strategy to be implemented.

We also introduce some improvements over Silver et al.'s approach in the code generation step. First, we also add a reflection step to the automated debugging of the Python programs. 
Second, we take inspiration from LLM-based code generation to produce several initial versions of the program \citep[e.g.][]{hao2024exploration, wang2024planningnaturallanguageimproves}. 
We pick the best program based on performance on the debugging tasks.

We empirically evaluate our 
method on 17 PDDL domains, including the ones Silver et al.\ ran their experiments on, using GPT-4o, Llama3.3, DeepSeek-V3.2 and Qwen3-Thinking as the LLMs. Compared to Silver et al., our approach improves average performance across domains substantially for all four LLMs. Our approach in combination with DeepSeek solves on average 82$\%$ of the evaluation tasks.   
In 14 domains, our approach achieves perfect coverage for at least one of three runs. We manually verified that our 100$\%$ coverage programs generalize beyond the evaluation data and to all tasks that can be generated using the respective instance generator.
In experiments on a range of "costumed" benchmarks that do not appear in the LLM training data, our approach also exhibits good performance, indicating its generalization capabilities.

\section{Background}
\label{background}

\paragraph{Classical planning.} In classical planning the task is to find a sequence of actions (a plan) that leads from a given initial state into a state that satisfies a goal condition. 
A commonly used formalism to define classical planning tasks is the Planning Domain Definition Language (PDDL) \cite{mcdermott20001998,DBLP:series/synthesis/2019Haslum}.
In PDDL, a planning task is specified by a domain along with a problem. The domain defines the world model, including the predicates for describing the possible world states and all actions that can be used to change the state. Each action has preconditions specifying what needs to be true in order to apply the action, and effects that specify how applying the action changes the world state.
A specific problem file defines a set of available objects, the initial world state and the goal. The solution is a plan consisting of actions from the domain. 

\begin{figure}[t]
    \centering
    \includegraphics[width=0.98\linewidth]{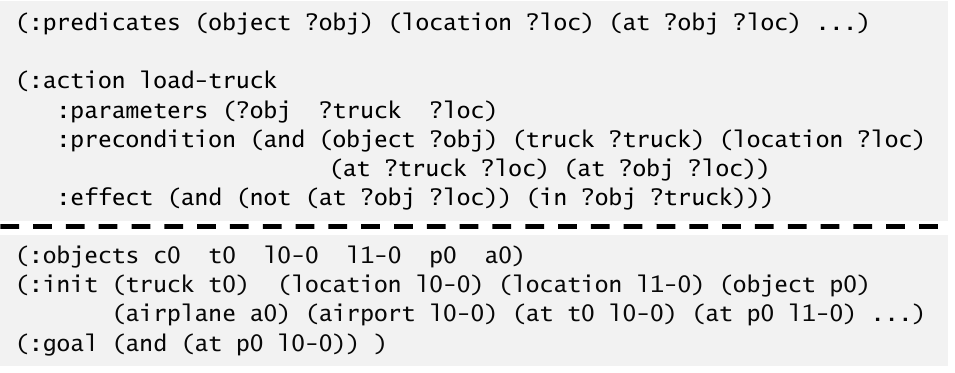}
    \caption{Excerpt from the Logistics PDDL domain (top) and a Logistics PDDL problem (bottom). }
    \label{fig:pddl}
\end{figure}

Figure~\ref{fig:pddl} (top) shows an excerpt from the Logistics domain that models transporting packages with trucks within cities and with planes between cities. 
The action ``load-truck'' can only be executed if the parameter ``?truck'' is a truck, ``?obj'' an object and ``?loc'' a location, and if ``?truck" and ``?obj" are both at ``?loc" (precondition). 
Applying the action changes the location of the package from ?loc to the ?truck (effect).
Figure~\ref{fig:pddl} (bottom) shows part of a task where the goal is to move package ``p0'' from ``l1-0'' to ``l0-0''. 

While there are no formal constraints on the possible initial states and goals, the instance generators used to construct benchmarks usually only generate a subset of all possible tasks.
For example, Logistics benchmarks only include tasks where the goal specifies locations of packages but never e.g. a location of a vehicle.

\paragraph{Generalized planning.} 

\begin{figure*}
    \centering
    \includegraphics[width=1.0\linewidth]{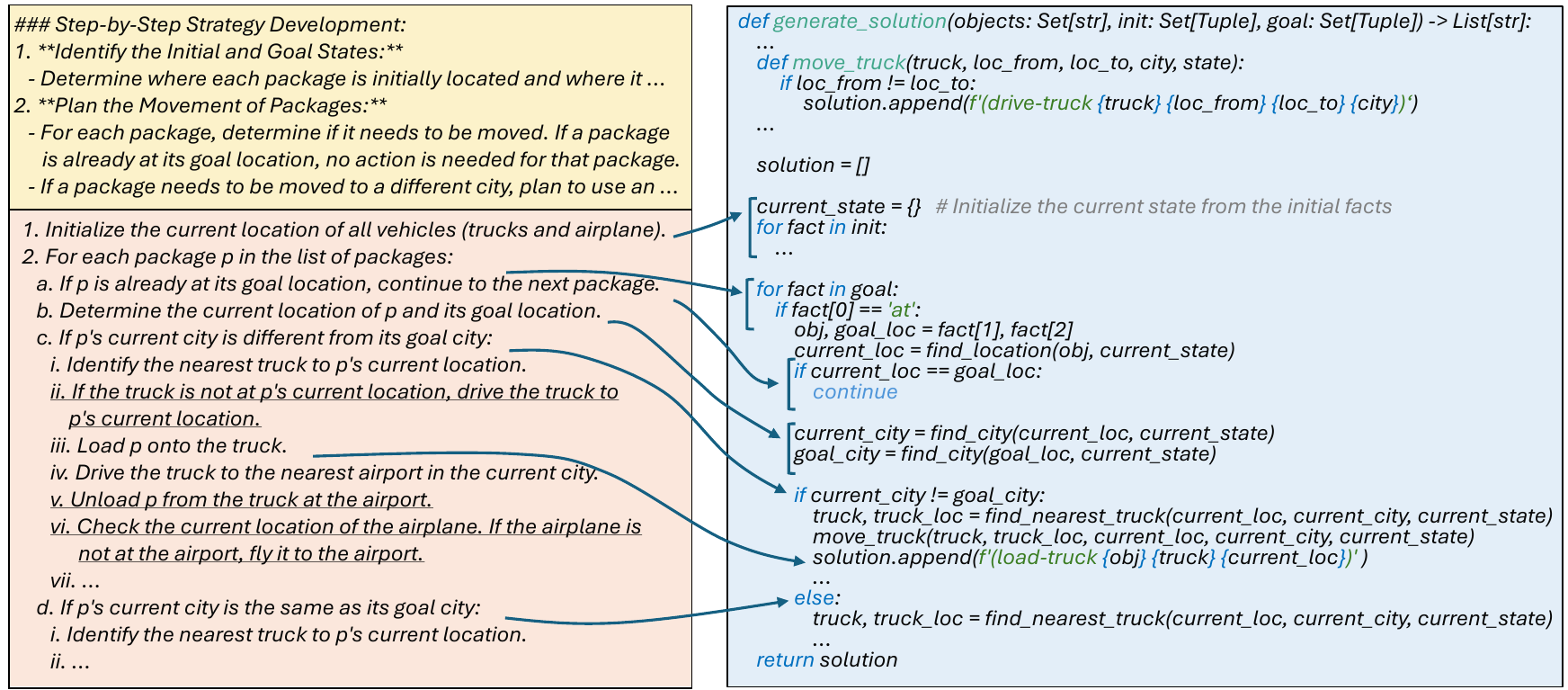}
    \caption{Left: Excerpt of the LLM output for generating a pseudocode strategy for the Logistics domain, consisting of a CoT (top, yellow color) and the pseudocode (bottom, peach color). Underlined steps were initially missing and added during debugging. Right: excerpt of a generalized plan implementing the pseudocode strategy. Arrows illustrate corresponding parts.}
    \label{fig:pseudo_and_code}
\end{figure*}

Generalized planning \cite[e.g.][]{Bonet_Palacios_Geffner_2009,DBLP:journals/ai/SrivastavaIZ11,Jiménez_Segovia-Aguas_Jonsson_2019} seeks plans that generalize over a set of planning tasks. Different variants of this problem have been discussed in the literature. Here, we follow up on Silver et al.'s (2024) work, which generates Python programs intended to generalize over all tasks in a given PDDL domain. The right part of Figure \ref{fig:pseudo_and_code} shows an excerpt of such a program that outputs a plan for a specific input task.

The top part of Figure \ref{fig:pipelines} illustrates Silver et al.'s pipeline. It consists of three steps, of which the first two serve as CoT steps. 
First, the LLM is prompted to generate a short summary of the domain (green color in the figure) based on the PDDL domain file and exemplary PDDL tasks. Afterwards, the LLM receives a prompt stating that there exists ``a simple strategy for solving all tasks in this domain without using search'', and is prompted to tell the strategy (peach color). 

Then, the LLM is asked to implement that strategy as a Python program, i.e. the generalized plan (blue). For this step, it receives the function signature and a short description of the inputs and output. 
Silver et al. then use an automatic debugging approach to iteratively revise the generalized plan based on the outcome of running the program on a set of training tasks. 
If the program interrupts with an error, reaches a timeout or does not return a correct plan - as determined by the plan validator VAL \cite{DBLP:conf/ictai/HoweyLF04}
- the LLM receives a new prompt with a feedback (coral color) and the instruction to fix the code. The feedback includes details about the error that occurred and the PDDL definition of the task for which it occurred. 
This process continues until all training tasks are solved or a maximum number of revisions, $\mathcal{K}_C$, is reached. The last Python program obtained in this manner is selected as the output.

\section{Generating and Refining Pseudocode Strategies}
\label{pseudocode}

Generating generalized plans for planning domains using LLMs is a complex task that poses two main challenges. 
Because the LLM only has access to the domain and example tasks, 
it first needs to abstract away from 
individual tasks to the higher-level logic that generalizes across the domain, i.e. a strategy. Second, the LLM is required to implement that strategy in an executable form, a Python program in our case. The correctness of the final program therefore heavily depends on the quality of the generated strategy, as this serves as a kind of program specification. We therefore treat the strategy generation as a separate subtask in our framework with the dedicated purpose of generating a strategy that is correct and closely matches the specification of the target program, hence reducing the complexity of the code generation itself.

\subsection{Generating Pseudocode Strategies}
Our goal is to improve the quality of the strategies that the LLM is asked to implement in order to shift most of the work beyond the mere conversion into Python to the previous step of the generation framework. We therefore instruct the LLM to generate the strategy in the form of pseudocode that should be detailed and specific enough to be converted into an executable program in a straightforward way. The prompt for this step consists of the NL descriptions of the domain and two example tasks and instructions to think step-by-step (zero-shot CoT, \citeauthor{kojima22zeroCoT}, \citeyear{kojima22zeroCoT}) for developing a strategy that can be turned into a program. 

The left part of Figure \ref{fig:pseudo_and_code} shows part of the output generated for the Logistics domain consisting of the thoughts (top, yellow) and the pseudocode (bottom, peach color) that gets extracted for the subsequent steps. We show inputs to the LLM in regular font and LLM outputs in \textit{italics} in all Figures. The right part of Figure \ref{fig:pseudo_and_code} shows an excerpt of a generalized plan for Logistics. While the pseudocode strategy is expressed in natural language, it includes key words such as ``for each'', ``if'', ``continue''. Furthermore, the steps are enumerated in a structured and nested way that closely matches the overall structure of the final Python program as indicated by the arrows. 

Pseudocode strategies hence express the strategy in a more detailed form, specifically structured in a way that is useful for its actual target use case. If an LLM is simply asked to generate a strategy and produces a simple, natural language summary of it, more work needs to be done (implicitly) to map this strategy into a program.

\subsection{Debugging at the Strategy Level}
If the strategy generated by the LLM is wrong, then an implementation of it will also result in a wrong generalized plan. 
We address the challenge of improving the correctness of the strategy by introducing an approach for automatically validating and refining the pseudocode.

\begin{figure}[ht]
    \centering
    \includegraphics[width=1.0\linewidth]{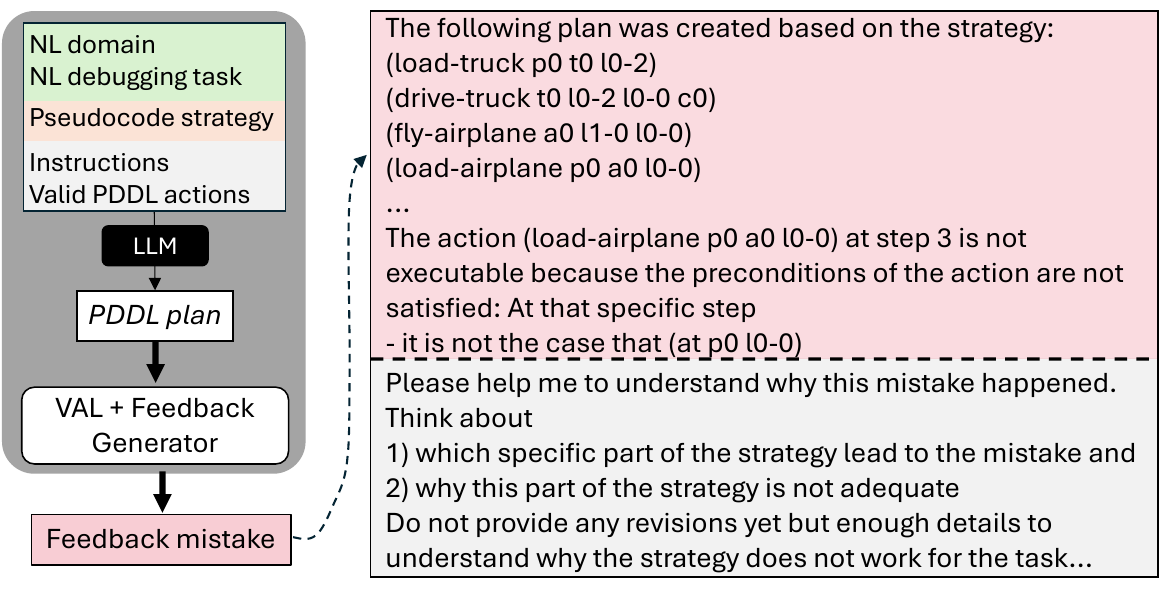}
    \caption{
    Left: approach for generating a PDDL plan for a debugging task based on the pseudocode and obtaining a feedback message (coral color). Right: example prompt for the reflection step consisting of a feedback message for a Logistics task and instructions (grey color).}
    \label{fig:plan_feedback}
\end{figure}

\begin{figure}[ht]
    \centering
    \includegraphics[width=1.0\linewidth]{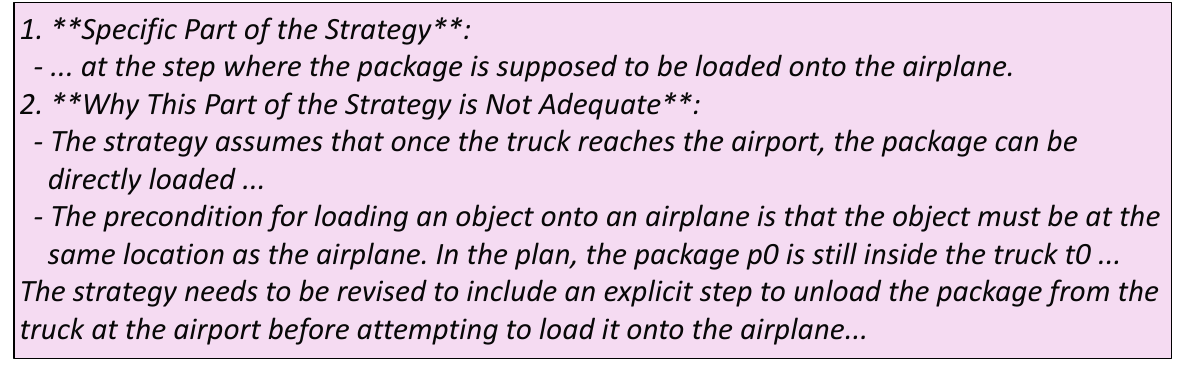}
    \caption{Excerpt of the reflection generated for the example in Figure \ref{fig:plan_feedback}}
    \label{fig:reflection_pseudo}
\end{figure}

Validating the pseudocode strategies without a human in the loop is hard as the pseudocode is not executable, i.e. we cannot run it on 
example tasks and assess the correctness of the outcome. Letting an LLM judge its own output for reasoning problems can even lead to worse performance \cite[e.g.][]{stechly2025on}. Therefore, we introduce an approach that indirectly validates the correctness of the pseudocode using an LLM and a symbolic plan validator as illustrated in Figure \ref{fig:plan_feedback} (left).
We use a small set of tasks from the target domain as \textit{debugging tasks}.
In particular, we provide the pseudocode strategy to an LLM and prompt it to generate the PDDL plan for a given debugging task (in NL) by following the strategy. 
The generated plan is then validated using VAL. If the plan is incorrect, the validation output is converted into a feedback message. For the conversion, we incorporate the feedback generator \citet{stein-25-icaps} used for their experiments on PDDL inputs. 

Instead of directly prompting the LLM to update the pseudocode based on the feedback, we add a reflection step, inspired by approaches that let LLMs reflect about ways to improve over previous outputs \cite[e.g.][]{self-refine23, reflextion23}. 
We combine the feedback about the mistake and the generated plan and with instructions to reflect about the part of the pseudocode that caused the mistake and the reason why that part is incorrect. After generating the reflection response based on that prompt, the LLM is then asked to correct the pseudocode by thinking step-by-step. 
This process is continued until the LLM generates correct plans for all debugging tasks or a maximum number of debugging iterations, $\mathcal{K}_S$, is reached. Then the pseudocode that resulted in the highest number of solved tasks is selected as the pseudocode for the code generation step. 

One bottleneck is that there is no guarantee that the LLM will generate a correct plan given a correct strategy or that a mistake in the plan is actually caused by a mistake in the strategy. However, our approach guarantees that the feedback the LLM achieves about the mistake is always correct with respect to the plan. Additionally, if the pseudocode is missing important details or steps, and the LLM generates a plan reflecting this issue, then our approach makes it possible to automatically find and potentially correct these issues.

Figure \ref{fig:plan_feedback} (right) shows an example of a reflection prompt, including the feedback message for a plan (coral color) that was generated based on the first version of the pseudocode in Figure \ref{fig:pseudo_and_code}, where the underlined steps were missing before debugging. In particular, the step of unloading the package from the truck before loading it onto the airplane (v.) was missing and the LLM generated a plan that was missing that step as well. In order to load a package onto an airplane in Logistics it needs to be at the same location and not in another vehicle. The excerpt of the generated LLM reflection in Figure \ref{fig:reflection_pseudo} illustrates that the LLM correctly identified the mistake and the required extension of the pseudocode. 

\paragraph{NL descriptions.} For the strategy validation approach, we provide the domain and debugging task in NL form. Therefore, we require a separate NL description for each debugging task. We obtain the NL descriptions in a two-step process (see NL Generation, Figure \ref{fig:pipelines}): First, the LLM is prompted to generate the NL domain description given the PDDL domain. Afterwards, the NL description of each debugging task is generated based on its PDDL definition and the PDDL and NL domain descriptions. We also use that NL domain description and two debugging task descriptions as input for the pseudocode generation.

\section{Adding Reflection to Code Debugging}
\label{reflection}

While LLMs perform well on generating short, single-fuction code, generating larger code with several, dependent functions
is complex \cite{hao2024exploration, du24classlevel}. Therefore, the automatic refinement based on feedback 
is important. However, debugging itself can also be complex, especially when the mistake that occurs needs to be traced back to the actual, logical error in the code. We therefore use a similar approach as for the strategy debugging, where the LLM is first asked to reflect on the location and reason of the error before revising the program (see Figure \ref{fig:pipelines}, step 3).

\begin{figure}
    \centering
    \includegraphics[width=0.99\linewidth]{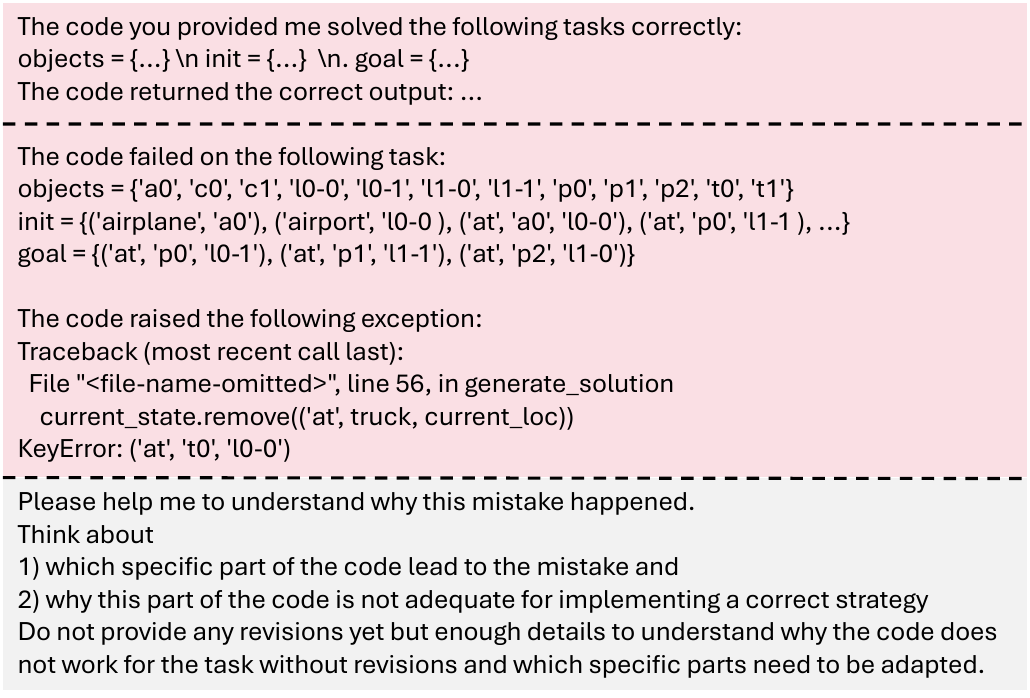}
    \caption{Prompt for the reflection about mistakes in the generated Python program, consisting of the reflection instructions (grey color) and an example feedback message obtained for a Logistics debugging task (coral color).}
    \label{fig:reflect_code}
\end{figure}

We run the generated program on all debugging tasks and create not only negative feedback but also positive feedback as additional information for the debugging. We include all solved debugging tasks in their Python format together with the correct outputs in the feedback prompt. We then add one task for which the program returned an incorrect output together with the feedback message. Figure \ref{fig:reflect_code} shows an example of the positive and negative feedback (coral color) and the reflection instructions (grey color). 
If the code returns an incorrect output, we again use the feedback generator from \citet{stein-25-icaps} to convert the output of VAL. 
Additionally, we also enumerate the steps in the output plan in the feedback message, to make it explicit to which action a feedback of the form ``the action ACTION in step X is not executable'' actually refers. 
We provide more details about the feedback messages and inputs in the Appendix.

\section{Producing Multiple Code Versions}
\label{multiple}

One common approach used for LLM-based code generation is to generate not only a single program 
but several output programs by using a higher temperature or nucleus sampling \cite{holtzman2020curiouscaseneuraltext}, i.e. increasing the cumulative probability threshold based on which the set of tokens to sample from is determined \cite[e.g.][]{hao2024exploration}.
We also propose to generate multiple program versions based on the same strategy,  
but operationalize this in a different way and keep greedy decoding and the temperature of 0. Instead, we randomly change the order in which the objects and the facts of the goal state of the Python example task are presented in the prompt. Apart from this small change, the input prompts are the same for generating all initial programs. 

The different initial programs are generated and debugged one after the other. Specifically, the LLM generates the first program, and the debugged versions of it, as described in the previous section. If none of the programs solves all the debugging tasks, the code generation part is restarted with the newly sampled ordering. 
The code generation stops if a program 
solves all debugging tasks or a defined limit $\mathcal{N}$ of initial programs is reached. Finally, the best program is selected from all generated ones based on the debugging data.

\section{Experiments}
\label{experiments}

\paragraph{Benchmarks.} We consider domains expressed using a STRIPS subset of PDDL that allows variable typing and is restricted to conjunctive conditions with negation. 
We conduct experiments on the seven domains on which \citet{Silver_2024} evaluated their approach, and on 10 of the domains on which \citet{stein-25-icaps} ran LLM action-choice experiments. We remove all action costs from the domains. For each domain, we compose a dataset of tasks taken from previous work and tasks generated by us using available instance generators. We randomly select 6 debugging tasks per domain that are small 
compared to the tasks in the evaluation data. In particular, we only consider tasks for which we can obtain optimal plans, and the number of objects and optimal plan length of each debugging task is among the 16 smallest values of object number and plan length in the overall dataset (see Appendix for more details). 

\paragraph{Costumed and \random\ benchmark variants.} 
Inspired by ideas outside planning \cite{duchnowski-etal-2025-knapsack}, we also run experiments on ``costumed'' variants of all domains from \citet{Silver_2024} that are structurally equivalent but phrased differently and therefore have not been part of the training data of the LLMs.
We define new names for the actions, predicates and objects in the original PDDL, in a way that is still semantically reasonable. 
For example, in the costumed ferry domain, the ferry is a squirrel that needs to jump between trees in order to move nuts. 

Additionally, we follow previous work and create \random benchmark variants without real-world related semantics \cite[e.g.][]{Silver_2024}. In particular, we replace all names with generic names of the form ``action'', ``predicate'', ``object'' and ``type'' and number these (e.g. ``action\_1''). For these  variants we slightly adapt the prompts for the NL generation step to emphasize that all names from the PDDL need to be included in the output in their exact form.
 
\paragraph{Set-up.} We run our experiments using two non-reasoning models, GPT-4o and Llama3.3-70B Instruct, and two reasoning models, DeepSeek-V3.2 and Qwen3-30B-A3B Thinking (see Appendix for details). We use the same prompts for all all models but remove instructions to think for the reasoning models. 

In all experiments, we select the generated program for the final evaluation based on the best performance on the debugging data.  In case of ties, we select the one generated at a later step. 
We also apply the same approach for selecting the pseudocode that is passed to the code generation. 
If a program does not terminate within 45 seconds, it is interrupted and a timeout feedback is generated. 

For each domain and version of the pipeline, we conduct three runs. We split the debugging tasks into three pairs and use a different pair as the examples for the generation of the strategy for each run. All six tasks are used for debugging.

When generating the initial programs, we provide the LLM with one debugging task in Python format and a corresponding plan as an example. 
If the LLM generated a correct plan for any debugging task during the pseudocode validation, we select that task and plan as the example. 
Otherwise, we show a plan generated by an optimal symbolic planner.  

\paragraph{Evaluation.} For running the Python programs on the evaluation tasks, we impose the same time limit of 45s as in debugging.
Our main evaluation metric is coverage, the percentage of evaluation tasks for which the Python program generates a correct plan. 
We report both the average over all runs and the coverage of the run with the highest coverage on the evaluation data.
As the Python program output can depend on the ordering of objects and initial/goal facts in the input, we run 4 random orderings and treat the output as correct only if all runs succeed.

\paragraph{Our framework.} We test our generalized planning framework for two different combinations of the maximum number of initial programs ($\mathcal{N}$) and code debugging steps ($\mathcal{K}_C$). 
For one experiment we set $\mathcal{N}$ = $3$ and $\mathcal{K}_C$ = $6$, resulting in a maximum of 21 generated programs. For the other experiment, we set $\mathcal{N}$ = $5$ and $\mathcal{K}_C$ = $3$, hence increasing the number of initial programs 
while keeping the maximum number of generated programs similar (20). We refer to the two versions as \exfull and \exfullinit.
For both versions we set  $\mathcal{K}_S$ = $5$.

\paragraph{Ablations.}
We conduct three ablation experiments to assess the effect of our pipeline extensions. The base approach for all ablation experiments is \exfull. We assess the effect of generating multiple initial programs by setting $\mathcal{N}$ = $1$ (\exnomulti). In order to test to what extent debugging at the strategy level is beneficial we set $\mathcal{K}_S$ to 0 (\exnostrat). 
Lastly, we prompt the LLM to revise the code directly based on the feedback, to assess the effect of the reflection step (\exnoref). 

\paragraph{Baselines.} We compare the performance of our approach to the framework by \citet{Silver_2024} (\exsilver) and to a re-implementation of their pipeline (\exbaseline).  
For \exbaseline 
we make a number of smaller changes to the original pipeline for a fairer comparison. First, we adapt the phrasing of the prompts to be more similar to our prompts. 
We also separate the three parts of the pipeline and use the output of the previous step as part of the input for the next step, as done in our main framework. 
To account for the fact that no PDDL is available at code generation time,  we provide the definition of the example task and of the failed task in Python format. 
The final program is selected based on the debugging data. 

\paragraph{Symbolic planners.} Our LLM-generated programs come without any guarantees, and are quite different in nature to symbolic planners providing guarantees through search, so a direct comparison is not possible. To nevertheless provide a bit of a measuring line, we run A$^\star$ with the LM-cut heuristic (\exLMC) \cite{lmcut} and GBFS with the FF heuristic (\exFF) \cite{hoffmann2001ff}, as baselines for optimal and satisficing symbolic planning respectively. We ran these planners on Intel Xeon E5-2687W processors with limits of 30m and 8GB. We also report coverage for 
the same 45s limit applied to the execution of generalized plans.

\begin{table*}[t]
  \centering
{\scriptsize
    \centering
    \setlength{\tabcolsep}{2.2pt}
    \begin{tabular}{|l||r|r|>{\columncolor[gray]{0.95}}r|>{\columncolor[gray]{0.95}}r|r|r|r||r|r|>{\columncolor[gray]{0.95}}r|>{\columncolor[gray]{0.95}}r|r|r|r||r|r|>{\columncolor[gray]{0.95}}r|>{\columncolor[gray]{0.95}}r|r|r|r||r|r|>{\columncolor[gray]{0.95}}r|>{\columncolor[gray]{0.95}}r|r|r|r|}
    \hline
        \multirow{3}{*}{Domains} & \multicolumn{14}{c||}{Avg coverage three runs}& \multicolumn{14}{c|}{Coverage best run} \\ 
         & \multicolumn{7}{c||}{GPT-4o} & \multicolumn{7}{c||}{Llama3.3} & \multicolumn{7}{c||}{GPT-4o} & \multicolumn{7}{c|}{Llama3.3}  \\
         & \exsilver & \exbaseline & \exfullinit & \exfull & \exnomulti & \exnostrat & \exnoref
         & \exsilver & \exbaseline & \exfullinit & \exfull & \exnomulti & \exnostrat & \exnoref
         & \exsilver & \exbaseline & \exfullinit & \exfull & \exnomulti & \exnostrat & \exnoref
         & \exsilver & \exbaseline & \exfullinit & \exfull & \exnomulti & \exnostrat & \exnoref \\ 
         \hline\hline
         \multicolumn{29}{|l|}{\qquad\quad Domains from \citet{Silver_2024}}\\
         \hline
         delivery & \textbf{100} & \textbf{100} & \textbf{100} & \textbf{100} & \textbf{100} & \textbf{100} & \textbf{100} & \textbf{100} & \textbf{100} & \textbf{100} & \textbf{100} & 67 & \textbf{100} & \textbf{100} & \textbf{100} & \textbf{100} & \textbf{100} & \textbf{100} & \textbf{100} & \textbf{100} & \textbf{100} & \textbf{100} & \textbf{100} & \textbf{100} & \textbf{100} & \textbf{100} & \textbf{100} & \textbf{100} \\
        ferry & 33 & \textbf{100} & \textbf{100} & \textbf{100} & 35 & \textbf{100} & \textbf{100} & \textbf{100} & 67 & \textbf{100} & \textbf{100} & \textbf{100} & \textbf{100} & \textbf{100} & \textbf{100} & \textbf{100} & \textbf{100} & \textbf{100} & \textbf{100} & \textbf{100} & \textbf{100} & \textbf{100} & \textbf{100} & \textbf{100} & \textbf{100} & \textbf{100} & \textbf{100} & \textbf{100} \\
        gripper & 79 & \textbf{100} & \textbf{100} & 88 & \textbf{100} & \textbf{100} & \textbf{100} & 64 & 55 & \textbf{100} & 67 & \textbf{100} & 88 & 88 & \textbf{100} & \textbf{100} & \textbf{100} & \textbf{100} & \textbf{100} & \textbf{100} & \textbf{100} & 64 & \textbf{100} & \textbf{100} & \textbf{100} & \textbf{100} & \textbf{100} & \textbf{100} \\
        heavy & \textbf{100} & 67 & \textbf{100} & \textbf{100} & \textbf{100} & \textbf{100} & \textbf{100} & \textbf{100} & \textbf{100} & \textbf{100} & 88 & 92 & 97 & 53 & \textbf{100} & \textbf{100} & \textbf{100} & \textbf{100} & \textbf{100} & \textbf{100} & \textbf{100} & \textbf{100} & \textbf{100} & \textbf{100} & \textbf{100} & \textbf{100} & \textbf{100} & \textbf{100} \\
        hiking & \textbf{100} & 0 & 33 & 67 & 0 & 0 & 67 & \textbf{100} & 43 & 95 & 67 & 81 & \textbf{100} & 67 & \textbf{100} & 0 & \textbf{100} & \textbf{100} & 0 & 0 & \textbf{100} & \textbf{100} & \textbf{100} & \textbf{100} & \textbf{100} & \textbf{100} & \textbf{100} & \textbf{100} \\
        miconic & 11 & 4 & \textbf{68} & 33 & 0 & 1 & 4 & \textbf{41} & 4 & 5 & 4 & 10 & 0 & 0 & 32 & 12 & \textbf{100} & \textbf{100} & 0 & 3 & 12 & \textbf{100} & 12 & 12 & 9 & 18 & 0 & 0 \\
        spanner & 0 & 6 & 33 & \textbf{67} & 33 & \textbf{67} & 33 & 0 & 0 & 33 & \textbf{67} & 0 & 1 & 12 & 0 & 15 & \textbf{100} & \textbf{100} & \textbf{100} & \textbf{100} & \textbf{100} & 0 & 0 & \textbf{100} & \textbf{100} & 0 & 3 & 35 \\
         \hline\hline
        \multicolumn{29}{|l|}{\qquad\quad Additional Domains}\\
        \hline
        beluga & 0 & 0 & 0 & 0 & 0 & 0 & 0 & \textbf{7} & 0 & 0 & 0 & 0 & 0 & 0 & 0 & 0 & 0 & 0 & 0 & 0 & 0 & \textbf{10} & 0 & 0 & 0 & 0 & 0 & 0 \\
        blocksw. & 2 & \textbf{12} & 6 & 7 & 6 & 5 & 4 & \textbf{50} & 0 & 14 & 11 & 5 & 8 & 11 & 4 & \textbf{20} & 12 & 13 & 8 & 6 & 6 & \textbf{100} & 1 & 20 & 22 & 14 & 12 & 14 \\
        goldminer & 6 & 0 & 4 & \textbf{11} & 2 & 3 & 2 & \textbf{3} & 0 & 0 & 1 & 0 & 2 & 0 & 14 & 0 & 6 & \textbf{24} & 6 & 6 & 6 & \textbf{5} & 0 & 0 & 4 & 0 & \textbf{5} & 0 \\
        grippers & \textbf{100} & 33 & \textbf{100} & \textbf{100} & 71 & \textbf{100} & \textbf{100} & 71 & \textbf{100} & \textbf{100} & \textbf{100} & \textbf{100} & \textbf{100} & \textbf{100} & \textbf{100} & \textbf{100} & \textbf{100} & \textbf{100} & \textbf{100} & \textbf{100} & \textbf{100} & 93 & \textbf{100} & \textbf{100} & \textbf{100} & \textbf{100} & \textbf{100} & \textbf{100} \\
        logistics & 2 & 45 & \textbf{100} & 94 & 94 & 77 & 74 & 7 & 12 & 42 & 46 & \textbf{60} & 21 & 16 & 6 & \textbf{100} & \textbf{100} & \textbf{100} & \textbf{100} & \textbf{100} & \textbf{100} & 14 & 19 & 94 & \textbf{100} & 83 & 26 & 41 \\
        minigrid & 0 & 31 & 48 & \textbf{61} & 37 & 36 & 42 & 21 & 26 & \textbf{65} & 51 & 41 & 53 & 46 & 0 & 42 & 54 & \textbf{72} & 68 & 42 & 47 & 42 & 37 & \textbf{82} & 60 & 42 & 64 & 54 \\
        rovers & 0 & 0 & \textbf{7} & 0 & 0 & 1 & 0 & 0 & 0 & 0 & 0 & 0 & 0 & 0 & 0 & 0 & \textbf{20} & 0 & 0 & 4 & 0 & 0 & 0 & 0 & 0 & 0 & 0 & 0 \\
        satellite & 33 & 48 & \textbf{69} & 29 & 67 & 60 & 45 & 31 & 4 & 36 & 36 & 35 & 32 & \textbf{37} & 60 & 68 & \textbf{100} & 44 & 72 & \textbf{100} & 52 & 36 & 12 & \textbf{44} & \textbf{44} & \textbf{44} & \textbf{44} & \textbf{44} \\
        transport & 0 & 0 & 33 & \textbf{67} & 0 & 0 & 0 & 0 & 0 & \textbf{60} & 26 & 19 & 59 & 33 & 0 & 0 & \textbf{100} & \textbf{100} & 0 & 0 & 0 & 0 & 0 & 89 & 79 & 57 & \textbf{100} & \textbf{100} \\
        visitall & 70 & 80 & 80 & \textbf{100} & 33 & 78 & 51 & 20 & 15 & 60 & 52 & 55 & 88 & 47 & \textbf{100} & \textbf{100} & \textbf{100} & \textbf{100} & \textbf{100} & \textbf{100} & \textbf{100} & 35 & 20 & 81 & \textbf{100} & \textbf{100} & \textbf{100} & 91 \\

         \hline\hline
         Avg & 37 & 37 & 58 & \textbf{60} & 40 & 49 & 48 & 42 & 31 & \textbf{54} & 48 & 45 & 50 & 42 & 48 & 50 & \textbf{76} & 74 & 56 & 57 & 60 & 53 & 41 & \textbf{66} & \textbf{66} & 56 & 56 & 58 \\
         \hline

    \end{tabular}
\captionof{table}
{\label{tab:results}
Percentage of solved tasks using non-reasoning LLMs for the original framework by \citet{Silver_2024} (\exsilver) and the re-implemented baseline (\exbaseline) and our generalized planning approach with  $\mathcal{N}$ = $3$, $\mathcal{K}_C$ = $6$ (\exfull) and $\mathcal{N}$ = $5$, $\mathcal{K}_C$ = $3$ (\exfullinit). The three ablations \exnomulti, \exnostrat and \exnoref are based on \exfull. We report the average coverage over three runs and coverage of the best run. For both, we show in \textbf{bold} the best generalized planning approach for each model. 
}
}
\end{table*}

\begin{table*}[t]
  \centering
{\scriptsize
    \centering
    \setlength{\tabcolsep}{2.2pt}
    \begin{tabular}{|l||r|r|>{\columncolor[gray]{0.95}}r|>{\columncolor[gray]{0.95}}r||r|r|>{\columncolor[gray]{0.95}}r|>{\columncolor[gray]{0.95}}r|r|r|r||r|r|>{\columncolor[gray]{0.95}}r|>{\columncolor[gray]{0.95}}r||r|r|>{\columncolor[gray]{0.95}}r|>{\columncolor[gray]{0.95}}r|r|r|r||r|r|r|r|}
    \hline
        \multirow{3}{*}{Domains} & \multicolumn{11}{c||}{Avg coverage three runs}& \multicolumn{11}{c||}{Coverage best run} & \multicolumn{4}{c|}{Cov. symbolic} \\ 
         & \multicolumn{4}{c||}{DeepSeek} & \multicolumn{7}{c||}{Qwen3 Thinking} & \multicolumn{4}{c||}{DeepSeek} & \multicolumn{7}{c||}{Qwen3 Thinking} & \multicolumn{2}{c|}{\exLMC} & \multicolumn{2}{c|}{\exFF} \\
         & \exsilver & \exbaseline & \exfullinit & \exfull 
         & \exsilver & \exbaseline & \exfullinit & \exfull & \exnomulti & \exnostrat & \exnoref
         & \exsilver & \exbaseline & \exfullinit & \exfull
         & \exsilver & \exbaseline & \exfullinit & \exfull & \exnomulti & \exnostrat & \exnoref 
         & 45s & 30m & 45s & 30m \\ 
         \hline\hline
         \multicolumn{27}{|l|}{\qquad\quad Domains from \citet{Silver_2024}}\\
         \hline
         delivery & \textbf{100} & \textbf{100} & 67 & \textbf{100} & 70 & \textbf{100} & \textbf{100} & 70 & \textbf{100} & \textbf{100} & \textbf{100} & \textbf{100} & \textbf{100} & \textbf{100} & \textbf{100} & \textbf{100} & \textbf{100} & \textbf{100} & \textbf{100} & \textbf{100} & \textbf{100} & \textbf{100} & 0 & 0 & 100  & 100  \\
        ferry & \textbf{100} & \textbf{100} & \textbf{100} & \textbf{100} & 67 & \textbf{100} & \textbf{100} & \textbf{100} & \textbf{100} & \textbf{100} & \textbf{100} & \textbf{100} & \textbf{100} & \textbf{100} & \textbf{100} & \textbf{100} & \textbf{100} & \textbf{100} & \textbf{100} & \textbf{100} & \textbf{100} & \textbf{100} & 31  & 43& 100 & 100\\
        gripper & 67 & 88 & \textbf{100} & \textbf{100} & 43 & 64 & 43 & 64 & 64 & 76 & 60 & \textbf{100} & \textbf{100} & \textbf{100} & \textbf{100} & 64 & 64 & 64 & 64 & 64 & \textbf{100} & 64 & 15  & 40 & 100 & 100\\
        heavy & 67 & \textbf{100} & \textbf{100} & \textbf{100} & \textbf{100} & \textbf{100} & \textbf{100} & \textbf{100} & \textbf{100} & \textbf{100} & \textbf{100} & \textbf{100} & \textbf{100} & \textbf{100} & \textbf{100} & \textbf{100} & \textbf{100} & \textbf{100} & \textbf{100} & \textbf{100} & \textbf{100} & \textbf{100} & 100 & 100 & 100 & 100\\
        hiking & 33 & 86 & \textbf{100} & 86 & \textbf{76} & 27 & 44 & 33 & 33 & 0 & 33 & \textbf{100} & \textbf{100} & \textbf{100} & \textbf{100} & \textbf{100} & 29 & \textbf{100} & \textbf{100} & \textbf{100} & 0 & \textbf{100} & 100 & 100 & 100 & 100\\
        miconic & 41 & 71 & \textbf{100} & 83 & 46 & 21 & 41 & 41 & 89 & 11 & 9 & \textbf{100} & \textbf{100} & \textbf{100} & \textbf{100} & \textbf{100} & 50 & \textbf{100} & \textbf{100} & \textbf{100} & 12 & 12 & 56  & 62 & 100 & 100\\
        spanner & 33 & \textbf{100} & \textbf{100} & \textbf{100} & 15 & 67 & \textbf{100} & \textbf{100} & \textbf{100} & \textbf{100} & \textbf{100} & \textbf{100} & \textbf{100} & \textbf{100} & \textbf{100} & 44 & \textbf{100} & \textbf{100} & \textbf{100} & \textbf{100} & \textbf{100} & \textbf{100} & 15 & 41 & 15 & 59\\
         \hline\hline
        \multicolumn{27}{|l|}{\qquad\quad Additional Domains}\\
        \hline
        beluga & 0 & 2 & 24 & \textbf{30} & 1 & 0 & 0 & \textbf{7} & 2 & 3 & 0 & 0 & 5 & \textbf{43} & \textbf{43} & 2 & 0 & 0 & \textbf{10} & 7 & 10 & 0 & 0 & 0  & 100 & 100 \\
        blocksworld & 67 & 42 & \textbf{100} & \textbf{100} & 34 & 71 & 42 & 78 & \textbf{85} & 59 & 44 & \textbf{100} & \textbf{100} & \textbf{100} & \textbf{100} & \textbf{100} & \textbf{100} & \textbf{100} & \textbf{100} & \textbf{100} & \textbf{100} & \textbf{100} & 79  & 87 & 100 & 100\\
        goldminer & 66 & 73 & 67 & \textbf{85} & 1 & 3 & 28 & \textbf{37} & 34 & 27 & 8 & \textbf{100} & \textbf{100} & \textbf{100} & 92 & 1 & 8 & 55 & \textbf{65} & 64 & 64 & 14  & 89  & 96 & 99 & 100\\
        grippers & 91 & \textbf{100} & 80 & \textbf{100} & 98 & 98 & \textbf{100} & \textbf{100} & 98 & \textbf{100} & 91 & \textbf{100} & \textbf{100} & \textbf{100} & \textbf{100} & \textbf{100} & \textbf{100} & \textbf{100} & \textbf{100} & \textbf{100} & \textbf{100} & \textbf{100} & 22  & 27 & 100 & 100\\
        logistics & 0 & \textbf{88} & \textbf{88} & \textbf{88} & 4 & 15 & \textbf{94} & 85 & 67 & \textbf{94} & \textbf{94} & 1 & \textbf{100} & \textbf{100} & \textbf{100} & 6 & 21 & \textbf{100} & \textbf{100} & \textbf{100} & \textbf{100} & \textbf{100} & 38  & 45 & 100 & 100\\
        minigrid & \textbf{77} & 62 & 72 & 64 & 14 & 18 & 55 & 53 & 34 & \textbf{61} & 37 & 85 & \textbf{96} & 78 & 78 & 27 & 43 & 77 & 65 & 60 &  \textbf{81} & 57 & 99 & 100 & 100 & 100\\
        rovers & 16 & 20 & 35 & \textbf{60} & 0 & 3 & 13 & \textbf{20} & 15 & 11 & 5 & 48 & \textbf{60} & \textbf{60} & \textbf{60} & 0 & 4 & 28 & \textbf{48} & 36 & 16 & 16 & 88  & 96 & 100 &100\\
        satellite & 15 & 52 & \textbf{63} & 45 & 33 & 35 & \textbf{47} & 43 & 44 & 44 & 44 & 44 & 72 & \textbf{100} & 48 & 52 & 44 & \textbf{48} & 44 & 44 & \textbf{48} & 44 & 76  & 84 & 100 & 100\\
        transport & \textbf{100} & 0 & 67 & 67 & 33 & 0 & 33 & \textbf{91} & 67 & 41 & 33 & \textbf{100} & 0 & \textbf{100} & \textbf{100} & \textbf{100} & 0 & \textbf{100} & \textbf{100} & \textbf{100} & \textbf{100} & \textbf{100} & 15  & 26 & 100 & 100 \\
        visitall & \textbf{83} & 81 & 82 & 82 & 32 & 30 & \textbf{77} & 65 & 74 & 70 & 52 & \textbf{100} & \textbf{100} & \textbf{100} & \textbf{100} & 51 & 50 & \textbf{100} & \textbf{100} & \textbf{100} & \textbf{100} & 54 & 82 & 89 & 99  & 100\\
         \hline\hline
         Avg & 56 & 68 & 79 & \textbf{82} & 39 & 44 & 60 & 64 & \textbf{65} & 59 & 54 & 81 & 84 & \textbf{93} & 89 & 62 & 54 & 81 & \textbf{82} & 81 & 72 & 68 & 53 & 61 & 95 & 98\\
         \hline

    \end{tabular}
\captionof{table}
{\label{tab:results_reasoning}
Percentage of solved tasks using reasoning models for the original framework by \citet{Silver_2024} (\exsilver) and the re-implemented baseline (\exbaseline) and our generalized planning approach with  $\mathcal{N}$ = $3$, $\mathcal{K}_C$ = $6$ (\exfull) and $\mathcal{N}$ = $5$, $\mathcal{K}_C$ = $3$ (\exfullinit). The ablations \exnomulti, \exnostrat and \exnoref are based on \exfull. For both, we show in \textbf{bold} the best generalized planning approach for each model. The symbolic baselines were run for the same time limit as the generalized plans (45s) and for 30m (\exLMC and \exFF).
}
}
\end{table*}

\subsection{Results}

\paragraph{Improvements over baselines.}  
Table \ref{tab:results} and Table \ref{tab:results_reasoning} show the percentage of solved tasks per domain for the best run as well as averaged over all three runs for the non-reasoning models and the reasoning models respectively. Comparing the average of the best baseline (\exsilver, \exbaseline) and our best approach (\exfull, \exfullinit), our approach improves over the baseline by 23 percentage points when using GPT-4o, 20 points when using Qwen Thinking, 14 using DeepSeek and 12 using Llama. Overall, the reasoning models perform better than the non-reasoning models but even for them our approach outperforms the baselines. In particular, the configuration with the highest across-domain average is our \exfullinit approach with DeepSeek.

Comparing the per-domain averages  of \exfull and \exfullinit, we observe that none is consistently better than the other. As the benefit of continuing to debug vs. generating a program from scratch depends the type of mistake and the complexity to fix it, it is likely that a good balance between both depends on the specific domain and program.

For the ablations, we find that removing each of the three ablated parts of the approach has a negative effect on some of the domains. 
Overall, the ablation results illustrate that all three of our contributions are needed to achieve high performance across different domains and LLMs. 

We also provide an overview of the distribution of the types of errors encountered in the automatic evaluation for each of the LLMs in the Appendix. 

\begin{figure}[t]
    \centering
    \includegraphics[width=0.98\linewidth]{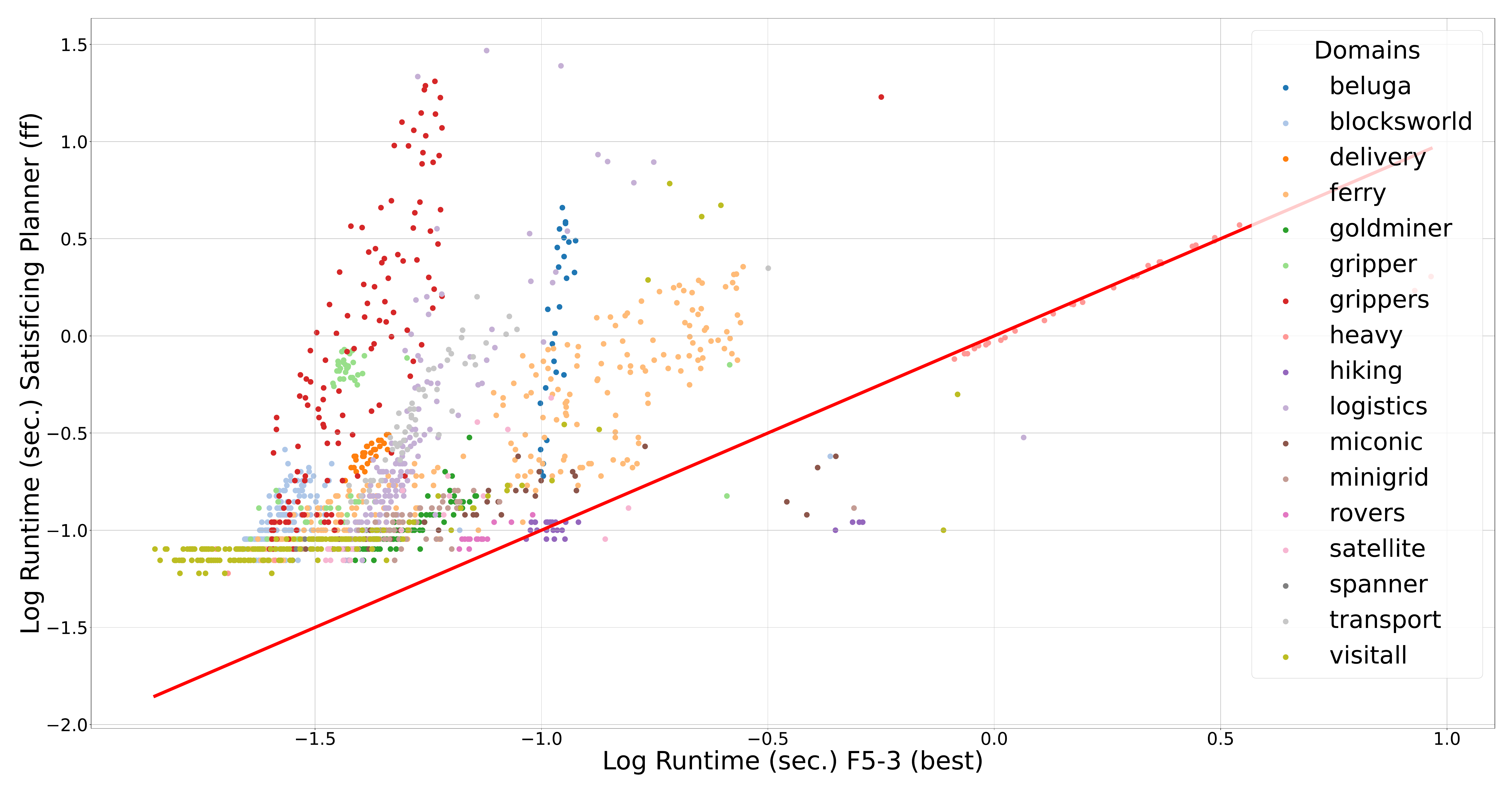}
    \caption{Runtime (log scale) of the best generalized plan by \exfullinit with DeepSeek (x-axis) and of \exFF (y-axis) for each commonly solved task. Diagonal is plotted in red.}
    \label{fig:rt_gbfs}
\end{figure}

\paragraph{Generalization power of our 100\% policies.} We manually analyzed the 100\% coverage programs generated by our \exfullinit configuration using GPT-4o (12 domains) and DeepSeek (14 domains) relative to the respective instance generators. In all these domains, the programs generalize beyond the evaluation dataset and indeed solve all tasks that can be generated with the instance generators. In particular, the programs generalize to tasks of arbitrary size. For example, the programs for Ferry can solve tasks with arbitrary numbers of cars and locations, provided that 
cars are initially not on the ferry and the ferry location is not part of the goal (which are exactly the restrictions inherent in the instance generators).
This shows that although LLMs fail to generalize to larger task sizes and plan lengths when generating plans directly \cite[e.g.][]{valmeekam2025a}, their knowledge from pretraining can be exploited to generate programs that do generalize. 

A full documentation of our manual program analysis is available in the Appendix. Briefly summarized, the main control structure of most of the analyzed programs is a loop that loops over all goal facts (or objects part of the goal) and that contains the code for generating the sub-plan required to arrive at a state satisfying that goal fact (e.g. see first for-loop in Figure \ref{fig:pseudo_and_code}). If the loops themselves correctly implement the sub-strategies and cover all relevant possible state conditions (e.g., whether a truck is already at the package location and if not) then solving tasks with a higher number of objects that require longer plans comes down to simply iterating through the loops more often. 

\paragraph{Comparison to symbolic planners.} 
As the rightmost columns in Table \ref{tab:results_reasoning} show, optimal planning becomes hard for our evaluation tasks within the given time limits, and is often outperformed by the Python programs. But satisficing planning still reigns supreme in coverage, being beaten only in the Spanner domain.

Looking beyond coverage however, the Python programs have substantial advantages. As pointed out above, many of the best programs generalize to the entire domain. Given the polynomial runtime in input task size, the programs are thus bound to eventually outscale any symbolic planner based on search. More generally, program execution is most of the time much faster than plan generation via search. 

To give an assessment of this aspect in our benchmarks (in many of which, state-of-the-art symbolic planners perform quite well), 
Figure \ref{fig:rt_gbfs} compares times for the best Python program (\exfullinit, DeepSeek) vs.\ the satisficing planner \exFF.
Focusing on tasks solved by both, program execution is considerably faster for 97$\%$ of the tasks (note the exponential scaling in Figure \ref{fig:rt_gbfs}; see Appendix for more comparisons).  

This runtime efficacy increase comes at a mild price in plan quality. Comparing plan length on commonly solved tasks, the plans generated by the Python programs are only 1.1 times longer on average than those generated by \exFF.

\paragraph{Cost of generating the programs.} 
Focusing on \exfullinit with DeepSeek, the generation of a program for Heavy took the least time, namely 664s on average, and for Goldminer the most time, almost 5.5h on average. Note however, that we used caching, and retrieving outputs for already processed inputs,  
is faster than generating them the first time. This concerns all parts of the pipeline that are shared between different variants of the framework, e.g. the NL domain description is only generated once per domain in our experiments and then retrieved from the cache for all other runs (with the exception of \exsilver which uses different prompts).  

In sum, almost 15M tokens (input + output) were processed by DeepSeek and \exfullinit for generating the Python programs across all domains. This  corresponds to the negligible cost of ca.\ 5.5 USD for the DeepSeek variant used. 

\begin{table}[t]
{\scriptsize
    \centering
    \setlength{\tabcolsep}{3.5pt}
    \begin{tabular}{|l||r|r|r||r|r|r|}
    \hline
    \multirow{2}{*}{Domain} & \multicolumn{3}{c||}{Avg three runs} & \multicolumn{3}{c|}{Best run} \\
    & original & \costumeabb & \randomabb & original & \costumeabb & \randomabb  \\
    \hline\hline
        delivery & 100 & 100 & 100 & 100 & 100 & 100 \\
        ferry & 100 & 100 & 1 & 100 & 100 & 2 \\
        gripper & 100 & 67 & 33 & 100 & 100 & 100 \\
        heavy & 100 & 100 & 0 & 100 & 100 & 0 \\
        hiking & 33 & 100 & 67 & 100 & 100 & 100 \\
        miconic & 68 & 67 & 33 & 100 & 100 & 100 \\
        spanner & 33 & 0 & 12 & 100 & 0 & 35 \\
    \hline\hline
    Avg & 76 & 76 & 35 & 100 & 86 & 62 \\
    \hline
    \end{tabular}
    \caption{Percentage of solved tasks for \exfullinit with GPT-4o on the original, \costume and \random versions of the domains from \citet{Silver_2024}.}
    \label{tab:costumes}
    }
\end{table}

\paragraph{Results on costumed and \random benchmark variants.} Table \ref{tab:costumes} shows the results of our \exfullinit approach ran with GPT-4o on the anonymized and costumized variants for the domains of \citet{Silver_2024}. 

Regarding the anonymized variants, unsurprisingly (and in line with previous work) we find that LLMs struggle, as no world knowledge can be leveraged when names in a domain carry no information.

Regarding the costumed variants however, interestingly we observe the performance of our LLM-generated programs does not degrade, with the single exception of the Spanner domain. 
This result indicates that the LLMs do not only replicate solutions that have already been part of the pretraining data, but are capable of actual reasoning over potential strategies for a domain, as long as the domains have a connection to real world semantics.

\section{Conclusion}
\label{conclusion}

We show that generalized planning with LLMs can be made substantially more effective through pseudocode strategy refinement, code reflection and generating multiple code candidates. Our approach generates Python programs that achieve an average coverage of 82$\%$ across 17 domains.

In future work, it would be interesting to investigate if and how knowledge about a domain can be exploited to create a more effective set of debugging tasks and potentially extend it automatically during the generation as needed. 
Another important direction, given the lack of intrinsic guarantees and the fundamental limitation to polynomial-time programs, is the combination with symbolic search methods.


\section{Acknowledgments}
The work was partially funded by the Deutsche Forschungsgemeinschaft (DFG, German Research Foundation) under the project number 232722074 – SFB 1102. It was also funded in part by the Deutsche Forschungsgemeinschaft -- GRK 2853/1 “Neuroexplicit Models of Language, Vision, and Action” - project number 471607914.

\bibliography{aaai2026}

\appendix
\newpage

\section{Model Parameters}

We run our experiments with the 
gpt-4o-2024-08-06\footnote{https://platform.openai.com/docs/models/gpt-4o} and 
LLama-3.3-70B-Instruct\footnote{https://huggingface.co/nvidia/Llama-3.3-70B-Instruct-FP8} non-reasoning models and the 
DeepSeek-V3.2-Exp\footnote{https://api-docs.deepseek.com/news/news250929} and 
Qwen3-30B-A3B-Thinking-2507\footnote{https://huggingface.co/Qwen/Qwen3-30B-A3B-Thinking-2507} reasoning models. We abbreviate the model names with GPT, DeepSeek, Llama and Qwen in the following.
For GPT and DeepSeek we use the OpenAI and DeepSeek APIs respectively. 
The experiments with Llama and Qwen were run on two GPUs of type Tesla V100-PCIE-32GB. 

The evaluation of the Python programs was run on the same processors as the symbolic planners, i.e. on Intel Xeon E5-2687W processors.

\begin{table}[h!]
\fontsize{8pt}{8pt}\selectfont
    \centering
    \begin{tabular}{|l|r|r|r|r|}
    \hline
        Parameter & GPT & DeepSeek & Llama & Qwen \\
        \hline\hline
        context window & 128000 & 128000 & 128000 & 262144 \\
        temperature & 0 & NA & 0.7 & 0.6 \\
        max tokens & 16384 & 32000 & 20000 & 81920\\
        top p & NA & NA & 0.8 & 0.95 \\
        top k & NA & NA & 20  & 20 \\
        presence pen. & NA & NA & 0 & 0\\
        seed & 1 & NA & NA & NA \\
        \hline
    \end{tabular}
    \caption{Parameters of the LLMs for the generation.}
    \label{tab:param}
\end{table}

\section{Additional Results.}

\paragraph{Error distributions.} Table \ref{tab:errors} provides an overview of the distribution of error types. Considering all three runs for each configuration, we report the number of runs where the generation itself stopped with an error (Incompl. run). All these errors were due to reaching the maximum number of output tokens or the maximum number of tokens the model can process, i.e. the context window limit. 

For all runs that finished without an error, we consider all final programs that resulted in an error for at least one evaluation task and computed the percentage of runs for which each error type occurred for at least one task. If running the program successfully terminated with an output (i.e. not Python exception nor timeout), we use VAL to validate whether the returned plan is correct, contains actions violating the preconditions (Invalid actions) or does not lead to a state satisfying the goal constraints (Goal not satisfied).

\begin{table}[h!]
\fontsize{8pt}{8pt}\selectfont
    \centering
    \begin{tabular}{|l|r|r|r|r||r|}
    \hline
         \multirow{2}{*}{Model} & \multicolumn{1}{c|}{Python}  & \multirow{2}{*}{Timeout} & Invalid  & Goal not  & \multicolumn{1}{c|}{Incompl.} \\
         & except. &  & actions & satisfied &  \multicolumn{1}{c|}{run} \\
         \hline\hline
        GPT & 23$\%$ & 14$\%$ & 37$\%$ & 27$\%$ & 0 \\
        DeepSeek  & 18$\%$ & 15$\%$ & 35$\%$ & 32$\%$ & 11 \\
        Llama & 13$\%$ & 15$\%$ & 35$\%$ & 32$\%$ & 3 \\
        Qwen & 13$\%$ & 12$\%$ & 51$\%$ & 24$\%$ & 11 \\
         \hline
    \end{tabular}
    \caption{The number of runs (out of all configurations and domains) for which the generation of the programs stopped with an error (Incompl. run) and the percentage of runs where one of the four error types occurred for at least one task out of all runs with any error.}
    \label{tab:errors}
\end{table}

\paragraph{Runtimes and Plan Lengths.}
We compare the time required for generating a plan using \exFF and \exfullinit for all commonly solved tasks as well as the length of these generated plans. 
We report the average over all per-task ratios, computed by dividing the runtime and plan length for \exfullinit by the corresponding value for \exFF. 
Additionally, we report the percentage of tasks for which the Python program is at least as fast as \exFF and the percentage of tasks for which the Python program generates plans of shorter or equal length.

Table \ref{tab:runtimesff} shows the result for the comparison to \exfullinit (best run) with each of the four tested LLMs. All four LLMs generate Python programs that are faster for a large majority of tasks. The programs generated by DeepSeek are the fastest - relative to \exFF - requiring only 0.36 time the time of \exFF on average while the programs generated by Llama have runtimes closer to \exFF.

\begin{figure*}
    \centering
    \subfloat[GPT \exfullinit vs. \exFF]
    {\includegraphics[width=0.358\textwidth]{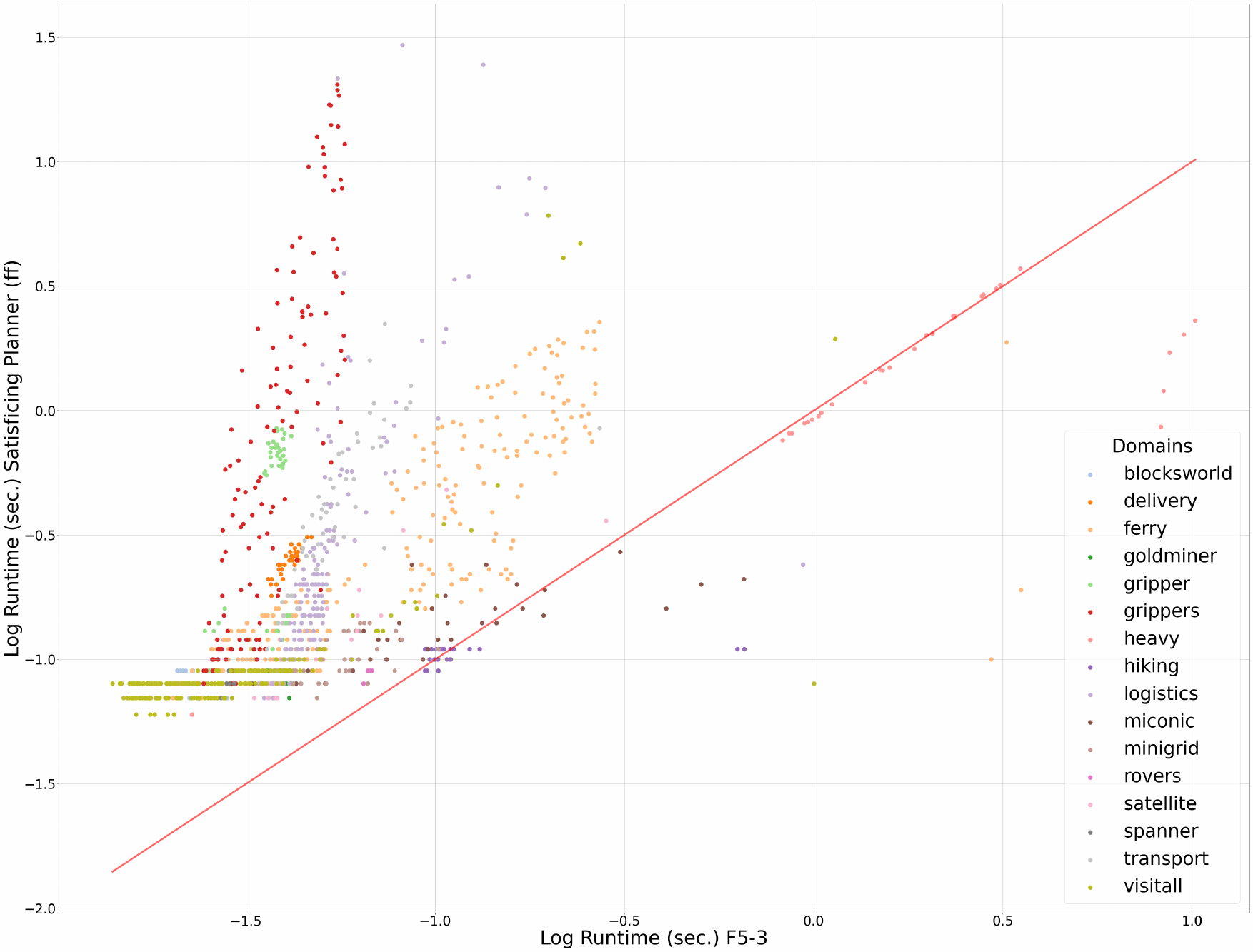}\label{fig:time_gpt_gbfs}}
    \qquad
    \subfloat[GPT \exfullinit vs. \exLMC]
    {\includegraphics[width=0.358\textwidth]{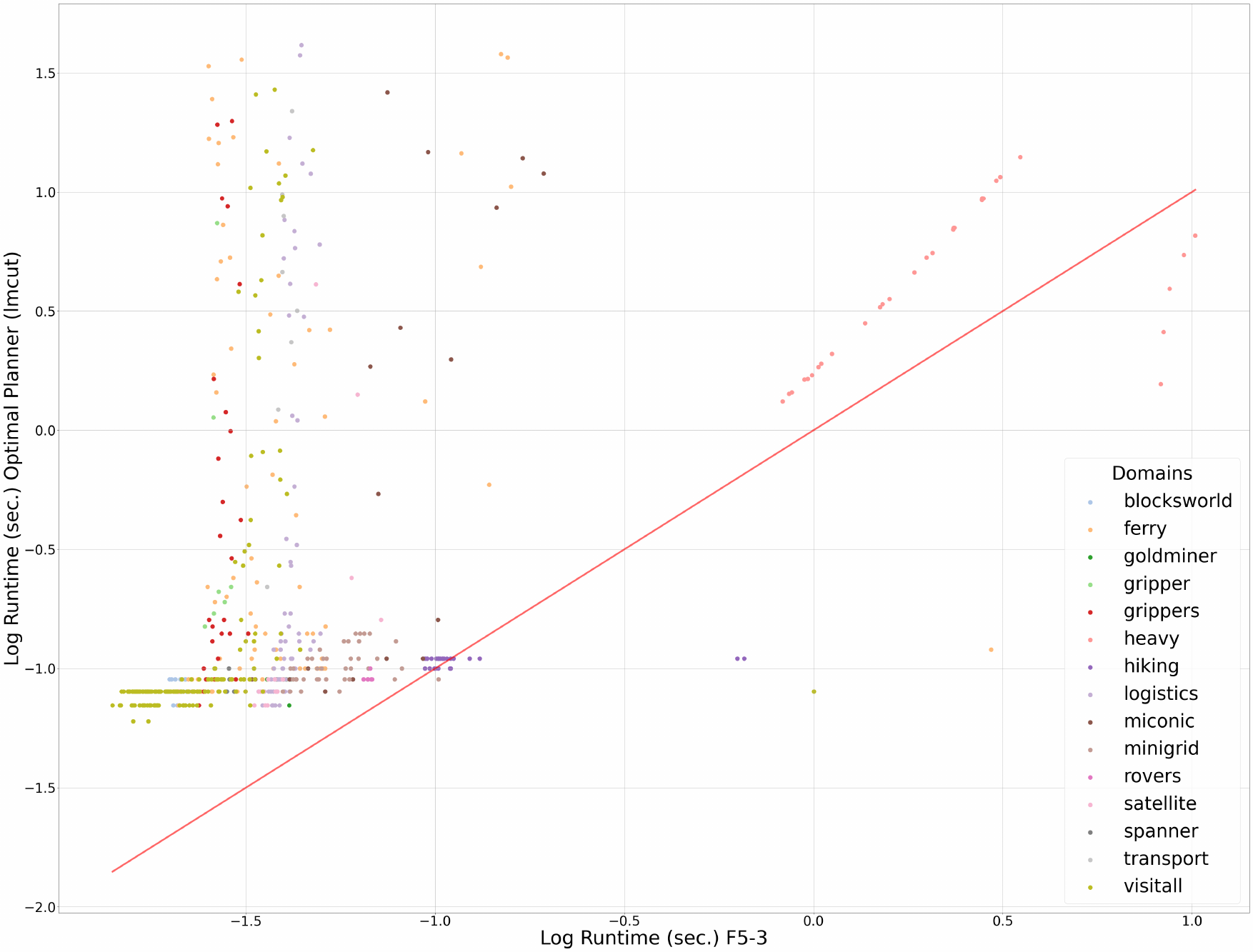}\label{fig:time_gpt_optimal}}
    
    \subfloat[DeepSeek \exfullinit vs. \exFF]
    {\includegraphics[width=0.358\textwidth]{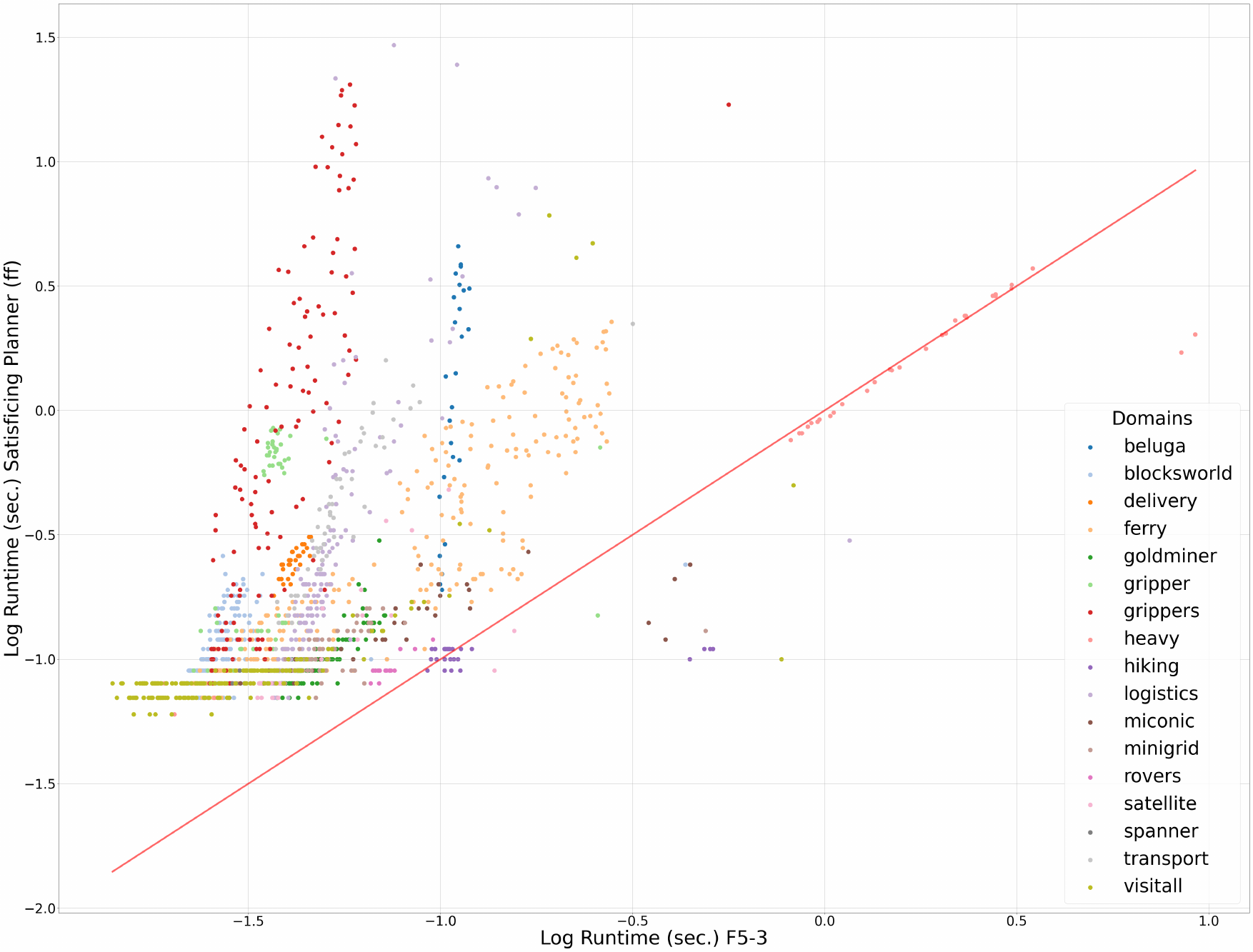}\label{fig:time_deepseek_gbfs}}
    \qquad
    \subfloat[DeepSeek \exfullinit vs. \exLMC]
    {\includegraphics[width=0.358\textwidth]{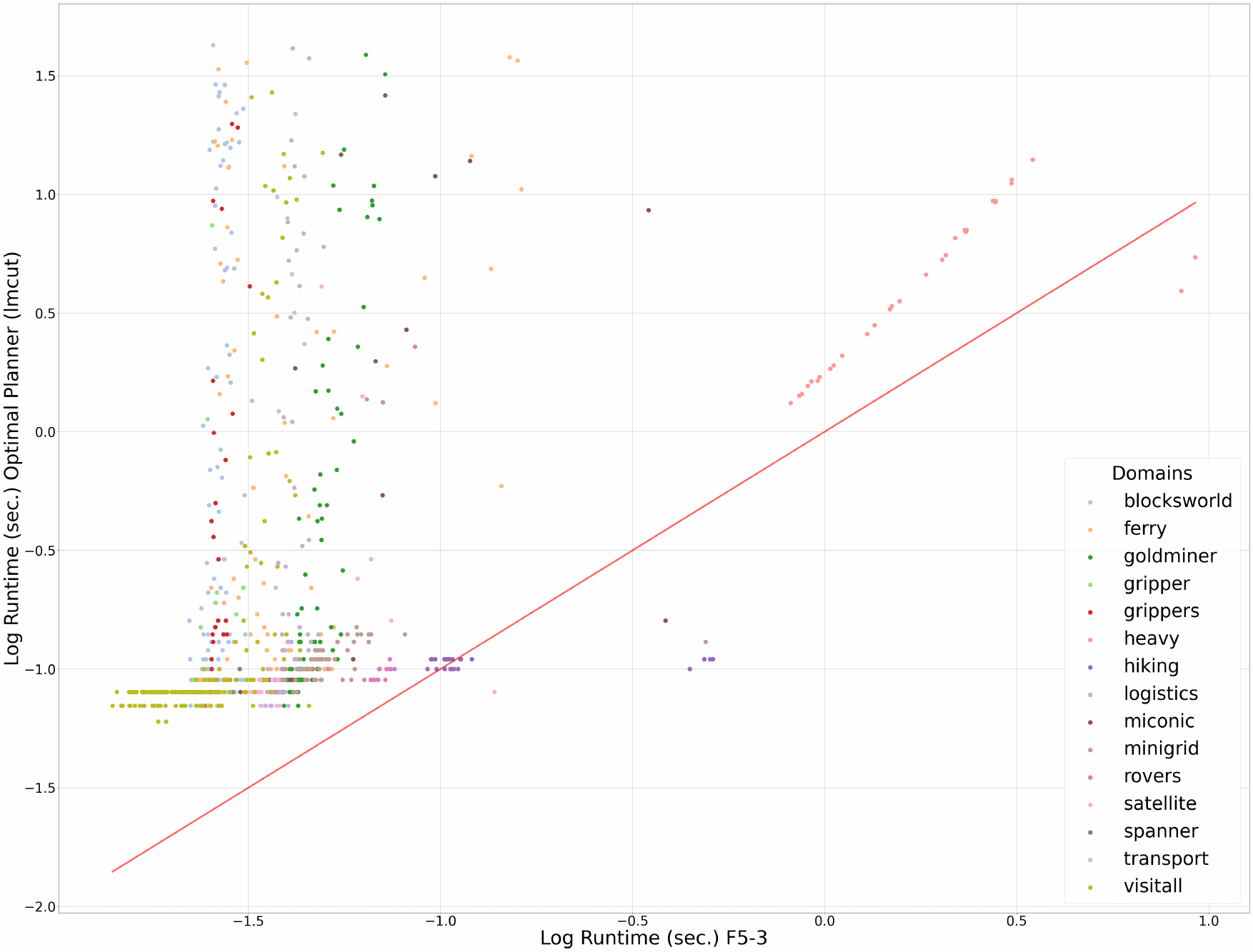}\label{fig:time_deepseek_optimal}}
    
    \subfloat[Llama \exfullinit vs. \exFF]
    {\includegraphics[width=0.358\textwidth]{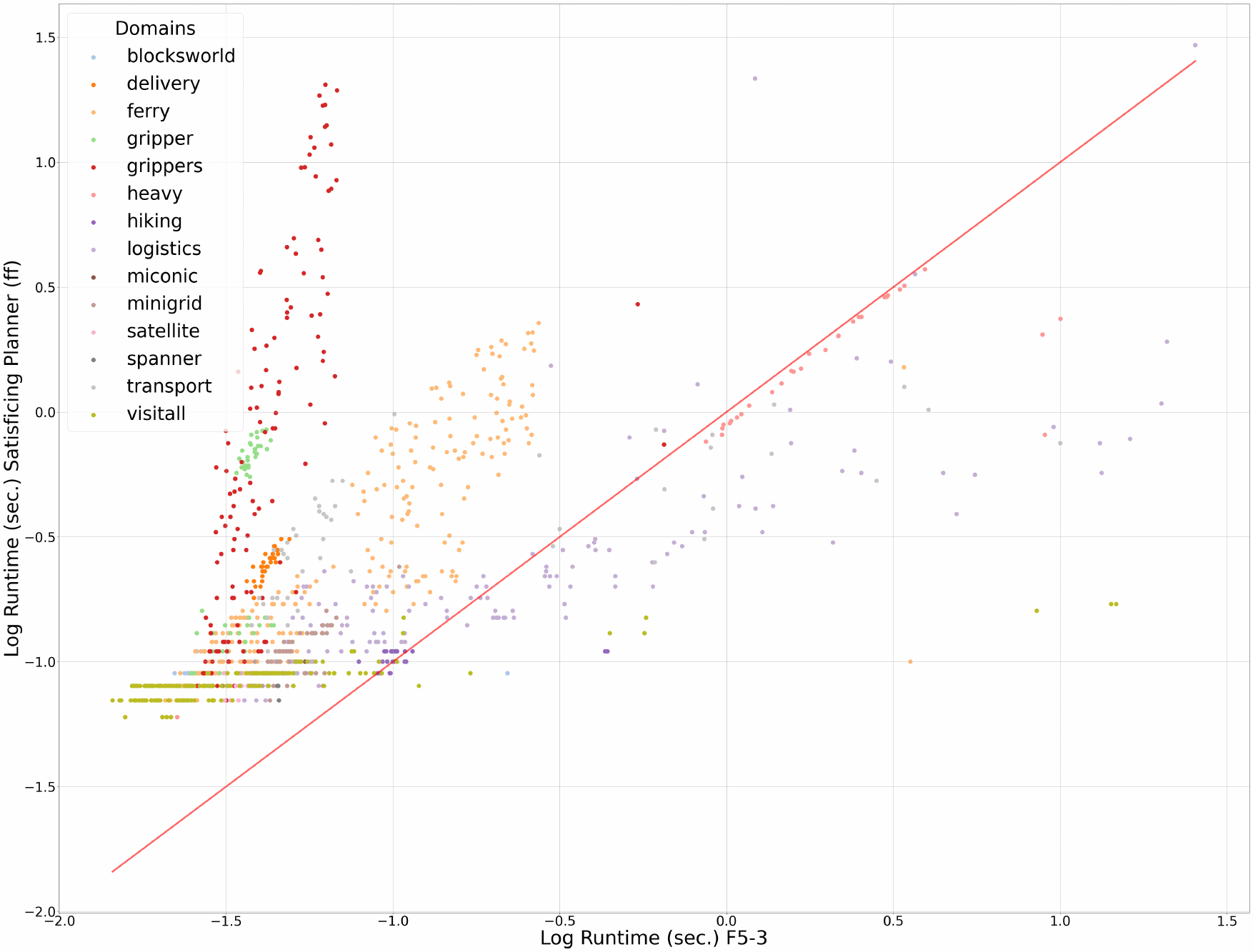}\label{fig:time_llama_gbfs}}
    \qquad
    \subfloat[Llama \exfullinit vs. \exLMC]
    {\includegraphics[width=0.358\textwidth]{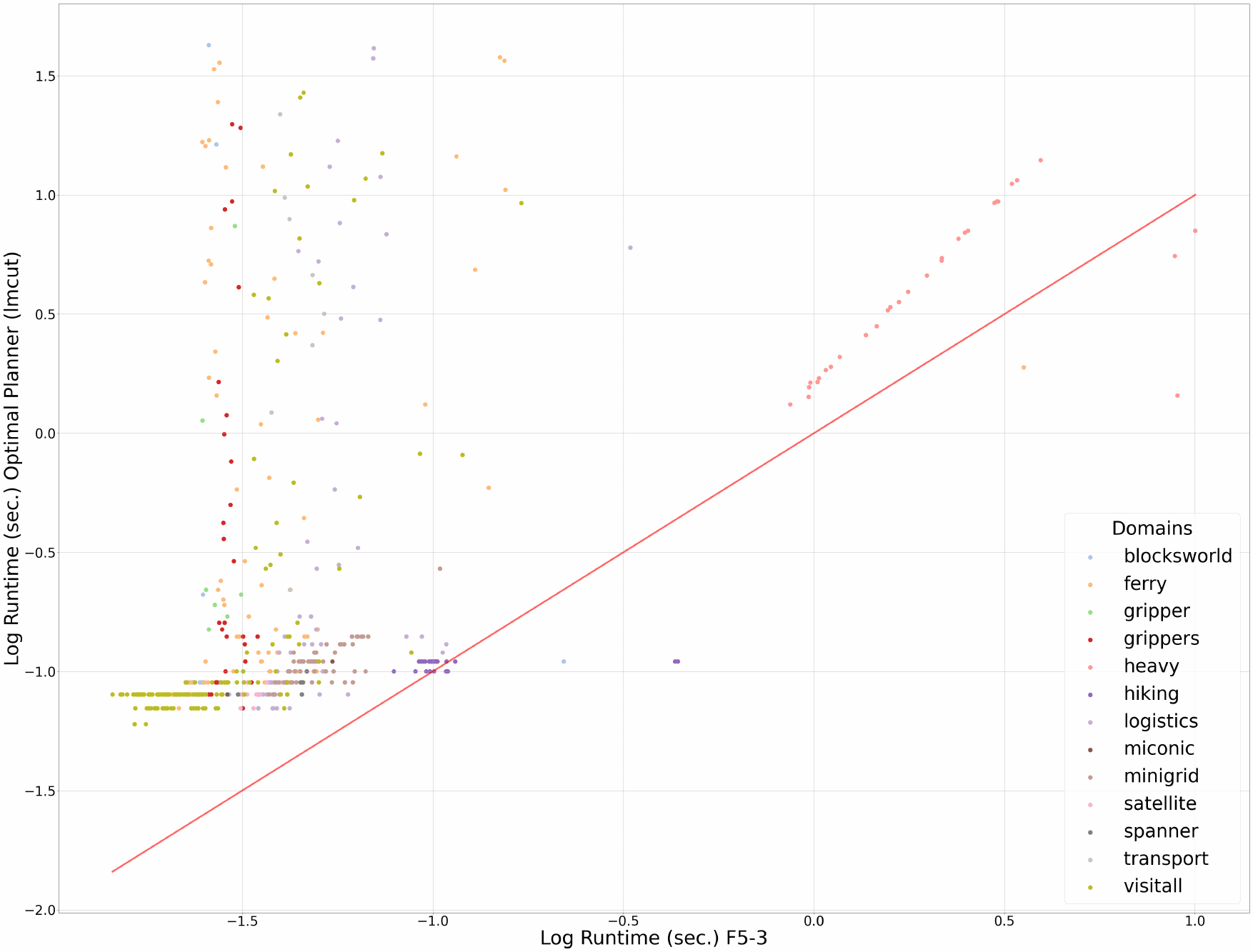}\label{fig:time_llama_optimal}}
    
    \subfloat[Qwen \exfullinit vs. \exFF]
    {\includegraphics[width=0.358\textwidth]{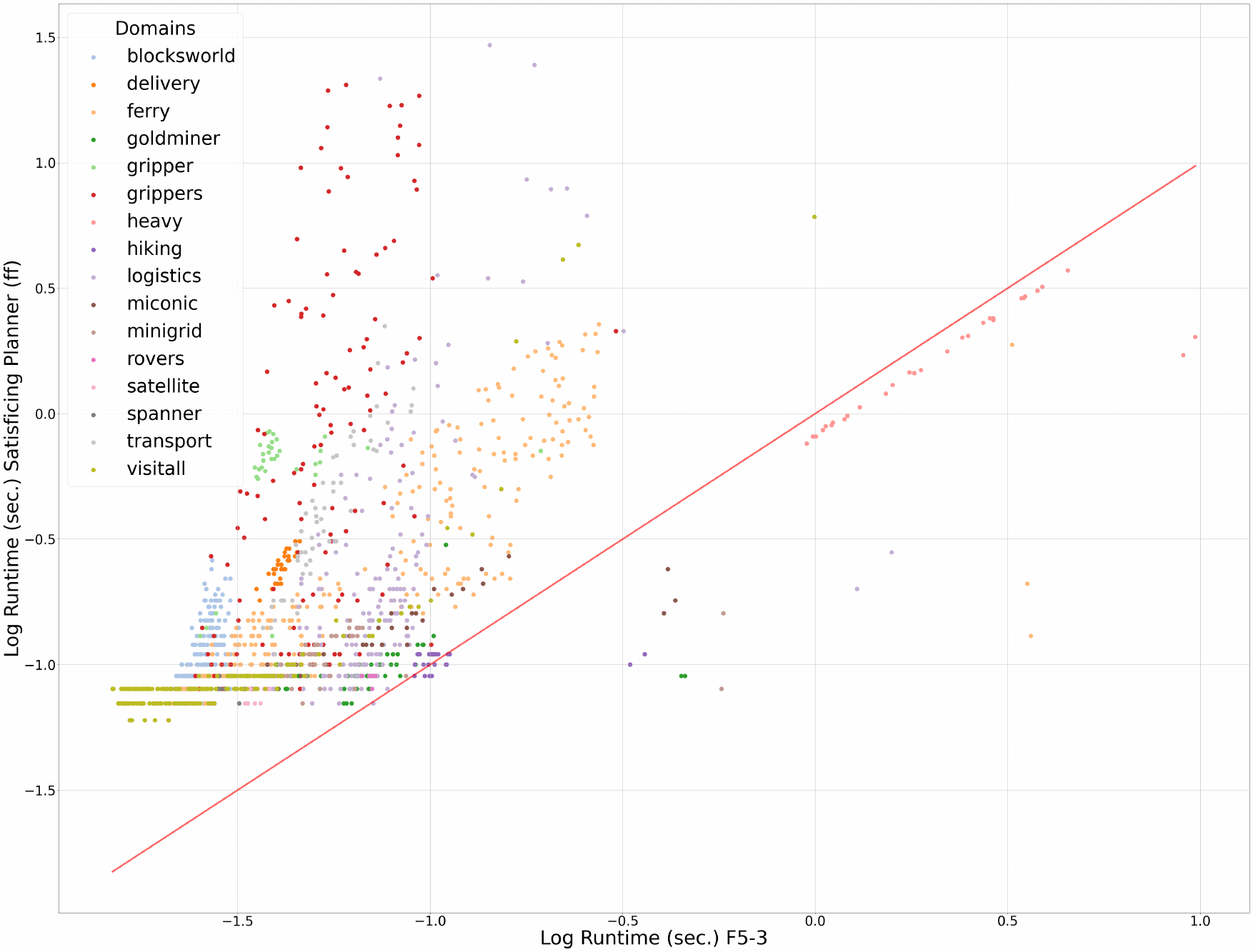}\label{fig:time_qwent_gbfs}}
    \qquad
    \subfloat[Qwen \exfullinit vs. \exLMC]
    {\includegraphics[width=0.358\textwidth]{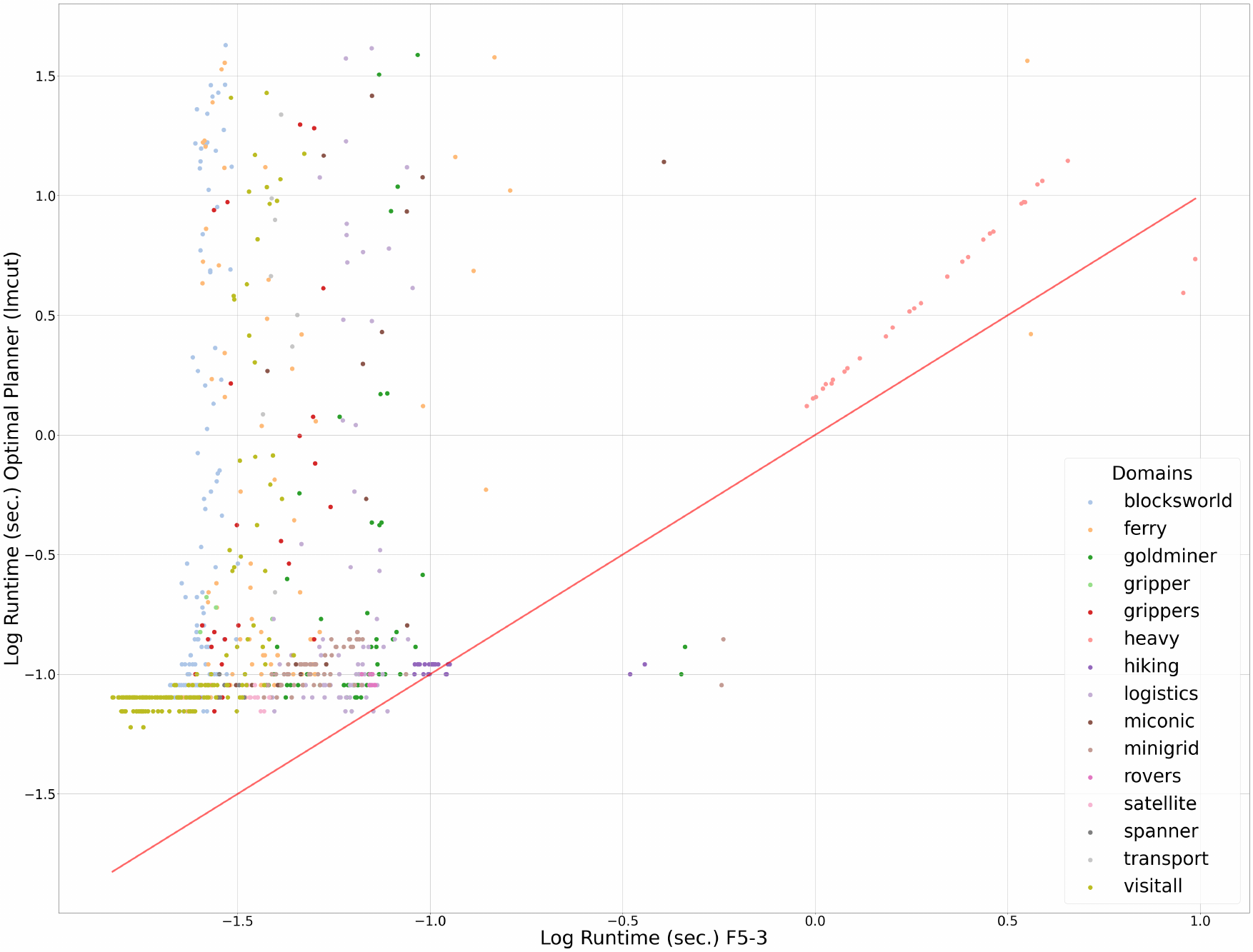}\label{fig:time_qwent_optimal}}
    \caption{\label{fig:time_plots}%
The runtime (log scale, seconds) of generating a plan using the best generalized plan (\exfullinit) and using the satisficing planner (\exFF,\ left column) or optimal planner (\exLMC, right column) for each commonly solved instance. Diagonal is plotted in red.  
}
\end{figure*}

We also provide the individual runtimes (log scale) of all commonly solved instances in Figure \ref{fig:time_gpt_gbfs}, \ref{fig:time_deepseek_gbfs}, \ref{fig:time_llama_gbfs} and \ref{fig:time_qwent_gbfs}. Tasks from different domains are plotted in different colors and the red diagonal visualizes the boundary between tasks for which \exfullinit is faster (above the diagonal) and for which \exFF is faster (below the diagonal). The comparisons indicate that there are some domains for which the Python programs generated by any model are faster than \exFF whereas e.g. for Heavy all four LLMs produce programs that have a similar runtime as \exFF.

\begin{figure*}
    \centering
    \subfloat[GPT \exfullinit vs. \exFF]
    {\includegraphics[width=0.358\textwidth]{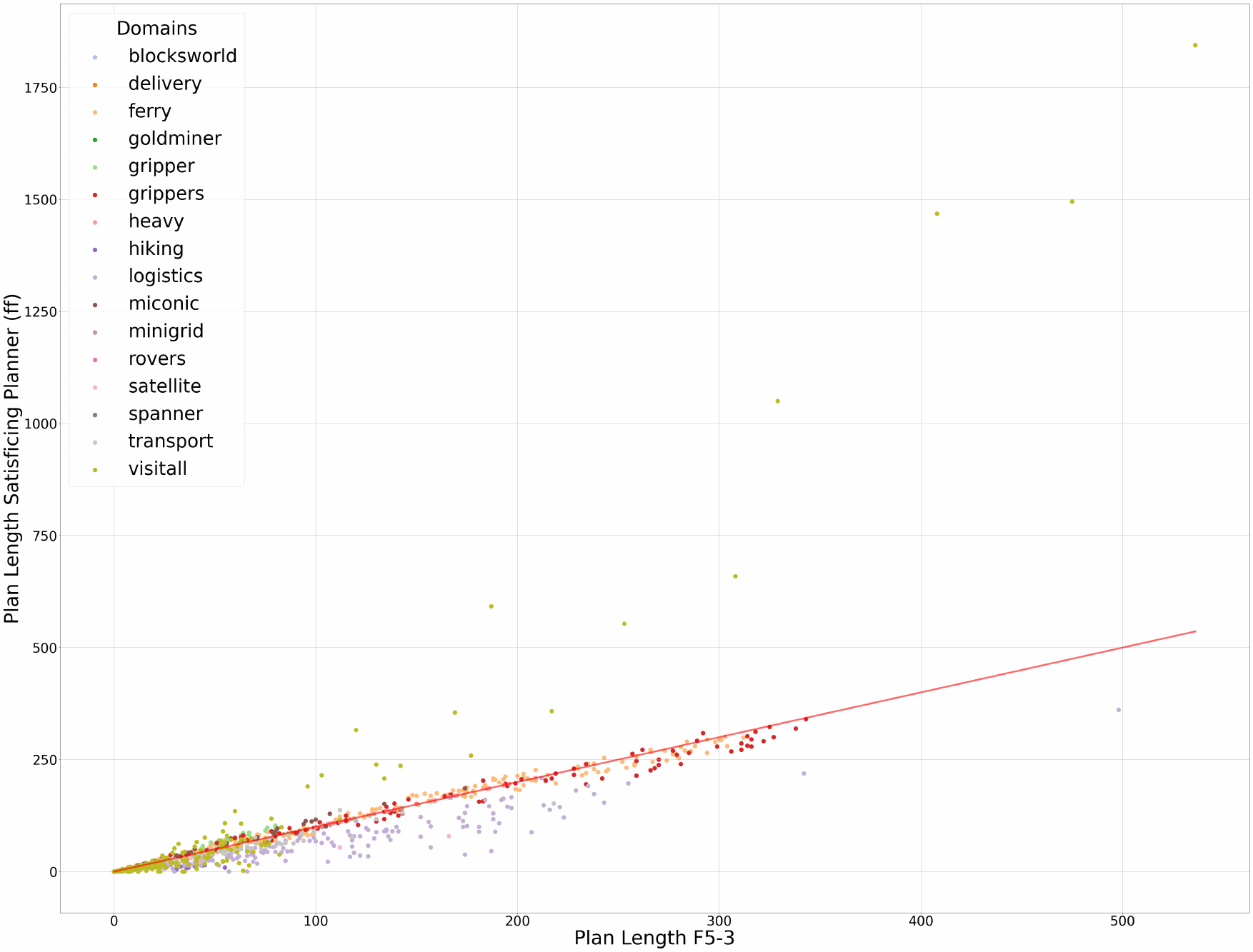}\label{fig:pl_gpt_gbfs}}
    \qquad
    \subfloat[GPT \exfullinit vs. \exLMC]
    {\includegraphics[width=0.358\textwidth]{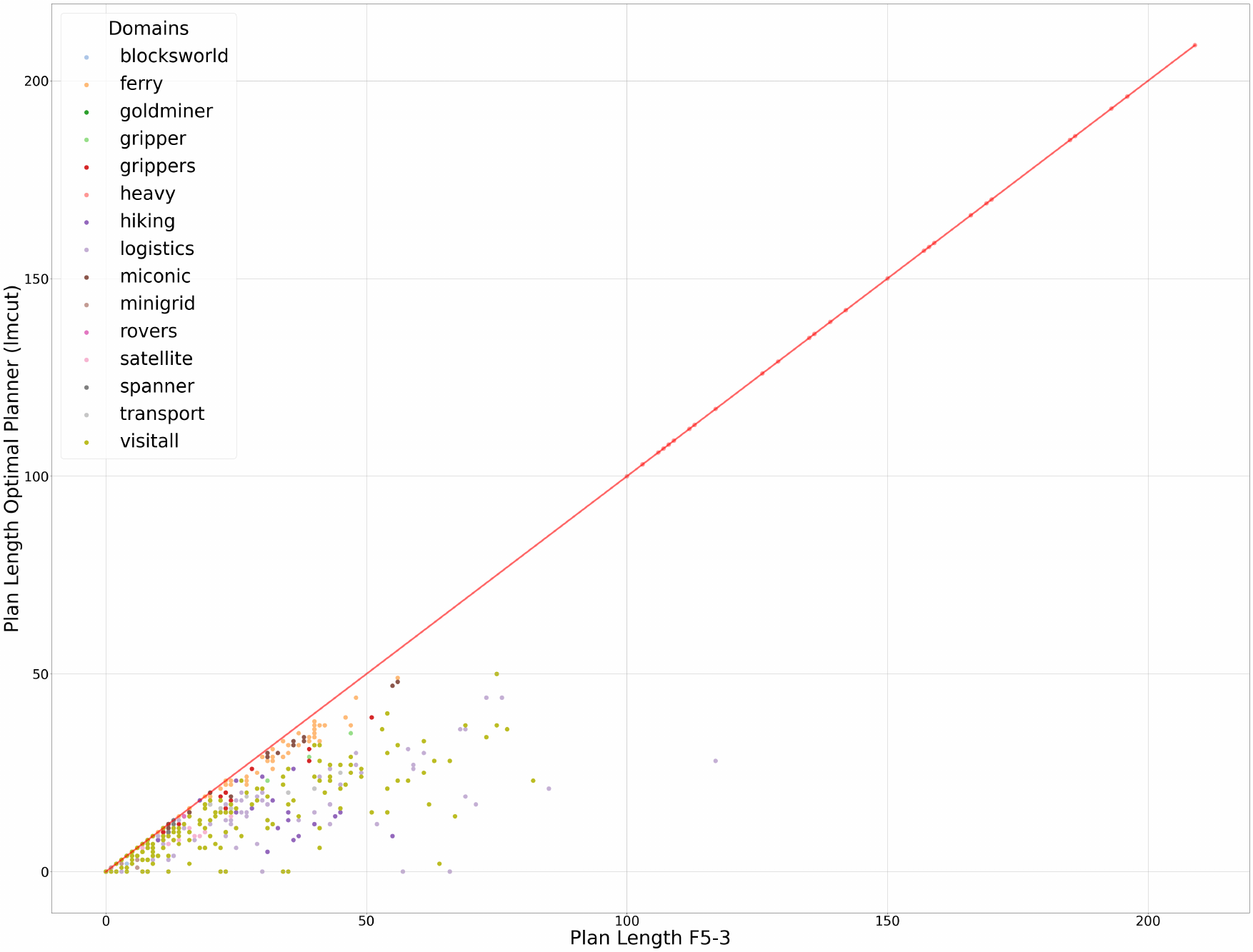}\label{fig:pl_gpt_optimal}}
    
    \subfloat[DeepSeek \exfullinit vs. \exFF]
    {\includegraphics[width=0.358\textwidth]{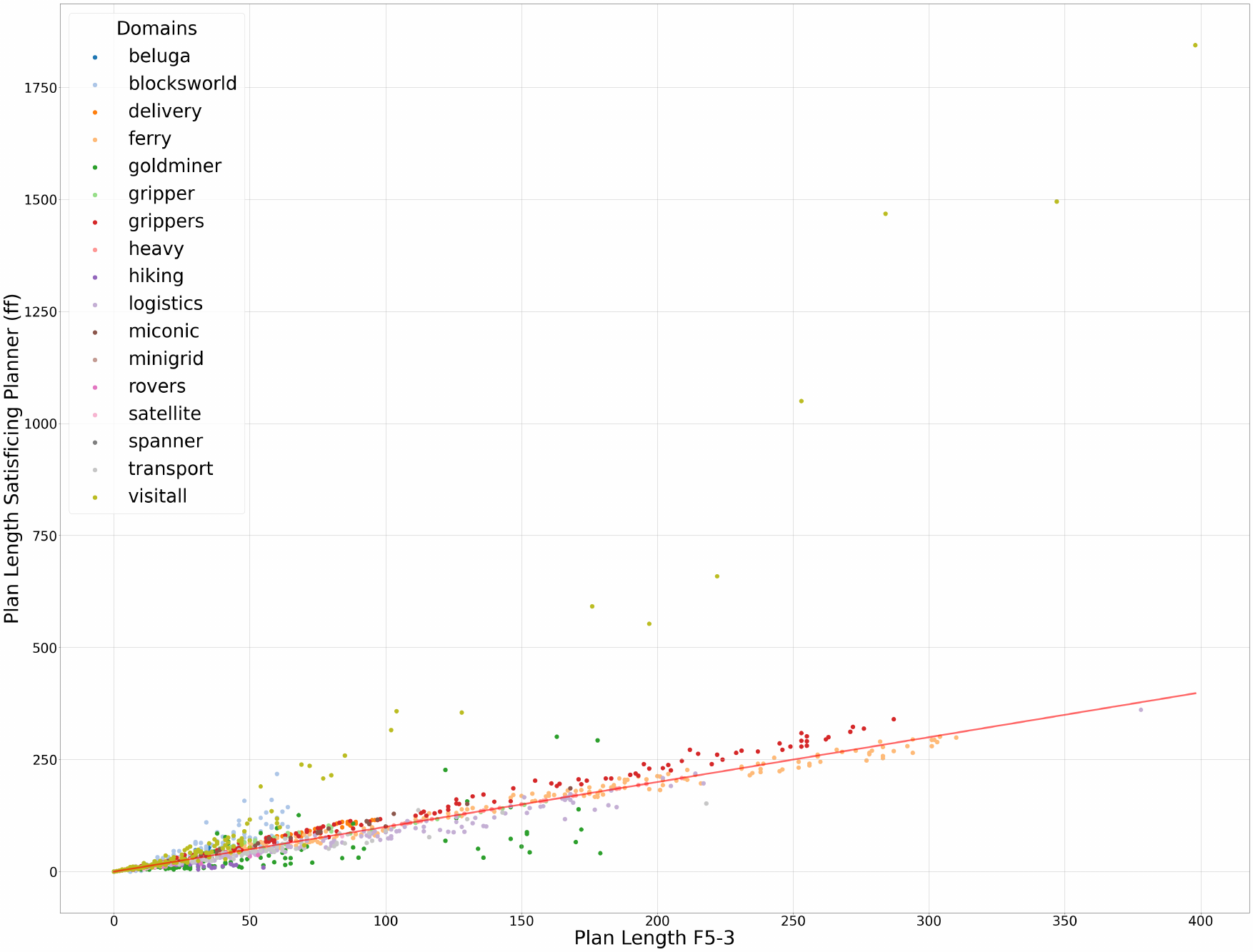}\label{fig:pl_deepseek_gbfs}}
    \qquad
    \subfloat[DeepSeek \exfullinit vs. \exLMC]
    {\includegraphics[width=0.358\textwidth]{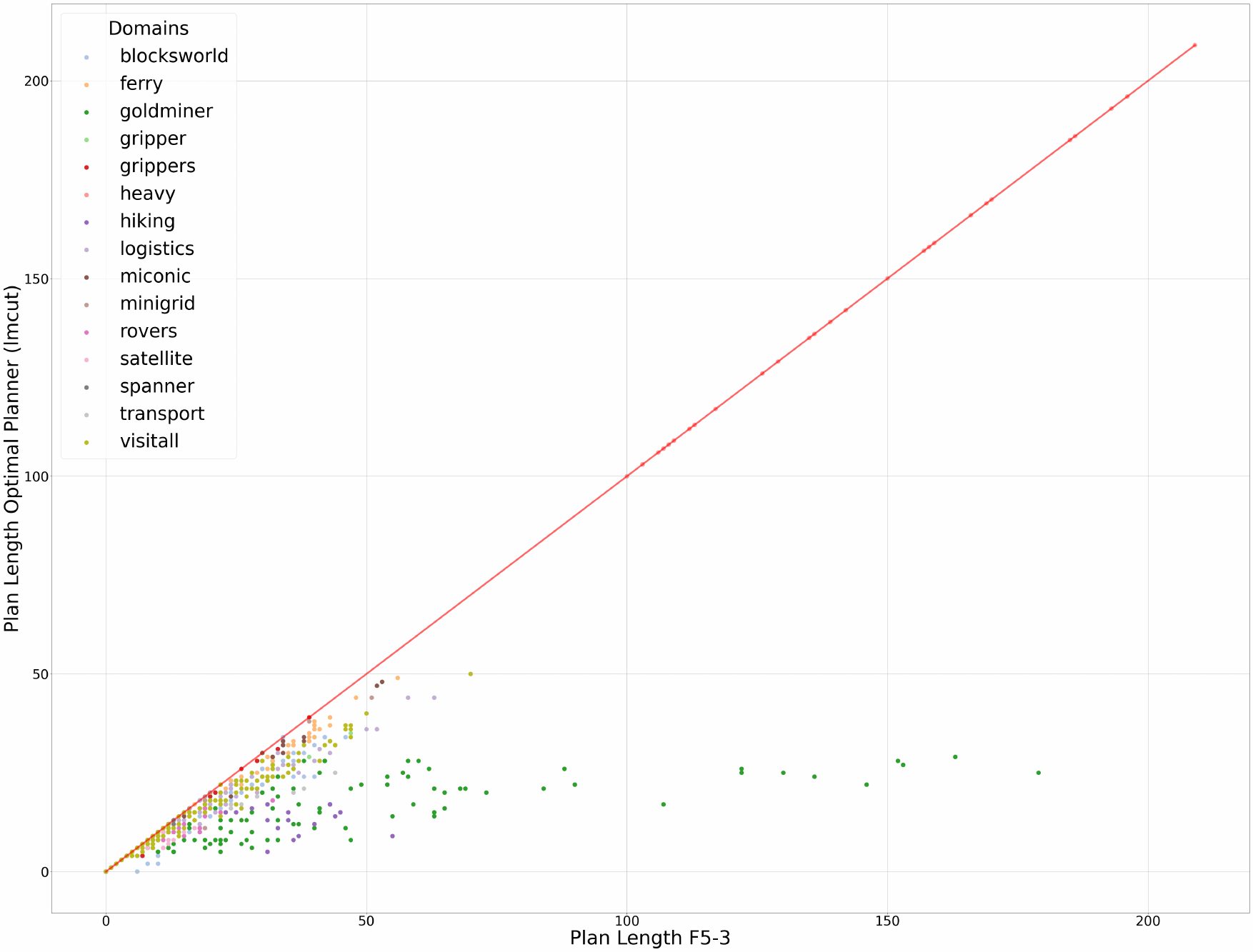}\label{fig:pl_deepseek_optimal}}
    
    \subfloat[Llama \exfullinit vs. \exFF]
    {\includegraphics[width=0.358\textwidth]{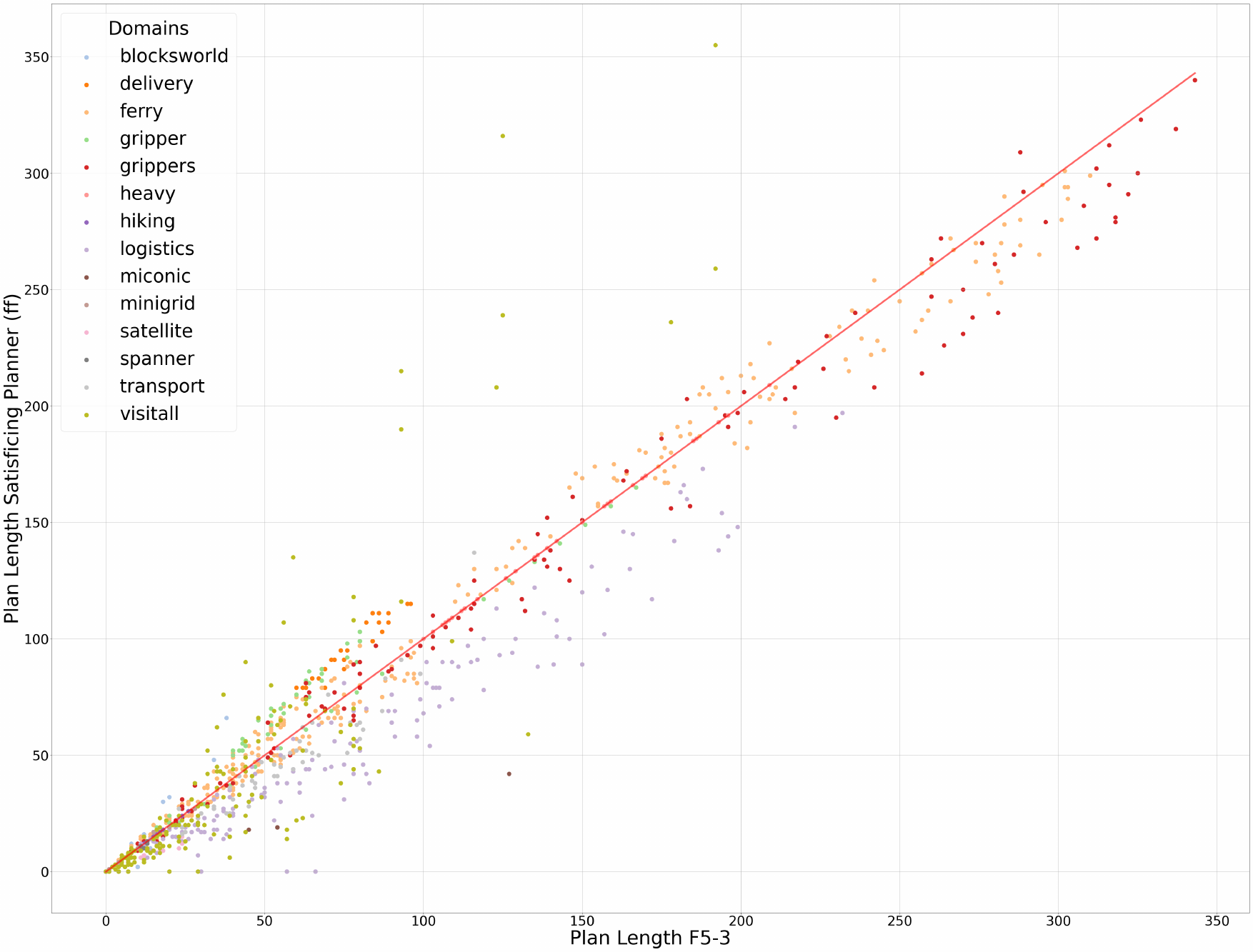}\label{fig:pl_llama_gbfs}}
    \qquad
    \subfloat[Llama \exfullinit vs. \exLMC]
    {\includegraphics[width=0.358\textwidth]{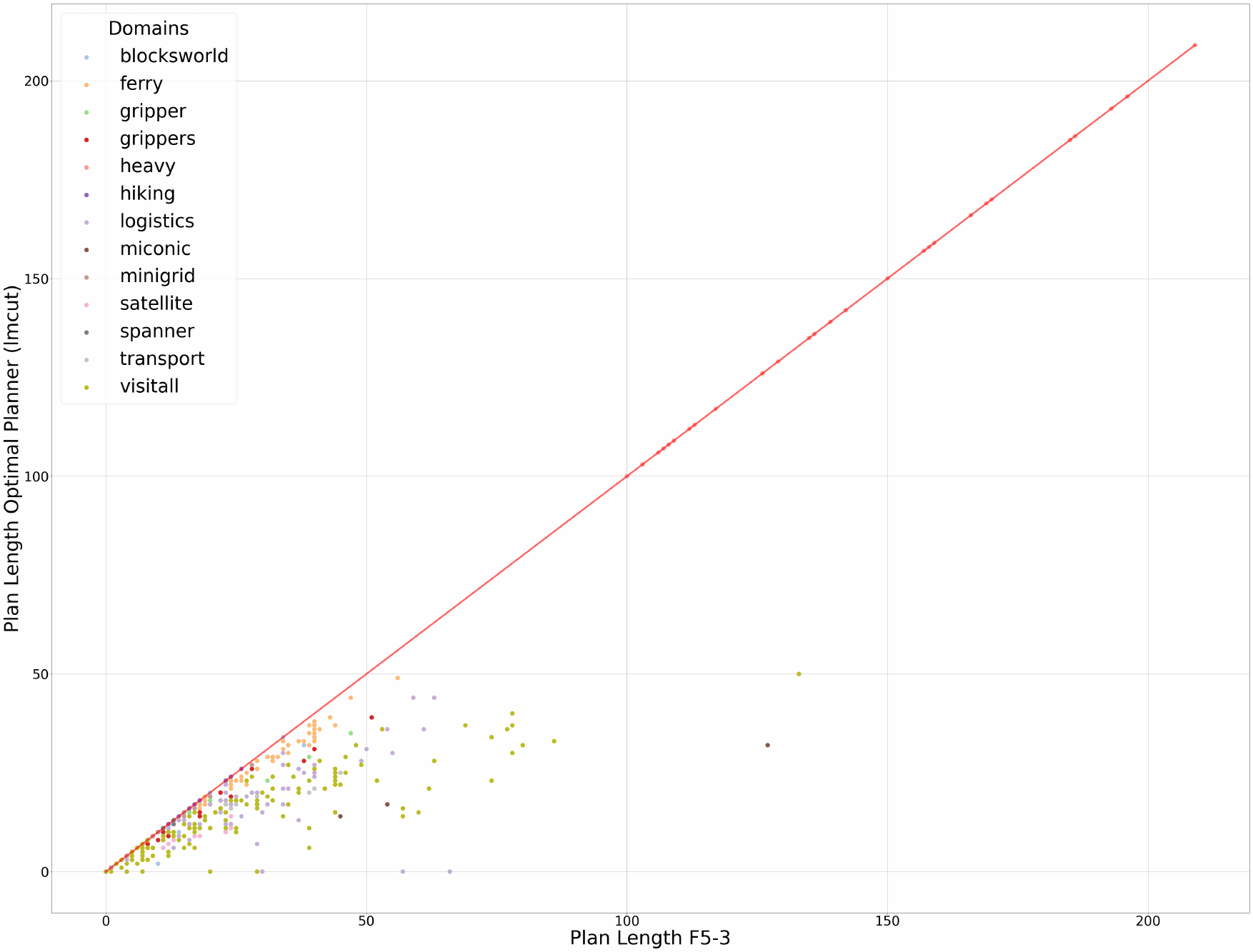}\label{fig:pl_llama_optimal}}
    
    \subfloat[Qwen \exfullinit vs. \exFF]
    {\includegraphics[width=0.358\textwidth]{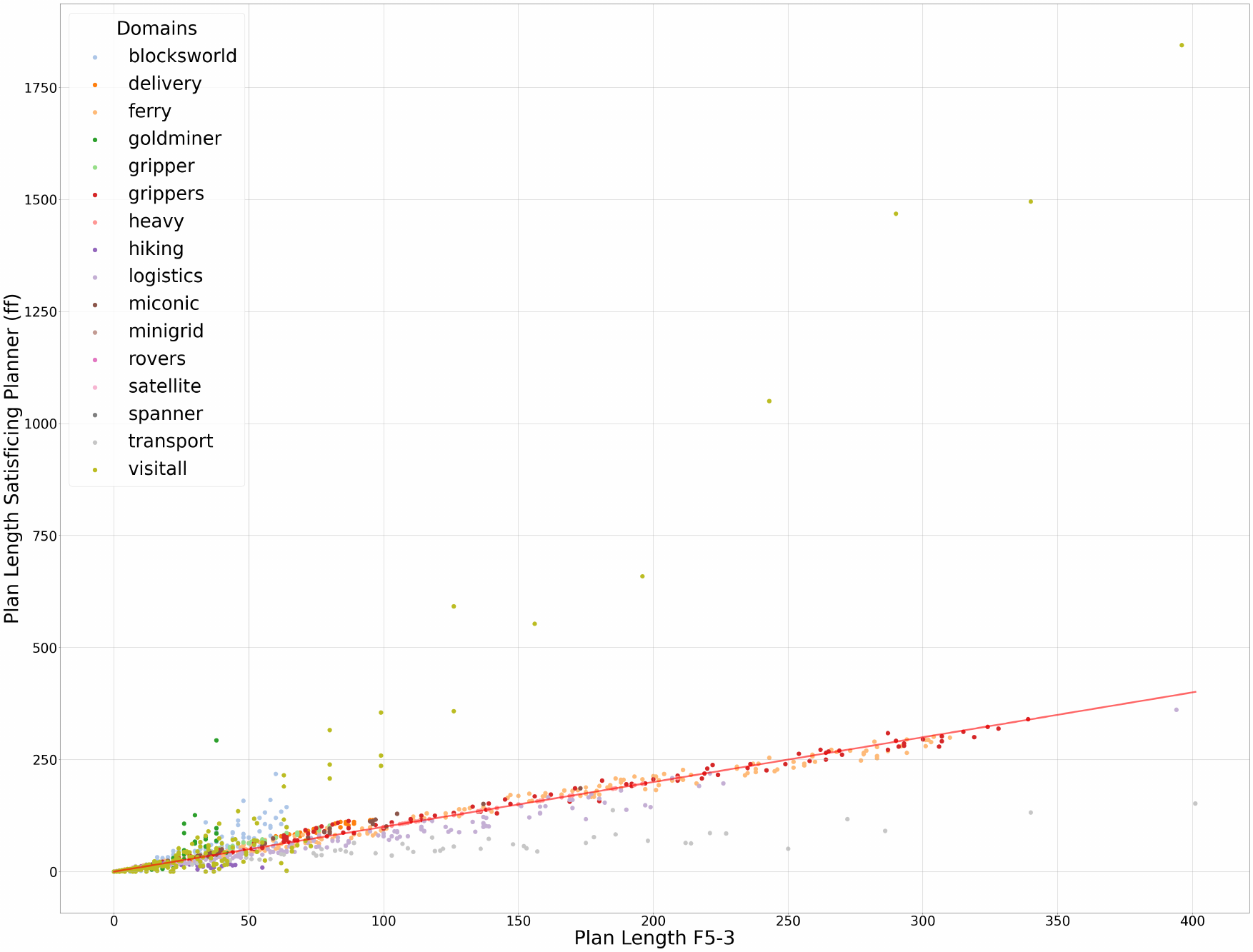}\label{fig:pl_qwent_gbfs}}
    \qquad
    \subfloat[Qwen \exfullinit vs. \exLMC]
    {\includegraphics[width=0.358\textwidth]{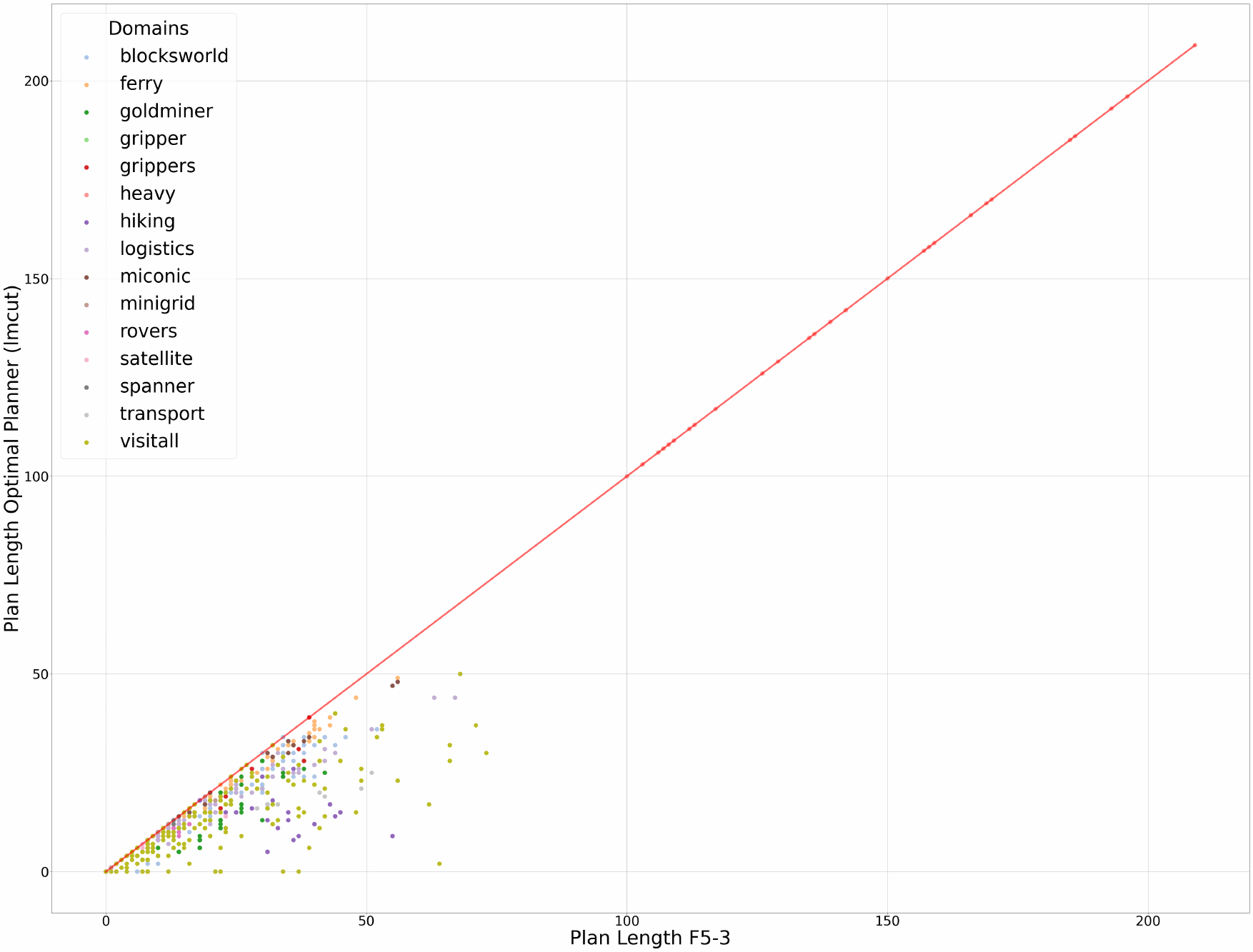}\label{fig:pl_qwent_optimal}}
     \caption{\label{fig:plan_lengths_plots}%
The lengths (log scale, seconds) of the plans generated using the best generalized plan (\exfullinit) and using the satisficing planner (\exFF,\ left column) or optimal planner (\exLMC, right column) for each commonly solved instance. Diagonal is plotted in red.  
}
\end{figure*}

Focusing on the plan length, we observe that \exFF generates plans that are on average between 1.1 (DeepSeek) and 1.57 (GPT) times shorter. Figures \ref{fig:pl_gpt_gbfs}, \ref{fig:pl_deepseek_gbfs}, \ref{fig:pl_llama_gbfs} and \ref{fig:pl_qwent_gbfs} show the plan lenghts for each commonly solved task. One noticeable outlier in all four plots are the tasks from Visitall where some of the plans generated by \exFF are considerably longer.

\begin{table}[h!]
\fontsize{8pt}{8pt}\selectfont
    \centering
    \begin{tabular}{|l|r|r|r|r|r|}
    \hline
        \multirow{2}{*}{Model} & \multicolumn{1}{c|}{\multirow{2}{*}{N}} & \multicolumn{2}{c|}{Runtime} & \multicolumn{2}{c|}{Plan length} \\
         &  & Ratio & \exfullinit $\leq$ & Ratio & \exfullinit $\leq$\\
        \hline\hline
        GPT & 1112 & 0.42 & 96$\%$ & 1.57 & 42$\%$  \\
        DeepSeek & 1442 &  0.36 & 97$\%$ & 1.10 & 61$\%$ \\
        Llama & 1043 &  0.89 & 88$\%$ &  1.34 & 49$\%$ \\
        Qwen & 1324 &  0.43 & 96$\%$ &  1.29 & 56$\%$ \\
        \hline
    \end{tabular}
    \caption{Comparison of the runtime and generated plan length for the best Python program generated using the \exfullinit (best run) configuration and \exFF on all commonly solved tasks (N: number of tasks).}
    \label{tab:runtimesff}
\end{table}

In our experiments, we do not expect the LLM to come up with generalized plans that generate optimal plans. In fact, we explicitly prompt the LLM that it is not required that the strategy is optimal. We still provide a comparison of the time to find a plan and the lengths of the plans when using \exfullinit and \exLMC to get a better idea of the quality of the plans. We apply the same analysis as for \exfullinit and \exFF. 

Table \ref{tab:runtimeslm} shows the results for commonly solved instances (best run of \exfullinit) averaged across domains. As optimal planning is slower than satisficing planning, using \exfullinit instead of the symbolic planner is even faster in direct comparison. When comparing the lengths of the plans generated by the best \exfullinit programs with optimal plan length, we observe that the programs generated by DeepSeek create plans closest to optimal plans. The programs generated by GPT create plans that are on average 2.19 times longer than the optimal plans. 

For DeepSeek, Llama and Qwen, the best programs generate optimal plans for more than 30\% of the commonly solved tasks. Looking at the differences for the individual domains, we observe that the tasks from the Heavy domain are always solved optimally (see Figure \ref{fig:pl_gpt_optimal}, \ref{fig:pl_deepseek_optimal}, \ref{fig:pl_llama_optimal}, \ref{fig:pl_qwent_optimal}). However, in the Heavy domain the optimal plans are the only valid plans and therefore each program that generates a correct plan for a task also generates an optimal plan.  

\begin{table}[h!]
\fontsize{8pt}{8pt}\selectfont
    \centering
    \begin{tabular}{|l|r|r|r|r|r|}
    \hline
        \multirow{2}{*}{Model} & \multicolumn{1}{c|}{\multirow{2}{*}{N}} & \multicolumn{2}{c|}{Runtime} & \multicolumn{2}{c|}{Plan length} \\
         &  & Ratio & \exfullinit $\leq$ & Ratio & \exfullinit $=$\\
        \hline\hline
        GPT & 536 & 0.43 & 97$\%$ & 2.19 & 28$\%$\\
        DeepSeek & 785 & 0.33 & 98$\%$ & 1.43 & 34$\%$\\
        Llama & 512 & 0.37 & 98$\%$& 1.77 &  39$\%$\\
        Qwen & 724 & 0.36 & 98$\%$ & 1.65 & 32$\%$\\
        \hline
    \end{tabular}
    \caption{Comparison of the runtime and generated plan length for the best Python program generated using the \exfullinit (best run) configuration and \exLMC on commonly solved tasks (N: number of tasks).}
    \label{tab:runtimeslm}
\end{table}

\section{Computational Budget}

\begin{table*}[t]
    \centering
    \begin{tabular}{|l|r|r|r|r|r|r|r|r|}
    \hline
        Approach & \multicolumn{1}{c|}{NL}  & \multicolumn{2}{c|}{Strategies} & \multicolumn{2}{c|}{Programs} & \multicolumn{2}{c|}{LLM calls}  \\
                 & \multicolumn{1}{c|}{descriptions}  & Min & Max  & Min & Max & Min & Max \\   
        \hline\hline 
        \exsilver & Domain + 2 tasks & 1 & 1 & 1 & 7 & 3 & 9\\
        \exbaseline & Domain + 6 tasks & 1 & 1 & 1 & 7 & 15 & 21 \\
        \exfullinit & Domain + 6 tasks & 1 & 6 & 1  &  20 & 17 & 91\\
        \hline
    \end{tabular}
    \caption{Comparison of the number of NL descriptions generated by the two baseline pipelines and our best variants and of the minimum and maximum number of strategies, programs and LLM calls possible with the configurations we used for our experiments}
    \label{tab:max_min}
\end{table*}

Our proposed pipeline enables the generation of better generalized plans across LLM models and domains. At the same time, extending the pipeline by additional parts also increases the computational cost. Our framework consists of more steps than the baselines because more strategies and programs can be generated and because of the new strategy validation part. Table \ref{tab:max_min} provides an overview over the minimum and maximum number of LLM calls, strategies and programs that could be generated using the two baselines and our \exfullinit approach with the parameters used in our experiments.

If the first strategy and program already result in full coverage of the validation tasks, then also \exfullinit only generates a single strategy and a single program. However, the strategy validation step requires one LLM call to get the NL description of each validation task and one LLM call per task during the strategy validation itself. Therefore, also the minimum number of LLM calls is higher than for the baselines. The individual LLM calls in our framework are very different, e.g. generating an NL description of a task is less costly than generating a pseudocode with respect to the number of tokens (and hence generation time). Therefore, the number of tokens gives a better idea of the cost than the number of calls alone. 

We provide the details about the actual number of LLM calls, processed tokens and the generation time from our experiments below. For all three resources, our approach has a higher cost than the baselines. However, all of these costs arise per domain and not per task. Once a correct program is obtained, it can be used on an arbitrary number of tasks from that domain with the only additional cost being the runtime of the program which is very small - as shown above. 

\begin{figure*}[ht]
    \subfloat[LLM calls GPT]
    {\includegraphics[width=0.43\textwidth]{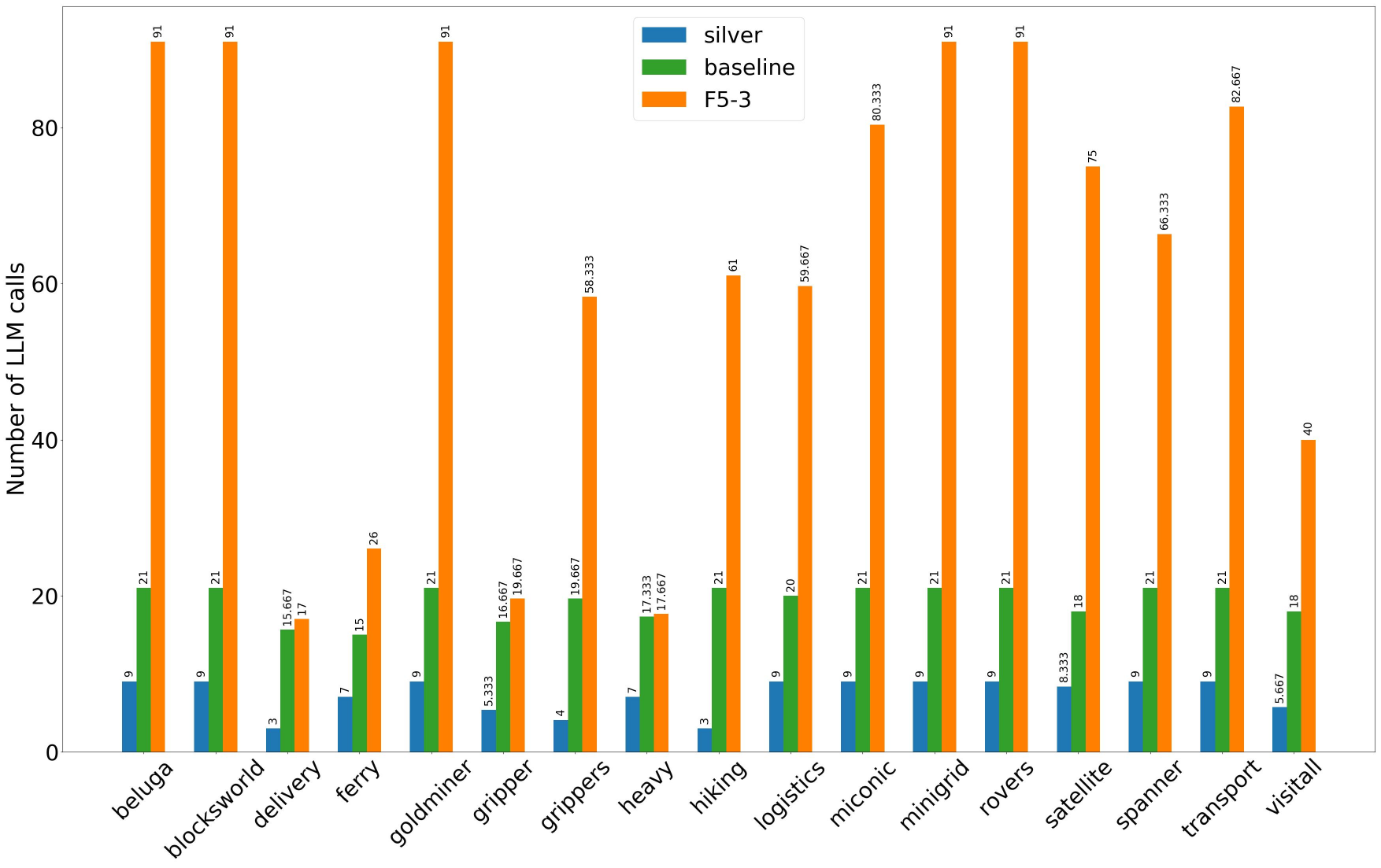}\label{fig:inf_calls_gpt}}
    \qquad
    \subfloat[Tokens GPT]
    {\includegraphics[width=0.43\textwidth]{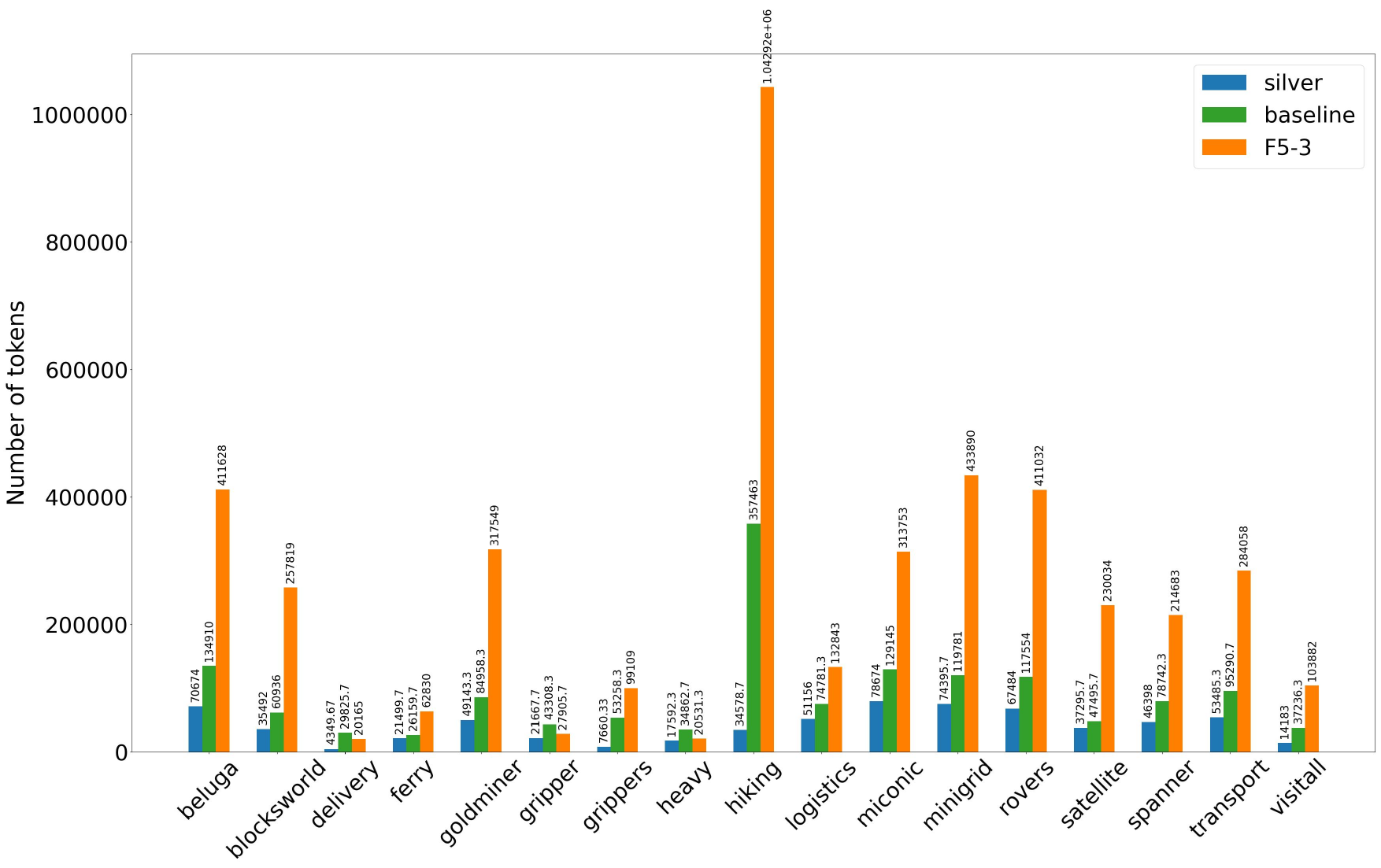}\label{fig:inf_token_gpt}}

    \subfloat[LLM calls DeepSeek]
    {\includegraphics[width=0.43\textwidth]{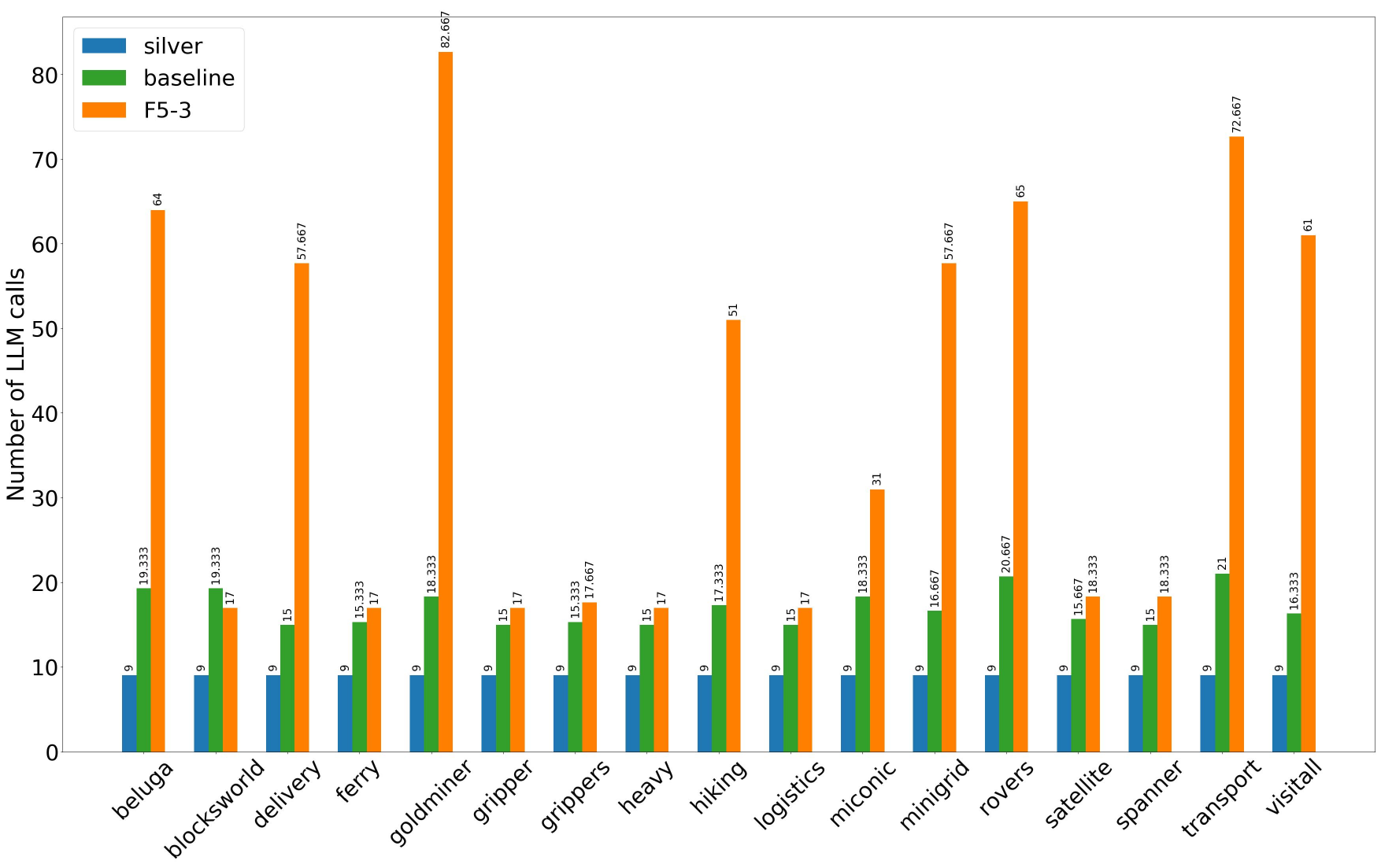}\label{fig:inf_calls_deepseek}}
    \qquad
    \subfloat[Tokens DeepSeek]
    {\includegraphics[width=0.43\textwidth]{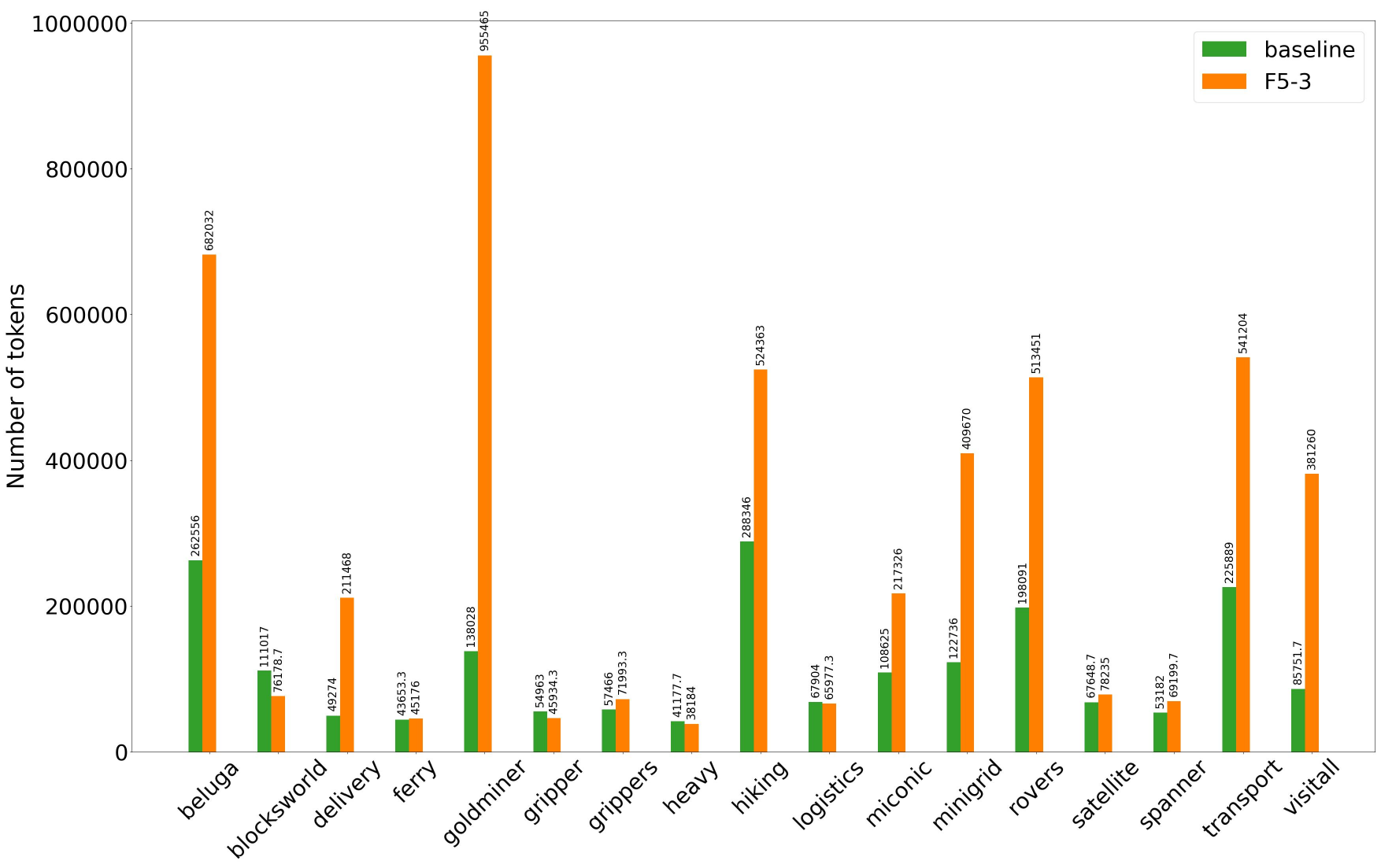}\label{fig:inf_token_deepseek}}

    \subfloat[LLM calls Llama]
    {\includegraphics[width=0.43\textwidth]{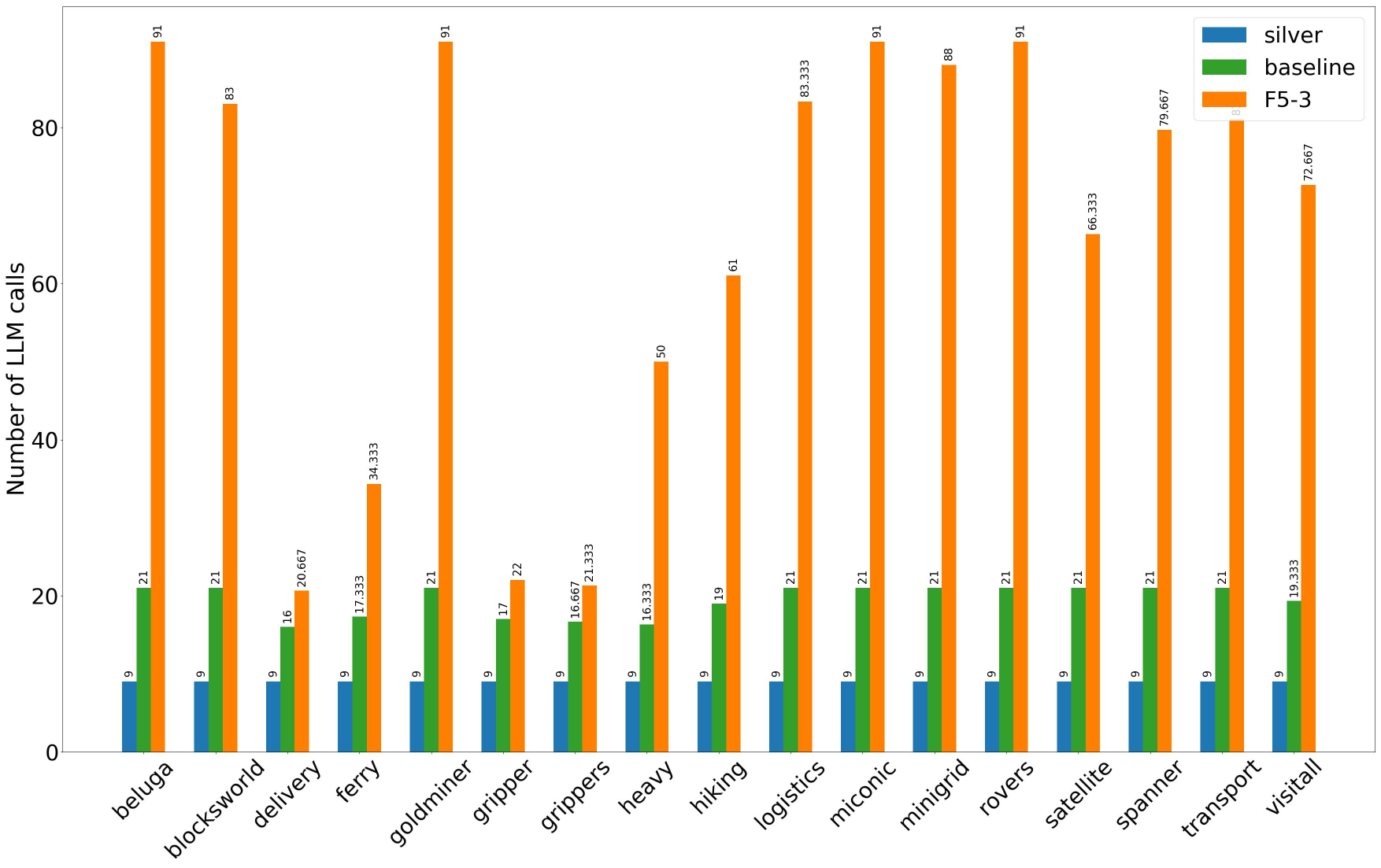}\label{fig:inf_calls_llama}}
    \qquad
    \subfloat[Tokens Llama]
    {\includegraphics[width=0.43\textwidth]{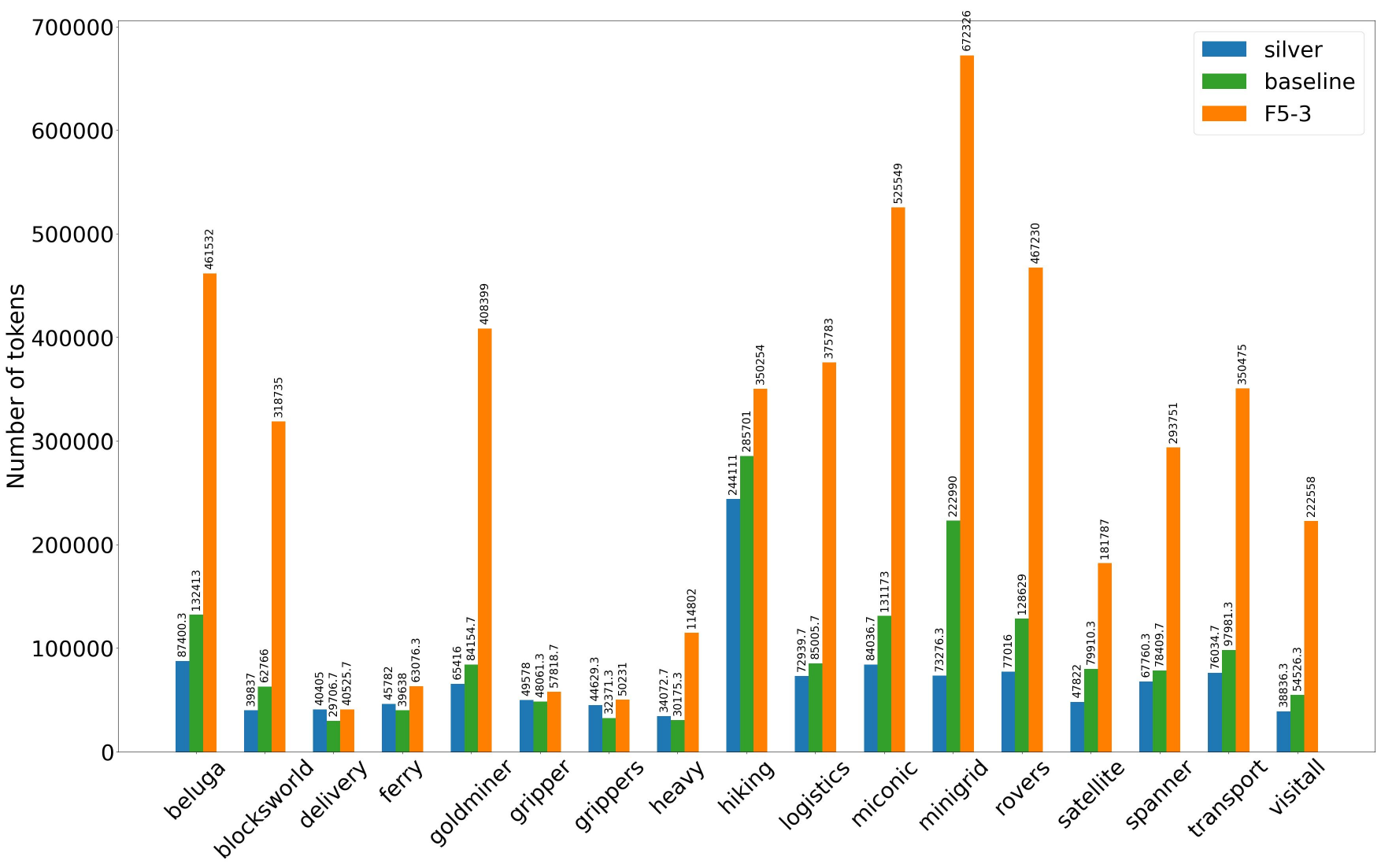}\label{fig:inf_token_llama}}

    \subfloat[LLM calls Qwen]
    {\includegraphics[width=0.43\textwidth]{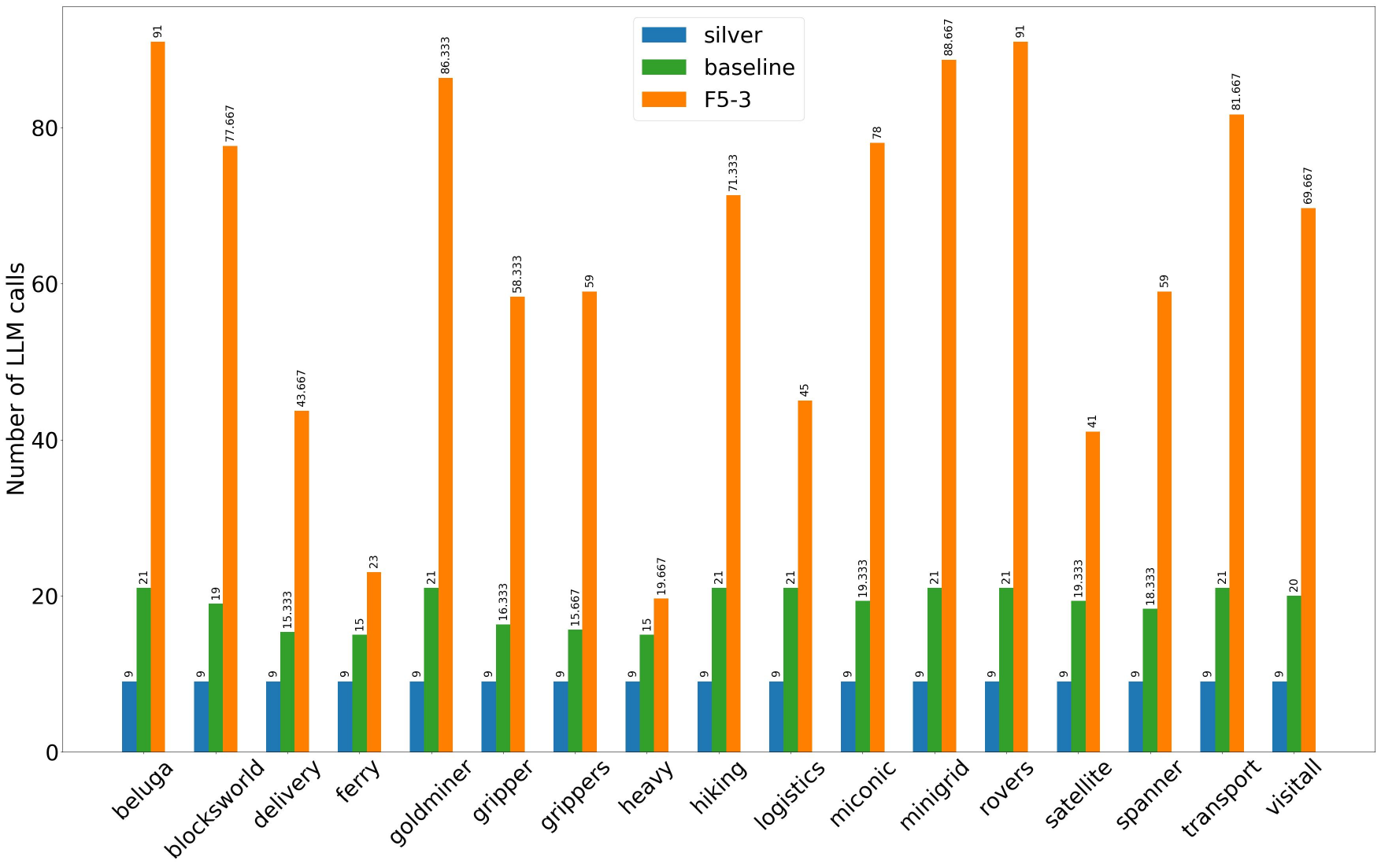}\label{fig:inf_calls_qwen}}
    \qquad
    \subfloat[Tokens Qwen]
    {\includegraphics[width=0.43\textwidth]{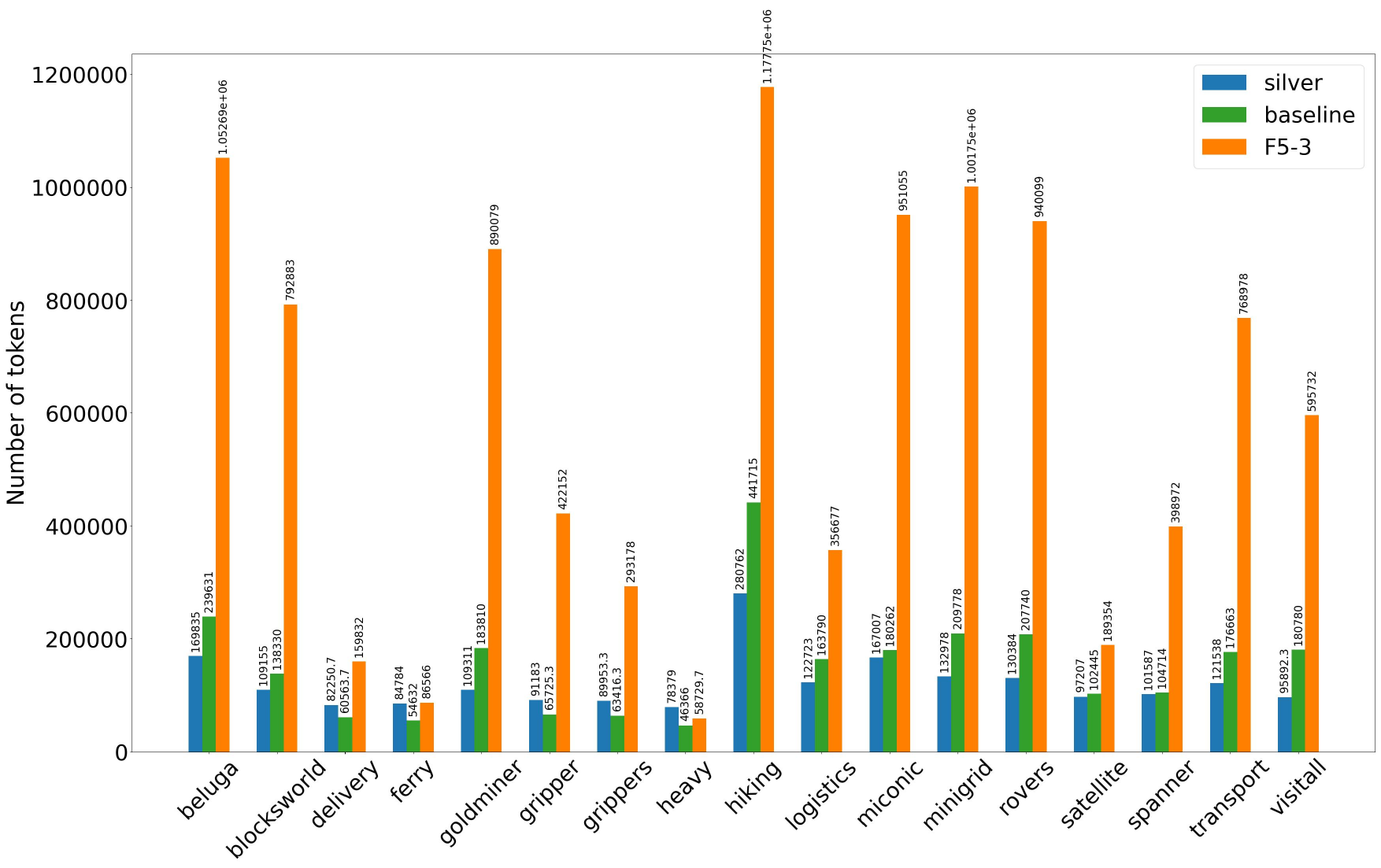}\label{fig:inf_token_qwen}}
    \caption{\label{fig:inf_tokens_plots}%
    The number of LLM calls (left column) and number of processed tokens (input + output tokens; right column) for generating the Python program for each domain, averaged over three runs.}
\end{figure*}

\paragraph{LLM Calls and Tokens.}  Table \ref{tab:inf_averages} shows number of actual LLM calls and processed tokens (input tokens + output tokens, including reasoning tokens) for the two baselines and our \exfullinit approach per model\footnote{The number of tokens for \exsilver with DeepSeek is missing in Table \ref{tab:inf_averages} and Figure \ref{fig:inf_token_deepseek} because the reasoning tokens were not tracked in our first version of the original code from \exsilver}. 
The numbers of calls and tokens were obtained by averaging over the three runs per domain and then over the domain averages. We also provide the relative difference in computational budget between \exfullinit and each of the two baselines. 


Figure \ref{fig:inf_tokens_plots} shows the number of LLM calls and the number of tokens processed for the individual domains, averaged over the three runs.
There are large differences between the domains with respect to the amount of LLM calls and tokens processed to generate the Python programs in our experiments. For some domains the number of tokens processed by \exfullinit is comparable to the baselines whereas for others we observe a larger difference. While there are also differences between the models, some domains tend to require less tokens in general than others. For example, Ferry, Heavy and Satellite are always among the domains with the least processed tokens whereas Beluga, Goldminer, Hiking, Minigrid and Rovers tend to require more tokens.

\paragraph{Generation Time.} Another important factor is the time it takes to actually obtain a generalized plan. The last column of Table \ref{tab:inf_averages} provides the generation time averaged over runs and then over the domains. However, the generation times for GPT and DeepSeek only give a rough idea of the required generation time but not an actual estimate. First, we use the respective APIs to access the two models which can have varying response times. Furthermore, we use caching when running GPT and DeepSeek, and retrieving a response from the cache is faster than generating it. As some of the tested approaches share the inputs up until some point in the pipeline, this leads to different measured generation times than one would obtain when running the pipeline without caching.

We measure the generation time from starting the pipeline until selecting the best program, i.e. the measured time includes the time of running each generated program on each validation task. Therefore, the total generation time also depends on how efficient the actual program implementation is and whether there are programs with infinite loops.

\begin{figure*}[ht]
    \subfloat[GPT]
    {\includegraphics[width=0.48\textwidth]{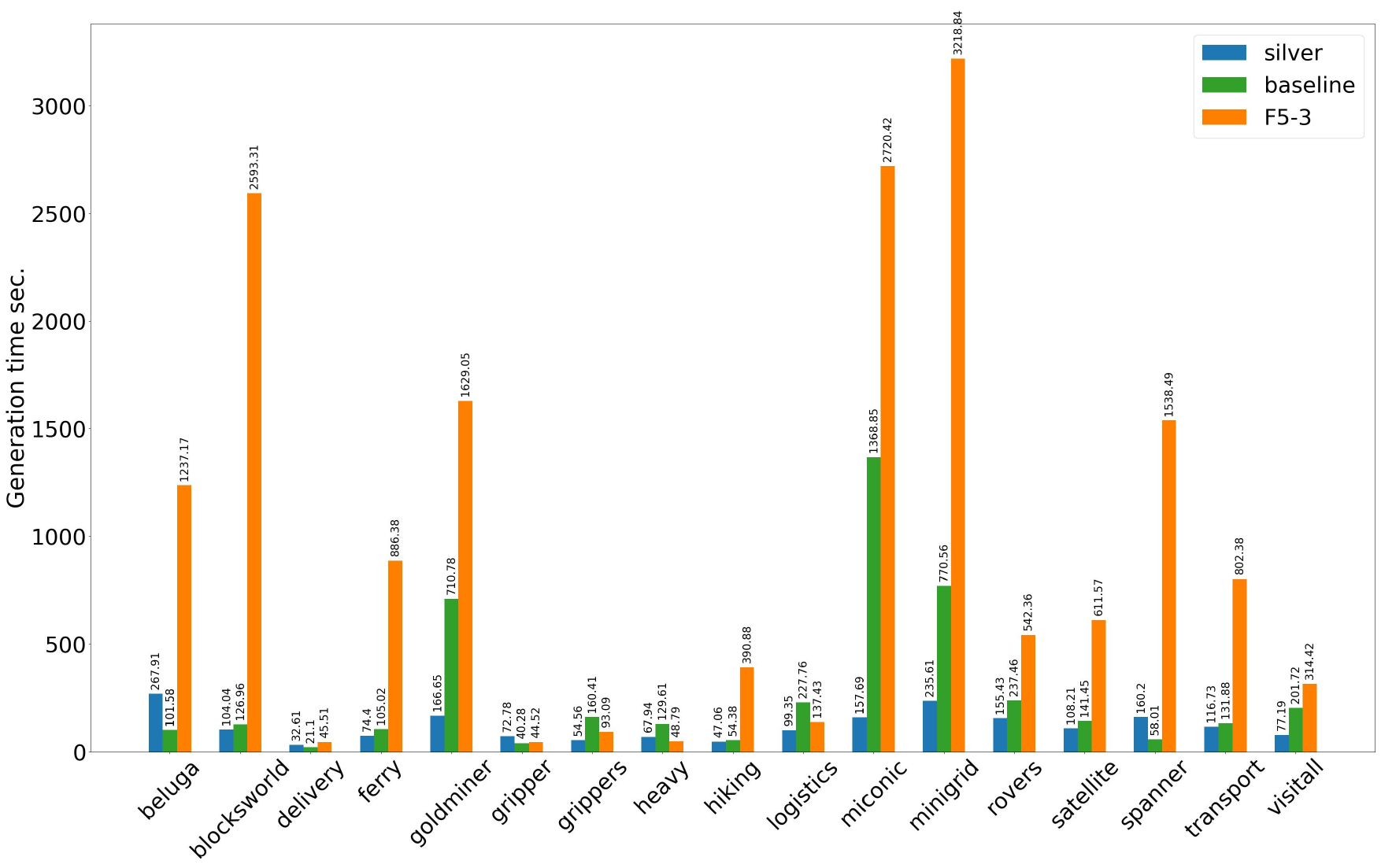}\label{fig:inf_time_gpt}}
    \qquad
    \subfloat[DeepSeek]
    {\includegraphics[width=0.48\textwidth]{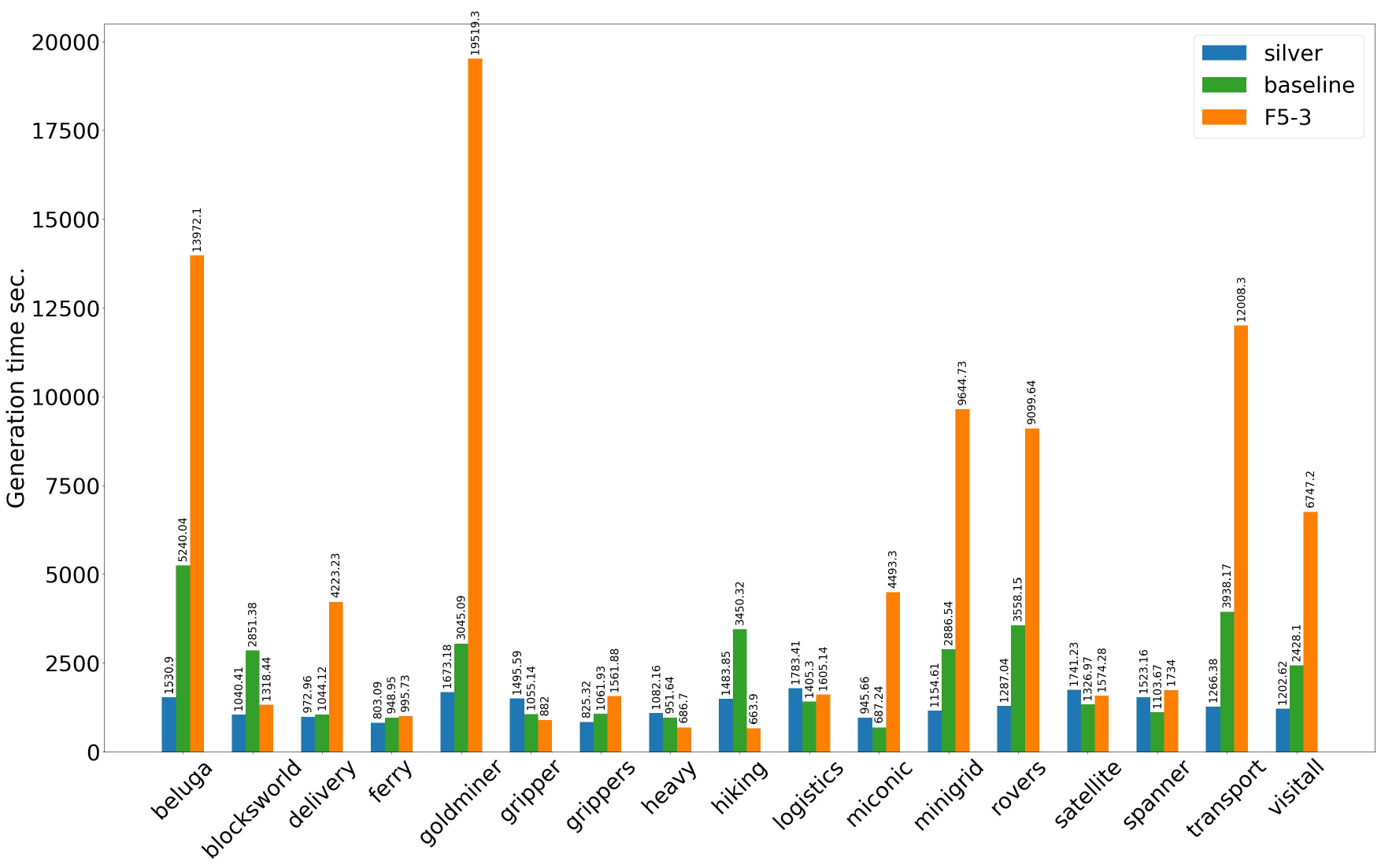}\label{fig:inf_time_deepseek}}

    \subfloat[Llama]
    {\includegraphics[width=0.48\textwidth]{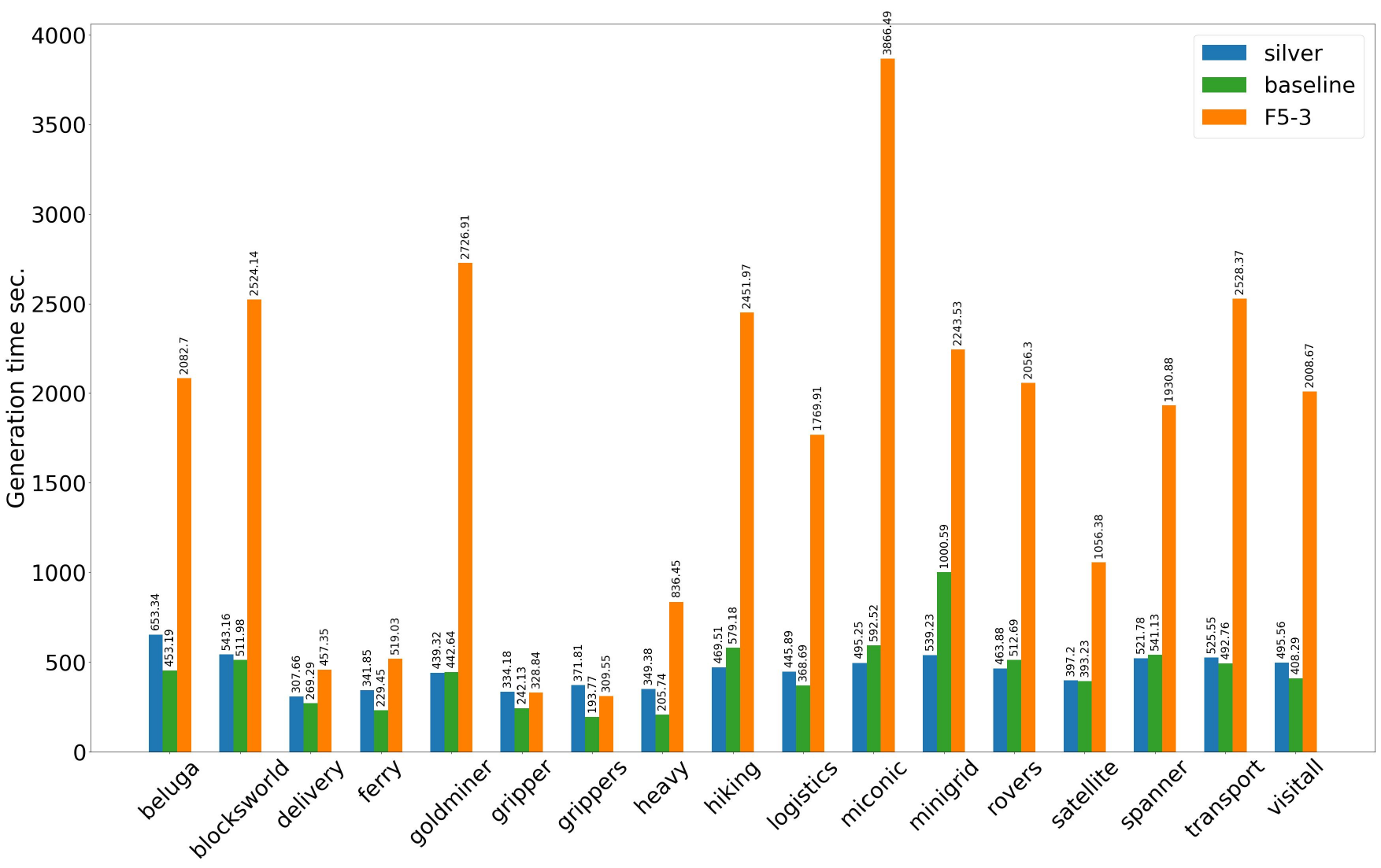}\label{fig:inf_time_llama}}
    \qquad
    \subfloat[Qwen]
    {\includegraphics[width=0.48\textwidth]{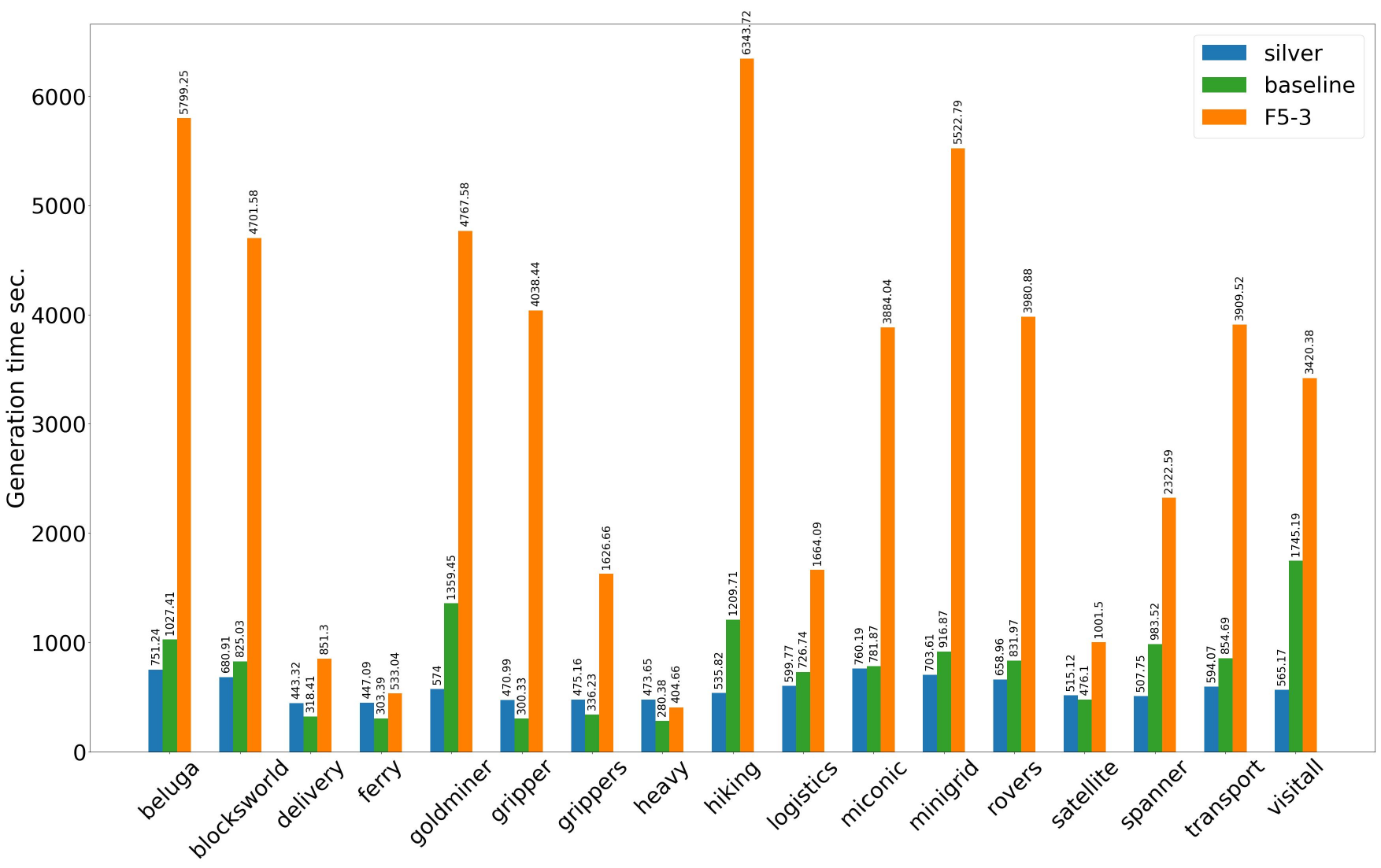}\label{fig:inf_time_qwen}}
     \caption{\label{fig:inf_time_plots}%
    The time required for generating the Python program for each domain, averaged over three runs. Time is measured from starting the pipeline until selection of the final best program is completed, i.e. includes running each generated program on the validation tasks.}
\end{figure*}

Figure \ref{fig:inf_time_plots} shows the generation time per domain, averaged over the three runs. Similar to the number of LLM calls and tokens, we observe large differences between the individual domains. For example, the generation time for \exfullinit with Qwen was below 30 minutes for 6 of the domains, including Delivery, Grippers and Satellite. The generation of programs for Hiking took the longest on average, namely 1.75 hours. Focusing on the longest run for Hiking, we observed that the long runtime was caused by the program validation because the majority of the 20 generated programs did not terminate but was stopped after 45 seconds. The generation time could hence be reduced considerably by lowering the runtime limit.

\begin{table}[ht]
    \centering
    \begin{tabular}{|l|l|r|r|r|}
    \hline
        Model & Approach & Calls & Tokens & Time s. \\
        \hline\hline
        \multirow{5}{*}{GPT} & \exsilver & 7.3 &  40337.0& 117.6\\
         & \exbaseline & 19.4 & 89747.5 & 269.9\\
         & \exfullinit & 62.3 & 257919.6 & 991.5\\
         \cdashline{2-5}
         & \exfullinit\ / \exsilver & 8.5 & 6.4 & 8.4\\
         & \exfullinit\ / \exbaseline & 3.2 & 2.9 & 3.7\\
         \hline\hline
         \multirow{5}{*}{DeepSeek} & \exsilver & 9.0 & - & 1283.0\\
         & \exbaseline & 17.0 & 116253.5 & 2175.5\\
         & \exfullinit & 40.1 & 289830.4 & 5337.1\\
         \cdashline{2-5}
         & \exfullinit\ / \exsilver & 4.5 & - & 4.2\\
         & \exfullinit\ / \exbaseline & 2.4  & 2.5 & 2.5 \\
         \hline\hline
         \multirow{5}{*}{Llama} & \exsilver & 9.0 & 69938.5 & 452.6 \\
         & \exbaseline & 19.5 & 95506.6 & 437.5\\
         & \exfullinit & 66.3 & 291460.7 & 1746.9\\
         \cdashline{2-5}
         & \exfullinit\ / \exsilver & 7.4 & 4.2 & 3.9\\
         & \exfullinit\ / \exbaseline & 3.4 & 3.1 & 4.0 \\
         \hline\hline
         \multirow{5}{*}{Qwen} & \exsilver & 9.0 & 121466.5 & 573.9 \\
         & \exbaseline & 18.8 & 154139.0 & 781.0\\
         & \exfullinit & 63.8 & 596263.3 & 3221.9\\
         \cdashline{2-5}
         & \exfullinit\ / \exsilver & 7.1 & 4.9 & 5.6 \\
         & \exfullinit\ / \exbaseline & 3.4 & 3.9 & 4.1 \\
         \hline
    \end{tabular}
    \caption{The average numbers of LLM calls and processed tokens and the average generation time (in seconds) per run for the baselines and our framework. The shown results are averaged over the three seeds and then averaged across domains. We also show the relative increase in computational budget when using \exfullinit compared to \exsilver (\exfullinit\ / \exsilver) and compared to \exbaseline (\exfullinit\ / \exbaseline).}
    \label{tab:inf_averages}
\end{table}

\section{Datasets}\label{app:data}

\paragraph{Sources.}For our experiments, we focus on domains that have previously been used in research on LLMs in the context of classical planning. In particular, we use the domains from \citet{Silver_2024}'s generalized planning experiments and \citet{stein-25-icaps}'s LLM action-choice experiments. Table \ref{tab:data_sources} shows for each of the domains the sources of the tasks we include in our experiments. All tasks included in our experiments are solvable.

\paragraph{Instance generators.} The right column of Table \ref{tab:data_sources} shows the origin of the instance generators that we used to generate additional tasks for some of the domains, and that we used for the manual evaluation of the generalization power of our generalized plans. For the domains from \citet{Silver_2024}, we mostly focused on their generators\footnote{https://github.com/tomsilver/llm-genplan} and for all other domains, except Minigrid, we used the generators from the PDDL-Generators repository\footnote{https://github.com/AI-Planning/pddl-generators} \cite{seipp-et-al-zenodo2022}. 

For Blocksworld, we renamed two of the predicates in the newly generated problem files to match the domain file used by \citet{stein-25-icaps} (``on-table'' to ``ontable'' and ``arm-empty'' to ``handempty'').
Additionally, we modified all domains and tasks with non-uniform action costs or functions.

\renewcommand{\arraystretch}{1.25}
\begin{table}[ht]
    \fontsize{8pt}{8pt}\selectfont
    \centering
    \setlength{\tabcolsep}{1.5pt}
    \begin{tabular}{|l|l|l|}
    \hline
         \textbf{Domain} & \textbf{Source of tasks} & \textbf{Instance Generators} \\
         \hline\hline
         delivery & \citet{Silver_2024} & \citet{Silver_2024} \\
         \hline
         \multirow{3}{*}{ferry} & \citet{Silver_2024} & \citet{Silver_2024} \\
         \arrayrulecolor{gray}\cline{2-3}
         \arrayrulecolor{black}
         & \citet{stein-25-icaps} & \multirow{2}{*}{\citet{seipp-et-al-zenodo2022}}\\
         & Generated by us & \\
         \hline
         \multirow{2}{*}{gripper} & \citet{Silver_2024} &  \citet{Silver_2024} \\
         \arrayrulecolor{gray}\cline{2-3}
         \arrayrulecolor{black}
         & IPC gripper98 & \citet{seipp-et-al-zenodo2022}\\
         \hline
         heavy & \citet{Silver_2024} & \citet{Silver_2024} \\
         \hline
         hiking & \citet{Silver_2024} & \citet{Silver_2024} \\
         \hline
         miconic & \citet{Silver_2024} & \citet{Silver_2024}\\
         \hline
         spanner & \citet{Silver_2024} & \citet{Silver_2024} \\
         \hline\hline
         beluga & \citet{Eisenhut2024} & \citet{Eisenhut2024} \\ 
         \hline
        \multirow{3}{*}{blocks.} & \citet{stein-25-icaps} & \multirow{3}{*}{\citet{seipp-et-al-zenodo2022}}\\
         & IPC blocks00 &  \\
         & Generated by us & \\
         \hline
         goldminer & \citet{stein-25-icaps} & \citet{seipp-et-al-zenodo2022}\\
         \hline    
         \multirow{2}{*}{grippers} & \citet{stein-25-icaps} & \multirow{2}{*}{\citet{seipp-et-al-zenodo2022}}\\
         & Generated by us & \\
         \hline 
         \multirow{3}{*}{logistics} & \citet{stein-25-icaps} & \multirow{3}{*}{\citet{seipp-et-al-zenodo2022}}\\
          & IPC logistics98& \\
           & IPC  logistics00& \\
           \hline
         \multirow{2}{*}{minigrid} & https://github.com/ & \multirow{2}{*}{https://github.com/bonetblai/mini-grid/} \\
         & bonetblai/mini-grid/ & \\
         \hline
         rovers & \citet{stein-25-icaps} & \citet{seipp-et-al-zenodo2022} \\
         & Generated by us & \\
         \hline
         satellite & \citet{stein-25-icaps} &  \multirow{2}{*}{\citet{seipp-et-al-zenodo2022}}\\
         & Generated by us & \\
         \hline
         \multirow{3}{*}{transport} & IPC transport08& \multirow{3}{*}{\citet{seipp-et-al-zenodo2022}}\\
           & IPC transport11& \\
           & IPC transport14 & \\
           \hline
         \multirow{4}{*}{visitall} & \citet{stein-25-icaps} & \multirow{4}{*}{\citet{seipp-et-al-zenodo2022}}\\
         & IPC visitall11 & \\
         & IPC visitall14 & \\
         & Generated by us & \\
         \hline
    \end{tabular}
    \caption{The origin of all tasks that we used for our experiments and the instance generators that we used for the manual evaluation and for generating additional data for some of the domains.}
    \label{tab:data_sources}
\end{table}

\paragraph{Debugging and eval tasks.} For the each domain, we randomly select 6 tasks that are small with respect to the sized of the evaluation tasks as debugging tasks. In particular, we only consider tasks for which the number of objects and optimal plan length of each debugging task is among the 16 smallest values of object number and plan length in the overall dataset. One exception is the Beluga domain for which the optimal planner (\exLMC) did not solve any task. We therefore based the debugging task selection on the lengths of the plans generated by the satisficing planner (\exFF).

\begin{table*}[!ht]
\fontsize{8pt}{8pt}\selectfont
    \centering
    \begin{tabular}{|l|r||r|r|r|r||r|r|r|r||r|r|r|r|}
    \hline
        \multirow{3}{*}{domain} & \multirow{3}{*}{N} & \multicolumn{4}{c||}{Optimal Plan Length (\exLMC)} & \multicolumn{4}{c||}{Satisficing Plan Length (\exFF)} & \multicolumn{4}{c|}{Number of objects} \\ 
        & & debug & \multicolumn{3}{c||}{eval} & debug & \multicolumn{3}{c||}{eval} & debug & \multicolumn{3}{c|}{eval} \\
         &  & \multicolumn{1}{c|}{avg} & \multicolumn{1}{c|}{avg} & \multicolumn{1}{c|}{min} & \multicolumn{1}{c||}{max}  & \multicolumn{1}{c|}{avg}  & \multicolumn{1}{c|}{avg} & \multicolumn{1}{c|}{min}  & \multicolumn{1}{c||}{max} & \multicolumn{1}{c|}{avg} & \multicolumn{1}{c|}{avg} & \multicolumn{1}{c|}{min} & \multicolumn{1}{c|}{max}   \\ 
          \hline\hline
        \multicolumn{14}{|l|}{\quad Domains from \citet{Silver_2024}}\\
         \hline
        delivery & 30 & 10 & None & None & None & 12 & 96 & 79 & 115 & 10 & 62 & 50 & 73 \\ \hline
        ferry & 275 & 7 & 28 & 4 & 56 & 7 & 104 & 4 & 301 & 8 & 45 & 5 & 116 \\ \hline
        gripper & 53 & 9 & 35 & 15 & 52 & 10 & 77 & 18 & 165 & 28 & 53 & 10 & 81 \\ \hline
        heavy & 34 & 6 & 128 & 4 & 209 & 6 & 128 & 4 & 209 & 6 & 128 & 4 & 209 \\ \hline
        hiking & 28 & 7 & 13 & 2 & 26 & 7 & 13 & 2 & 26 & 97 & 108 & 80 & 121 \\ \hline
        miconic & 34 & 23 & 27 & 11 & 
        66 & 25 & 63 & 12 & 186 & 20 & 53 & 9 & 104 \\ \hline
        spanner & 34 & 16 & 33 & 10 & 52 & 16 & 37 & 10 & 52 & 19 & 47 & 13 & 64 \\ \hline
         \hline
        \multicolumn{14}{|l|}{\quad Additional Domains}\\
         \hline
        beluga & 61 & None & None & None & None & 12 & 48 & 10 & 159 & 17 & 27 & 15 & 34 \\ \hline
        blocksworld & 191 & 7 & 18 & 0 & 42 & 9 & 38 & 0 & 218 & 4 & 9 & 3 & 20 \\ \hline
        goldminer & 115 & 10 & 16 & 5 & 32 & 11 & 42 & 5 & 370 & 7 & 17 & 4 & 49 \\ \hline
        grippers & 131 & 5 & 18 & 4 & 54 & 5 & 126 & 4 & 340 & 11 & 63 & 9 & 144 \\ \hline
        logistics & 178 & 7 & 17 & 0 & 48 & 7 & 59 & 0 & 361 & 12 & 54 & 9 & 438 \\ \hline
        minigrid & 74 & 6 & 12 & 0 & 83 & 6 & 13 & 0 & 122 & 25 & 34 & 10 & 96 \\ \hline
        rovers & 25 & 8 & 14 & 6 & 36 & 8 & 15 & 8 & 37 & 12 & 16 & 11 & 28 \\ \hline
        satellite & 25 & 6 & 10 & 6 & 23 & 6 & 17 & 6 & 79 & 8 & 31 & 9 & 110 \\ \hline
        transport & 53 & 10 & 22 & 16 & 36 & 10 & 55 & 16 & 152 & 14 & 37 & 18 & 75 \\ \hline
        visitall & 193 & 5 & 15 & 0 & 50 & 5 & 87 & 0 & 2308 & 6 & 31 & 1 & 324 \\ \hline
    \end{tabular}

\caption{\label{tab:task_sizes}
The number of tasks for each domain (N), and the average (avg), minimum (min) and maximum (max) values of the plans derived by the \exLMC and \exFF symbolic planners and number of objects for the evaluation tasks (eval) as well as the average values of the debugging tasks (debug). Tasks for which the symbolic planner did not find a plan were left out in the computation of the average plan length values. 
}
\end{table*}

Table \ref{tab:task_sizes} gives an overview of the sizes of all tasks from our experiments. It shows the average length of the plans generated by the optimal planner and the satisficing planner for the debugging and evaluation tasks, as well as the average number of objects for both sets of tasks. Additionally, we include the minimum and maximum values for the plan lengths and number of objects of the evaluation tasks. The overview shows that the tasks on which we evaluate our generate programs include tasks that are much larger than the ones used during the generation of the generalized plans (i.e. as examples and for debugging). Figure \ref{fig:nobj_distr} illustrates the distribution of the number of objects per debugging and evaluation task for the Miconic and Logistics domains. 

\begin{figure*}[ht]
\centering
    \subfloat[Miconic domain]{\includegraphics[width=0.6\linewidth]{pictures_appendix/nobj_counts/miconic_nobj_counts.pdf}\label{fig:nobj_distr_spanner}}
    \hfill
    \subfloat[Logistics domain]{\includegraphics[width=0.8\linewidth]{pictures_appendix/nobj_counts/logistics_nobj_counts_wide.pdf}\label{fig:nobj_distr_logistics}}
    \hfill
    \caption
    {\label{fig:nobj_distr}
 Number debugging tasks (blue) and number of evaluation tasks (orange) with a specific number of objects (x-axis).
}
\end{figure*}

\subsection{Costumed variants}
\citet{duchnowski-etal-2025-knapsack} showed for NP-hard problems that different ways of phrasing the same mathematical problem impact the performance of LLMs. We adapt their idea of generating costumed versions to planning domains. For each actions, predicate, object and type name we manually create a new replacement name, hence generating a new variant of a domain that preserves the exact logical structure of the original domain. We provide a brief description of the original and costumed versions. 

\paragraph{Ferry.} Cars must be transported between different locations using a ferry that can carry only one car at a time.
\paragraph{Costumed ferry.} A squirrel needs to jump between different trees in order to move nuts to the goal tree. It can only carry one nut in its paws.

\paragraph{Delivery.} The goal is delivery newspapers to different locations. All newspapers need to be picked up at the home based and be carried to the locations that want a newspaper.

\paragraph{Costumed delivery.} The goal is to deliver seedlings to places that are planning to create a garden. All seedlings need to be collected at the nursery and driven to the target location where the seedling is planted. 

\paragraph{Gripper.} A robot needs to carry balls between rooms. The robot has two grippers, each can carry one ball.

\paragraph{Costumed gripper.} The goal is to sail between different hideouts in order to find chests at their initial hideout and hide them at the goal hideout. The boat has space for two chests, one at the port and one at the starboard side.

\paragraph{Heavy.} Objects need to be placed on top of each other in a box such that the heaviest item is the bottom-most one and no object is placed on an object that is lighter than itself.

\paragraph{Costumed heavy.} A number of tasks needs to be scheduled such that the easiest task is scheduled first and no task is scheduled after a more difficult task.

\paragraph{Hiking.} The goal is to navigate from an initial location in a 2D grid to a goal location. Some locations are water or a hill. Moving to a hill location requires a climbing action instead of walking and it is not possible to move to locations with water. There is one defined trail leading from the initial to the goal location but other paths are possible.

\paragraph{Costumed hiking.} The goal is to navigate from an initial location in a 2D grid to a goal location. Some locations are colored white, black or red. Moving to the white locations requires a jump actions instead of moving and it is not possible to move to a black location. All red locations form a path from the initial to the goal location. 

\paragraph{Miconic.} The goal is to transport passengers between different floors in a building. Passengers can be picked up at their initial floor and dropped off at another floor. There can be several buildings and passengers can only move within the same building. 

\paragraph{Costumed miconic.} The goal is submit a process when a machine is in a specific mode.
There can be several machines and each machine has different modes that are ordered. It is possible to switch the mode down to lower modes or up to higher modes. Each process requires a specific mode in order to be set up before the mode can be switched to the goal mode for submitting the process.  

\paragraph{Spanner.} An agent must tighten all nuts using spanners. Each spanner can only be used once to tighten a nut. All locations are connected in the form of a one-way path and all nuts are at the last location. The agent needs to move from the its start location to the last location and pick up the spanners needed for tightening the nuts along the way. 

\paragraph{Costumed spanner.} A squirrel must feed its babies with nuts. There is a number of tree hollows along a tree that contain nuts and the squirrel can only move from the bottom of the tree to the top of the tree where the babies are located. It needs to climb upt the tree and pick up the nuts needed for feeding each baby one nut. 

\section{Details about Inputs of Pipeline Parts}

\begin{table*}[t]
    \centering
    \begin{tabular}{|l|l|l|l|l|}
    \hline
    \multirow{2}{*}{Main Step} & \multirow{2}{*}{Sub-step} & \multicolumn{2}{c|}{Input to LLM } & \multirow{2}{*}{Output} \\
    & & LLM history & New information in addition to instructions &  \\
    \hline\hline
        1) NL & 1.1 NL domain & --- & - PDDL domain definition & NL domain\\
        \hdashline
                & 1.2 NL task & step 1.1 & - PDDL task definition & NL task \\
                \hline
        2) strategy & 2.1 first pseudocode & --- & - NL domain and 2 NL tasks & pseudocode \\
                      
                      \hdashline
                      & 2.2 strategy validation & --- & - NL domain and 1 NL task & PDDL plan \\
                      & & & - pseudocode (first or revised) & \\
                      & & & - valid PDDL actions (lifted) &\\
                      \hdashline
                      & 2.3 reflection & step 2.2 & - feedback for wrong PDDL plan & reflection on mistake\\
                      \hdashline
                      & 2.4 strategy revision & step 2.2 - 2.3 & --- & pseudocode (revised)\\
                      \hline
        3) code & 3.1 first code & --- & - NL domain & program \\
                  & & & - pseudocode (best) & \\
                  & & & - function signature & \\
                  & & & - types of the inputs and output & \\
                  & & & - valid PDDL predicates and actions (lifted) & \\
                  & & & - example of input and output & \\
                  \hdashline
                  & 3.2 reflection & step 3.1 & - input and output of solved tasks (if any) & reflection on mistake \\
                  & & & - input of one failed task & \\
                  & & & - returned output (if any) of the failed task & \\
                  & & & - feedback message about type of mistake & \\
                  \hdashline
                  & 3.3 code revision & step 3.1 - 3.2 & --- & program (revised)\\
    \hline
    \end{tabular}
    \caption{The information provided as input to the LLMs at the different steps in our fulle pipeline and the output generated. The LLM history specifies from which previous step(s) the inputs and outputs are kept as part of the history (i.e. context). At each step, specific instructions about the output (content and format) are provided in addition to the information listed in the table.}
    \label{tab:inputs}
\end{table*}

\paragraph{Inputs and outputs.} Table \ref{tab:inputs} provides an overview of the information that is provided to the LLM at each generation step of our framework and the output generated during this step. At each of the steps, the LLM is provided an instruction specifying what to generate and what format the output should have. The instructions are not explicitly listed in Table \ref{tab:inputs}. 

At some of the steps, only one prompt (and the system prompt) is provided to the LLM (e.g. steps 1.1 and 2.1). For other steps, the inputs and outputs of the previous step(s) are still kept in the LLM history and the prompt with the new instructions and information is added (e.g. steps 1.2 and 2.3).

\begin{figure*}[!ht]
\centering
    \subfloat[]{\includegraphics[width=0.38\linewidth]{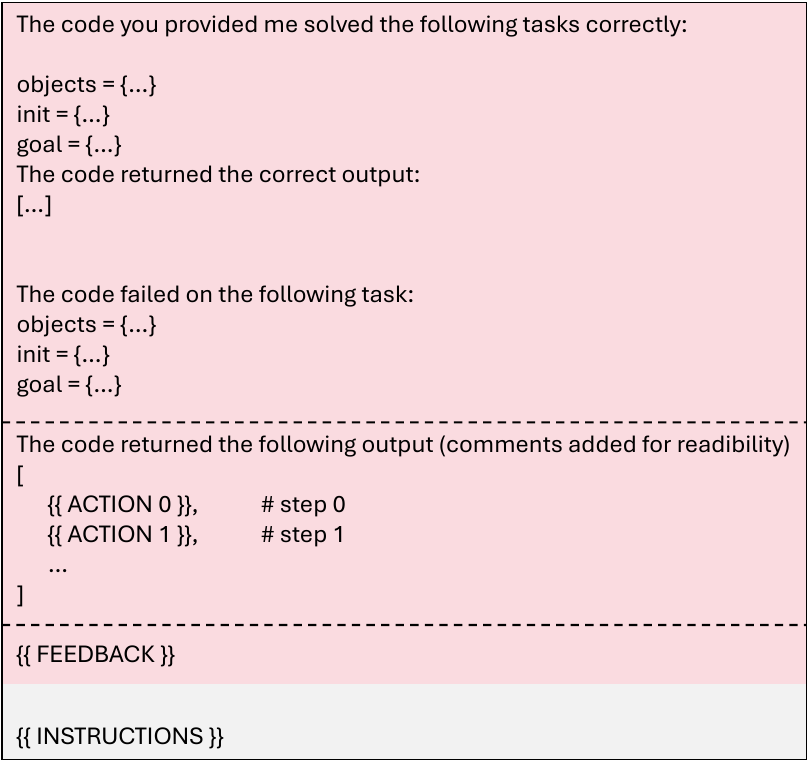}\label{fig:feedback_pos}}
    \hfill
    \subfloat[]{\includegraphics[width=0.3\linewidth]{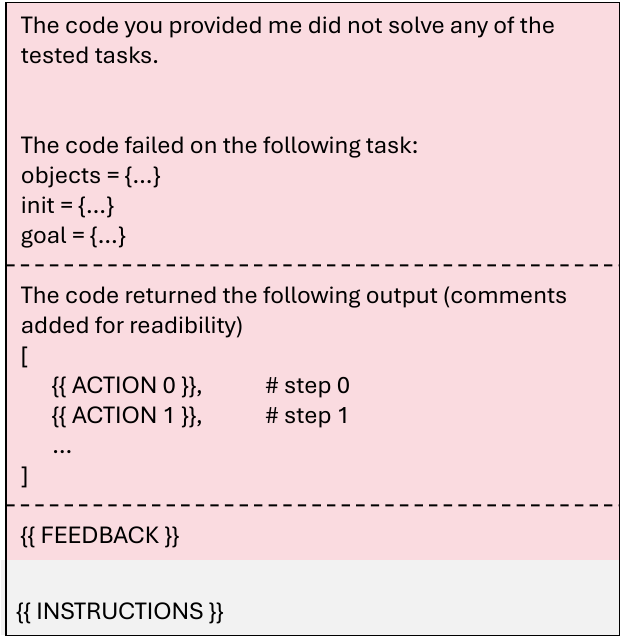}\label{fig:feedback_none_solved}}
    \hfill
    \subfloat[]{\includegraphics[width=0.3\linewidth]{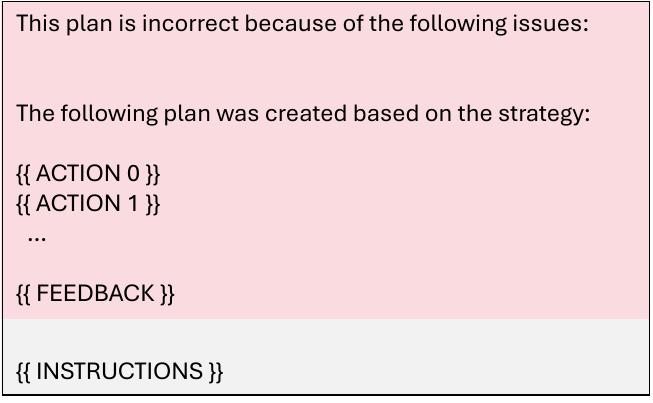}\label{fig:feedback_plan}}
    \hfill
    \caption
    {\label{fig:feedback_prompts}
   \ref{fig:feedback_pos} and \ref{fig:feedback_none_solved} show the templates for the reflection prompts provided during the code debugging step when some of the debugging tasks were solved by the program (\ref{fig:feedback_pos}) and none were solved (\ref{fig:feedback_none_solved}). The part between the dashed lines is only included if running the program on the task did actually generate an output. \ref{fig:feedback_plan} shows the template for the reflection prompt provided during the debugging of the pseudocode strategy. 
}
\end{figure*}

\paragraph{Feedback code debugging.} For the debugging of the generated Python programs, the LLM is provided feedback about the outcome of running the program on the debugging task. In particular, we provide feedback on one of the tasks for which the program did not return any output or an incorrect output. For our approach (but not for the baseline), we also include the tasks for which the program returned the correct output together with the actual outputs. Figure \ref{fig:feedback_pos} and \ref{fig:feedback_none_solved} show the templates used for creating the feedback messages if at least one task was solved by the program and if none was solved respectively. 

We provide the information about the failed task (and the solved ones if available) in Python format.
If the program returned an incorrect output, this output is included in the feedback prompt (part between the dashed lines). The actual feedback message depends on the type of error that occurred. 

Table \ref{tab:feedbacks} shows the feedback messages generated in our pipeline for the different types of errors.
Following \citet{Silver_2024} we differentiate between timeouts (1), Python exceptions (2), an output of an invalid type (3) and outputs that are not a valid plan for the task (4). We make the feedback message for outputs that are not valid plans for the input task more informative by incorporating the feedback generator from \citet{stein-25-icaps} (see Table \ref{tab:feedbacks}, 4.1 - 4.6). This approach allows us to give more precise feedback, e.g. for the wrong number of action parameters, or parameters not matching objects defined by the specific task.

\paragraph{Feedback pseudocode debugging.} Figure \ref{fig:feedback_plan} shows the template used for creating the feedback prompt for the debugging of the pseudocode strategy. The prompt does not include the definition of the task for which a wrong plan was generated because the NL task description is already included in the context of the LLM. The different types of errors considered for the pseudocode debugging and the corresponding feedback messages are the same as 4.1 - 4.6 in Table \ref{tab:feedbacks}. 

\begin{table*}[ht]
    \centering
    \begin{tabularx}{\textwidth}{r|h|s}
        & \textbf{Feedback Messages} & \textbf{Error Type} \\
        \hline\hline
       1 & The code was interrupted because it did not terminate within the time limit ($\{\{$ TIME LIMIT $\}\}$ seconds). This is likely caused by an infinite loop. Please check the loops again. & Execution of the program timed out. \\
        \hline
        2 &The code raised the following exception:\newline$\{\{$ TRACEBACK without file paths $\}\}$ & Execution of the program raised a Python exception. \\

        \hline
        3 & The code returned $\{\{$ OUTPUT $\}\}$ which is not a list of actions. Please make sure that your code returns a list of actions, i.e. of type List[str].
 & The returned output does not have the correct type. \\
    \hline\hline
    4.1 & The action $\{\{$ ACTION X $\}\}$ at step $\{\{$ X $\}\}$ is not executable because $\{\{$ STRING $\}\}$ is not an available object in this task. &  The returned plan contains parameters that do not match objects in this task.\\
    \hline
    4.2 & The action $\{\{$ ACTIONX $\}\}$ at step $\{\{$ X $\}\}$ is not executable because $\{\{$ ACTIONX $\}\}$ does not match any possible actions. & The returned plan contains actions that are not part of the domain.\\
    \hline
    4.3 & The action $\{\{$ ACTIONX $\}\}$ at step $\{\{$ X $\}\}$ is not executable because $\{\{$ ACTIONX $\}\}$ requires exactly $\{\{$ CORRECT NUMBER PARAMETERS $\}\}$ objects as arguments but $\{\{$ INCORRECT NUMBER PARAMETERS $\}\}$ were given. &  The returned plan contains actions with wrong number of parameters.\\
    \hline
    4.4 & The action $\{\{$ ACTIONX $\}\}$ at step $\{\{$ X $\}\}$ is not executable because the preconditions of the action are not satisfied: \newline
    At that specific step\newline
    $-$ it is not the case that $\{\{$ PRECONDITION $\}\}$ &  The returned plan contains actions with unsatisfied preconditions which are non-static predicates, i.e. there are actions that can make the precondition true. \\
    \hline
    4.5 & The action $\{\{$ ACTIONX $\}\}$ at step $\{\{$ X $\}\}$ is not executable because the preconditions of the action are not satisfied: \newline
    In this task instance\newline
    $-$ it is not the case that $\{\{$ PRECONDITION $\}\}$ & The returned plan contains actions with unsatisfied preconditions which are static predicates, i.e. there is no action that can make this precondition true.\\
    \hline
    4.6 & The generated plan does not reach the goal:\newline
    The following needs to be false but is true after executing all actions:\newline
    $\{\{$ NOT SATISFIED NEGATIVE GOAL FACTS $\}\}$\newline
    The following needs to be true but is false after executing all actions:\newline
    $\{\{$ NOT SATISFIED POSITIVE GOAL FACTS $\}\}$& The returned plans does only contain applicable actions but not all goal facts are satisfied in the end. \\
    \hline
    \hline
    \end{tabularx}
    \caption{The feedback messages generated for the different types of errors.}
    \label{tab:feedbacks}
\end{table*}

\section{Manual Evaluation of Generalization Power}

We manually evaluated the generalization power of all generalized plans generated by \exfullinit using GPT-4o and \exfullinit using DeepSeek-R1 that achieved 100\% coverage on our evaluation datasets. We find that for all 12 domains where \exfullinit with GPT-4o generated at least one program that solved all evaluation tasks, that program generalizes to all tasks that can be generated with the respective instance generators. The some holds for the 14 domains for which \exfullinit generated at least one program that solved all evaluation tasks when using DeepSeek-R1.

Here, we provide more details about the results of the manual analysis. We report the details for the programs generated by GPT-4o but the observations for the programs generated by DeepSeek-R1 are similar. For each domain, we provide a summary of the types of tasks generated by the instance generator and briefly describe the most relevant parts of the evaluated Python program. If several programs achieved 100\% coverage, we evaluated all of them but provide a description only for one of them. All evaluated programs are included in the supplementary material. 
We report whether the evaluated programs can generalize to all tasks generated by the instance generator. Additionally, we provide a brief overview of the generalization beyond the tasks generated by the specific instance generators. For both, we only consider solvable tasks. 

\subsection{Delivery} 
\paragraph{Instance generator.} The generator creates tasks with a specified number of locations. ``location-0'' is always the home base and always the start location. Newspapers must be picked up in the home base, moved to a location that wants a paper and then be delivered. The locations that want a newspaper are randomly distributed over the locations. The goal specifies that each location that wants a paper should be satisfied in the end.
\paragraph{Program description (seed 1).} First, the home base, all safe locations, all locations that want newspapers and all unpacked newspapers are determined. Afterwards, the code loops over all locations that want a newspaper and adds all actions required for delivering a newspaper from the home base to that location to the solution.
\paragraph{Generalization.} All three evaluated programs solve every task generated by the instance generator. They generalize to tasks with an arbitrary number of locations, newspapers and locations that want packages. 
\paragraph{Generalization beyond generator.} If the start location is different from the home base, two of the programs will not generate correct plans as they set the start location to the home base. Additionally, all three programs will fail if the goal specifies the final position of the deliverer. However, all programs generalize to tasks where the home base, i.e. the location of the newspapers, is different from ``location-0''. 

\subsection{Ferry}
\paragraph{Instance generator.} For all generated tasks, there is a ferry, a number of cars, and a number of possible locations. The cars are randomly distributed across locations. The ferry is initially empty. The goal specifies for each car a goal location, which can be identical to the initial location.
\paragraph{Program description (seed 1).} First, all cars and the initial location of the ferry are determined. Then the program loops over all cars and checks whether the current and goal location are the same. If not, all actions for transporting the car to the goal location are added to the solution. 
\paragraph{Generalization.} All three evaluated programs solve every task generated by the instance generator. They generalize to tasks with an arbitrary number of cars and locations. 
\paragraph{Generalization beyond generator.} The programs cannot generalize to tasks where a car is initially already on the ferry. Additionally, the programs will fail on tasks where the goal specifies a target position of the ferry or if a car needs to be on the ferry.

\subsection{Gripper}
\paragraph{Instance generator.} The instance generator generates initial states where the balls are randomly distributed over all rooms. The robot always starts in ``room-0'' and has two grippers that are initially free. The generated goals specify for some of the balls one room as the goal location, i.e. the goal is always to transport some, but not necessarily all, balls to the goal room.
\paragraph{Program description (seed 1).} First, the initial location of the robot is determined and afterwards the program loops over the goal input set and checks for goal facts starting wit ``at''. For each of those, it checks for the current location, moves the robot there, frees up a gripper and continues with the remaining actions required for moving the ball to the goal location. 
\paragraph{Generalization.}  All three evaluated programs solve every task generated by the instance generator. They generalize to tasks with an arbitrary number of balls, rooms and goal locations.
\paragraph{Generalization beyond generator.} All three programs cannot generalize to tasks where the goal specifies a target location for the robot. Additionally, they do not generalize to tasks where balls are already being carried in the initial state.

\subsection{Heavy}
\paragraph{Instance generator.} For all generated tasks, the box is initially empty and all objects are unpacked. Additionally, the initial state fully defines the heavier relation between all objects, i.e. if there are $n$ objects, then the heaviest object is heavier than $n-1$ objects, the second heaviest is heavier than $n-2$, and so on. The goal is that every object is packed in the box. 
\paragraph{Program description.} All three programs organize the objects into a list, sorted in descending order based on the frequency with which each object is heavier than another object. To generate the plan, the objects are stacked on top of each other in exactly that order. 
\paragraph{Generalization.} All three evaluated programs solve every task generated by the instance generator. They generalize to tasks with an arbitrary number of objects. 
\paragraph{Generalization beyond generator.} All programs pack all available objects into the box. If the goal state specifies that some objects should not be packed the programs will not be able to generalize valid plans.

\subsection{Hiking}
\paragraph{Instance generator.} The instance generators generate tasks that are grids where some cells are of type dirt, water or hill. In the initial state, these types are randomly distributed over the cells and a trail from the initial start position to the target position is generated, consisting of cells which are not of type water. The goal is to reach the target position. This can be achieved by simply following the trail, or finding another path through the grid.
\paragraph{Program description.} First, the start and target position are determined. Afterwards, the program loops over all facts of the initial state to find the location that is adjacent to the current one and on the trail. If that location is a hill, the action for climbing to the location is added to the solution, otherwise the action for walking is added. This process is continued until the current location equals the goal location. 
\paragraph{Generalization.} The evaluated program solves every task generated by the instance generator. It generalizes to tasks with an arbitrary grid size, as long as there exists a trail between the initial and target position. 
\paragraph{Generalization beyond generator.} If the trail is interrupted, or the initial or target position are not part of the trail, then the program will fail.

\subsection{Miconic}
\paragraph{Instance generator.} The generator generates tasks with a specified number of buildings. Every building has the same number of floors and passengers. The passengers are then randomly distributed across the floors of a building. Furthermore, every building has one lift which is initially on a random floor. The goal is to bring all passengers from their initial floor to their destination floor within the same building using the lifts.
\paragraph{Program description (seed 2).} First, the position of the lifts is determined, as well as the connections between the floors, i.e. which floors are in the same building, and the current and destination floor for each passenger. The program then loops over all passengers and their initial location, moves the lift that is in the same building to the passenger and adds all remaining actions for moving the passenger to the destination floor.  
\paragraph{Generalization.} All three evaluated programs solve every task generated by the instance generator. They generalize to tasks with an arbitrary number of buildings, passengers and floors. 
\paragraph{Generalization beyond generator.} If the goal specifies a floor as target location for any lift, then the all three programs will fail. Additionally, if the names of the floors do not follow the naming scheme of FLOOR$-$BUILDING (e.g. f1-b0) the programs cannot correctly determine anymore which floors belong to the same building.

\subsection{Spanner}
\paragraph{Instance generator.} The instance generator generates tasks with a specified number of locations, and two special locations, the shed and the gate. In the initial state, a man is at the shed and a number of loose nuts is at the gate. An arbitrary number of spanners is distributed across the locations and all locations are connected such that they form a one-way path from the shed to the gate. The goal is to tighten all loose nuts. 
\paragraph{Program description.} First, a direct path from the gate to the shed is determined using a breadth-first search approach. Then the man is moved from location to location along this path. If there are spanners at an location they are all picked up. After arriving at the gate, the program loops over the loose nuts and selects one spanner after the other to tighten each nut. 
\paragraph{Generalization.} The evaluated program solves every task generated by the instance generator. It generalizes to tasks with an arbitrary number of nuts, locations and spanners. 
\paragraph{Generalization beyond generator.} The program will fail if the nuts are not only at the gate location or if the goal specifies a target location of the man. Additionally, the program would fail if the connections between locations would allow moving in more than one direction, and the man would need to move between locations in a way different from a direct path between shed and gate. 

\subsection{Grippers}
\paragraph{Instance generator.} The instance generator generates initial states where the robots and balls are randomly distributed over all rooms, and all robots have two grippers that are initially free. The generated goals specify for each ball one room as the goal location, i.e. the goal is always to transport each ball, that is not already at its goal location, to the goal room. In contrast to Gripper above, there can be several robots.
\paragraph{Program description (seed 1).} First, all robots and objects (i.e. balls) are determined. The program then loops over all facts in the goal input set and checks for goal facts starting with ``at''. For each of them, it checks whether the initial location of the specified ball is identical to the goal location. If not, it selects a robot, moves it to the initial location of the ball and frees up a gripper if necessary, and continues with the remaining actions required for moving the ball to the goal room.
\paragraph{Generalization.} All three evaluated programs solve every task generated by the instance generator. They generalize to tasks with an arbitrary number of balls, rooms, and robots. 
\paragraph{Generalization beyond generator.} All three programs cannot generalize to tasks where the goal specifies a target location for any robot. Additionally, they do not generalize to tasks where balls are already being carried in the initial state.

\subsection{Logistics}
\paragraph{Instance generator.} In all initial states generated by the instance generator, all cities have the same number of locations and each city has one airport.
A specified number of airplanes are randomly distributed across all airports. Each task also includes a specified number of trucks, with the only condition that there are at least as many trucks as cities. There can be multiple trucks and airplanes in the same location. A specified number of packages is distributed over all possible locations. 
The goal specifies for each package a goal location which can be identical to the initial location. 
\paragraph{Program description (seed 2).} First, the locations of all trucks and a randomly selected airplane are determined. Afterwards, the program loops over all facts in the goal input set and identifies all goal facts that are about the location of packages. For each of them, it checks whether the current location and goal location are in the same city or not. For both cases, the required vehicles and steps for moving the package from the current location to the goal are then determined. 
\paragraph{Generalization.} All three evaluated programs solve every task generated by the instance generator. They generalize to tasks with an arbitrary number of vehicles, cities, locations, and packages. 
\paragraph{Generalization beyond generator.} All three programs cannot generalize to tasks where the location of a vehicle is part of the goal or where the target location of a package is inside a vehicle. If some packages are initially in a vehicle the generalized plans will fail as well. 

\subsection{Satellite}
\paragraph{Instance generator.} The initial state of the tasks generated defines which instruments a satellite has on board and the modes the instruments support. Furthermore, it gives the calibration targets for the instruments. Additionally, all satellites have ``power\_avail'' which is needed to switch on instruments, and satellites are pointing in some direction. The goal is to take images of some observations and, in some instances, to additionally point the satellite to a specific direction (target or observation).
\paragraph{Program description.} The program first loops over all facts in the goal input set and checks for goals that are about taking images. It then determines the required mode, instrument, satellite and calibration target and adds all steps taking the picture and turning off the instruments afterwards. After taking care of all pictures, the code loops over all goal facts that are about pointing in some direction and adds the actions required for turning the satellites accordingly.
\paragraph{Generalization.} 
The evaluated program solves every task generated by the instance generator. It generalizes to tasks with an arbitrary number of satellites, modes, targets, observations and maximum number of instruments per satellite. 
\paragraph{Generalization beyond generator.} If in the initial state, there is no ``power\_avail'' for some satellites, i.e. an instrument has ``power\_on'' from the beginning, the program might fail depending on whether that specific instrument is used first or not. The order in which the instruments are used depends on the order in which the program iterates over an (unordered) set.

\subsection{Transport}
\paragraph{Instance generator.} Each task generated by the instance generator consists of a specified number of locations, vehicles (trucks), packages, and vehicle capacity. Initially, packages and trucks are distributed across the available locations and trucks are assigned a capacity of at least 2. Additionally, there exist roads between locations. The goal is to bring every package to its goal location.
\paragraph{Program description.} First, all goal locations of packages are determined. The program then loops over all packages and their initial locations, checks for a vehicle with capacity that is not equal to ``capacity-0'' and finds a path from the location of the vehicle to the package using a breadth-first search based approach. Then the vehicle is move there, the package is picked-up, the capacity is updated and the path for getting to the goal location is determined in the same way. The package is dropped at the goal location, the capacity is updated and the loop continues with the next package.
\paragraph{Generalization.} 
The evaluated program solves every task generated by the instance generator. It generalizes to tasks with any number of trucks, packages, locations, cities, roads, and capacity numbers. 
\paragraph{Generalization beyond generator.} Similar to Logistics, the program cannot generalize to tasks where the location of a vehicle is part of the goal or where the target location of a package is inside a vehicle. Additionally, if a package is initially already in a truck the program will fail as well. 

\subsection{Visitall}
\paragraph{Instance generator.} The generator creates tasks consisting of a grid of some size $n$ x $m$. The initial state consists of the random location of the robot and the connections between adjacent locations. It is possible to define unavailable locations, i.e. locations for which no connections to other locations in the grid are defined. Depending on the ratio of cells in the goal state, the goal is to reach either all locations or a random subset of all locations.
\paragraph{Program description.} First, the initial location of the robot and all connections between locations are determined, as well as all already visited and not yet visited locations. The program then loops over all locations adjacent to the current location to find a not yet visited one. If one is found, the robot moves there. Otherwise, the program backtracks to the last visited location that has still unvisited adjacent locations (and moves the robot there), and continues from there. 
\paragraph{Generalization.} The evaluated program solves every task generated by the instance generator. It generalizes to tasks with an arbitrarily large grid, any number of unavailable locations, and any ratio of locations in the goal state. 
\paragraph{Generalization beyond generator.} The program would fail only for tasks where the goal specifies a specific target position of the robot.

\newpage
\onecolumn
\section{Example Generalized Plans}

\subsection{Generalized Plan for Gripper (DeepSeek, F5-3)}
\begin{python}
from typing import List, Tuple, Set

def generate_solution(objects: Set[str], init: Set[Tuple], goal: Set[Tuple]) -> List[str]:
    solution = []
    
    # Extract initial state information
    robby_location = None
    ball_locations = {}
    grippers = []
    free_grippers = []
    
    for fact in init:
        if fact[0] == 'at-robby':
            robby_location = fact[1]
        elif fact[0] == 'at' and len(fact) == 3:
            ball_locations[fact[1]] = fact[2]
        elif fact[0] == 'gripper':
            grippers.append(fact[1])
        elif fact[0] == 'free':
            free_grippers.append(fact[1])
    
    # Extract goals
    goals = []
    for fact in goal:
        if fact[0] == 'at' and len(fact) == 3:
            goals.append((fact[1], fact[2]))
    
    # Process each goal
    for ball, target_room in goals:
        current_room = ball_locations[ball]
        
        # Skip if ball is already in target room
        if current_room == target_room:
            continue
        
        # Move Robby to ball's current room if needed
        if robby_location != current_room:
            solution.append(f"(move {robby_location} {current_room})")
            robby_location = current_room
        
        # Pick up the ball with a free gripper
        if free_grippers:
            gripper = free_grippers.pop(0)
            solution.append(f"(pick {ball} {current_room} {gripper})")
            
            # Move Robby to target room if needed
            if robby_location != target_room:
                solution.append(f"(move {robby_location} {target_room})")
                robby_location = target_room
            
            # Drop the ball
            solution.append(f"(drop {ball} {target_room} {gripper})")
            free_grippers.append(gripper)
            
            # Update ball location
            ball_locations[ball] = target_room
    
    return solution
\end{python}

\subsection{Generalized Plan for Blocksworld (DeepSeek, F5-3)}

\begin{python}
from typing import List, Tuple, Set

def generate_solution(objects: Set[str], init: Set[Tuple], goal: Set[Tuple]) -> List[str]:
    solution = []
    
    # Initialize current state
    current_clear = set()
    current_ontable = set()
    current_hand = None
    current_on = {}
    
    for fact in init:
        if fact[0] == 'clear':
            current_clear.add(fact[1])
        elif fact[0] == 'ontable':
            current_ontable.add(fact[1])
        elif fact[0] == 'handempty':
            current_hand = None
        elif fact[0] == 'holding':
            current_hand = fact[1]
        elif fact[0] == 'on':
            current_on[fact[1]] = fact[2]
    
    # Phase 1: Move all blocks to the table
    while any(obj not in current_ontable for obj in objects):
        if current_hand is not None:
            # Put down the block in hand
            block = current_hand
            solution.append(f"(put-down {block})")
            current_ontable.add(block)
            current_clear.add(block)
            current_hand = None
        else:
            # Find a clear block that is not on the table
            found_block = None
            for block in current_clear:
                if block not in current_ontable:
                    found_block = block
                    break
            if found_block is None:
                break
                
            base_block = current_on[found_block]
            solution.append(f"(unstack {found_block} {base_block})")
            current_hand = found_block
            current_clear.remove(found_block)
            current_clear.add(base_block)
            del current_on[found_block]
    
    # Phase 2: Build goal stacks
    goal_on = {}
    for fact in goal:
        if fact[0] == 'on':
            goal_on[fact[1]] = fact[2]
    
    # Compute goal stacks
    goal_ontable = objects - set(goal_on.keys())
    goal_children = {}
    for top, bottom in goal_on.items():
        goal_children[bottom] = top
    
    stacks = []
    visited = set()
    for block in goal_ontable:
        if block in visited:
            continue
        stack = [block]
        visited.add(block)
        current = block
        while current in goal_children:
            next_block = goal_children[current]
            stack.append(next_block)
            visited.add(next_block)
            current = next_block
        stacks.append(stack)
    
    # Build each stack from bottom up
    for stack in stacks:
        for i in range(1, len(stack)):
            solution.append(f"(pick-up {stack[i]})")
            solution.append(f"(stack {stack[i]} {stack[i-1]})")
    
    return solution

\end{python}

\end{document}